\documentclass[final]{siamltex}
\usepackage{subfigure}
\usepackage{times}
\usepackage{epsfig}
\usepackage{graphicx}
\usepackage{rotating}
\usepackage{multirow}
\usepackage{array}
\usepackage{amsmath}
\usepackage{amssymb}
\usepackage{lscape}
\usepackage{setspace}
\usepackage[sectionbib,numbers,sort&compress]{natbib}
\usepackage{float}
\usepackage{url}
\usepackage[lined,boxed]{algorithm2e}
\usepackage{rotating}

\title{A robust framework for moving-object detection and vehicular traffic density estimation}
\author{Pranam Janney\thanks{National ICT Australia (NICTA) and Department of Computer Science and Engineering, 
        The University of New South Wales (UNSW), Australia. (Email: {\tt pranam@unswalumni.com})} \and Glenn Geers \footnotemark[1] 
}

\begin{document}
%

\maketitle

\begin{abstract}
Intelligent machines require basic information such as moving-object detection from videos in order to deduce higher-level semantic information. In this paper, we propose a methodology that uses a texture measure to detect moving objects in video. The methodology is computationally inexpensive, requires minimal parameter fine-tuning and also is resilient to noise, illumination changes, dynamic background and low frame rate. Experimental results show that performance of the proposed approach is higher than those of state-of-the-art approaches.  We also present a framework for vehicular traffic density estimation using the foreground object detection technique and present a comparison between the foreground object detection-based framework and the classical density state modelling-based framework for vehicular traffic density estimation.

\end{abstract}


\section{Introduction}
Urban environments are inundated with installations of closed-circuit television (CCTV) camera-based surveillance systems, currently used for monitoring traffic, security, etc. Video analytics are useful in lending a helping hand to people monitoring these surveillance systems. Foreground detection is one of the fundamental processes in video analysis. Higher level processes could use the information provided by foreground detection to hypothesize scenarios and activities occurring in the video. Motion detection is one of the methods of detecting foreground objects in videos.

\textit{Background subtraction}-based techniques are the most widely used approach for moving object detection with a fixed camera in the absence of any \textit{a priori} knowledge about objects or background~\cite{amam,gmm1,harita,lbp,ohta,sigdel}. Detailed surveys of background subtraction techniques and change detection algorithms have been published ~\cite{Piccardi04,Rich05}.

A typical motion-based foreground detection system would comprise of an estimate of the background (\textit{background model}), with the moving-foreground objects determined by comparing the current frame to the background model. An ideal system should be able to segment moving-foreground objects in the presence of noise, camera jitter, aperture effects, etc; it should also be resilient to gradual or sudden changes in illumination and new objects settling in the background  i.e., the background model should be temporally adaptive; and it must operate in real-time without being resource intensive.

A non-parametric approach, using kernel density estimation for representing the background was proposed  by Elgammal et. al~\cite{elga}. Without assuming any underlying distribution, a density function for pixel intensity was determined directly from the data. There exists a large volume of literature for background modelling based on quantisation/clustering~\cite{kim}, auto-regression~\cite{toya,monn}, Hidden Markov Model (HMM)~\cite{kato}, active contours~\cite{kass,cohe,yoko} and geodesic active contours~\cite{case,para}. All these methods are computationally expensive with large memory requirements.

 A mixture-of-Gaussians method to model pixel behaviour and update the background model using an exact EM algorithm (online K-means approximation) was proposed by Stauffer et. al ~\cite{gmm}. This approach is commonly used and many researchers have proposed improvements and extensions to this algorithm. Researchers\cite{kaew} have also proposed a new update algorithm for the mixture-of-Gaussians approach using a combination of an online EM algorithm and a $L$-recent window technique.  An adaptive version that can constantly update the number of mixtures was also proposed \cite{zivk}. The mixture-of-Gaussians approach is based on the assumption that the underlying distribution of pixel intensities is Gaussian. The above methods rely on adhoc parameters such as learning rate for recursive linear filters, weights of background states and new observations. Tuning these parameters in order to achieve high performance is often difficult. Moreover, the computational complexity of these methods is very high. 

  A simple and effective technique for background modelling based on the \textit{sigma-delta} estimation technique was proposed in ~\cite{sigdel}. Sigma-delta estimation is predominantly used as an analog-to-digital converter in the audio signal processing domain. Very low computation cost along with robustness of the non-linearity over the linear recursive average are some of the attractive features of this methodology.  

Researchers in ~\cite{maso}, explored techniques for foreground detection by modelling the background using color and/or edge histograms of small image blocks/cells. Every $24$-bit color pixel was transformed into $12$-bit color pixel using a color depth reduction formula. Color histograms were computed on these transformed color pixels. In order to compute edge histogram, a bin index value is computed for each pixel-based edge using edge orientation while the bin index is incremented using the edge magnitudes. The edge histogram-based technique was found to perform better than the color histogram-based technique. Researchers~\cite{lbp} have proposed a texture-based method for modelling the background and detecting moving objects. They use a mixture of local binary pattern (LBP) histograms to generate a background model. The method is computationally less intense but parameters have to be tuned depending on the video, which is nontrivial.

From the literature it can be gathered that most frameworks use single-pixel based~\cite{gmm,sigdel,elga,kato} or block-based (group of pixels)~\cite{lbp,maso,jabr} techniques. We intend to work towards a block-based technique.

Foreground detection is a mature area of research, however, most of the existing methodologies work under the assumption that the video frames are de-noised and of good quality and high frame rate. Existing techniques require numerous parameters to be tuned depending on the video.  Modern  surveillance systems use numerous cameras and the video frames are of low resolution and noisy. Tuning parameters for numerous cameras or scenes individually is an arduous process. So systems need to be  illumination-invariant and resilient to noisy, low frame rate  and low resolution frames  while requiring minimal calibration. 
  
In this paper we present a methodology that uses texture representation~\cite{pranam10} to describe a video frame and a filter operator which updates a background model and then determines the moving objects in the frame. The texture representation is computationally efficient and the texture measures are computed over a larger area than a single pixel. The proposed framework addresses the problem of detecting moving foreground objects in the presence of noise, illumination changes, dynamic background and low resolution objects, and requires few generic parameters to be set. The methodology, experimental setup and analysis is presented in sections~\ref{sec:propmethod}, ~\ref{sec:exp_setup} and~\ref{sec:analy}, respectively. In section~\ref{sec:appln} we present an example application of the proposed moving-object detector for vehicular traffic density estimation including comparison with other foreground detector-based vehicular traffic density estimation approaches.

%
%

%
\section{Proposed Methodology}
\label{sec:propmethod}


The proposed methodology is a block-based approach for detecting moving objects. Every grey-scale frame of the video is divided into blocks\footnote{overlapping or non-overlapping} and the aim is to classify every block into one of two states: \textit{stationary} or \textit{moving}.

\subsection{Texture Representation}

\begin{figure}
\begin{center}
\includegraphics[width=1.3in]{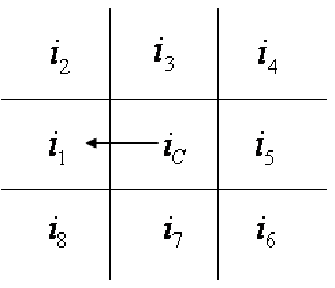}
\caption{A $3 \times 3$ neighbourhood of pixels}
\label{fig:3x3}
\end{center}
\end{figure}

We calculate the first-order finite directional differences in all directions within a $3 \times 3$ neighbourhood of pixels with  respect to the centre pixel. Directional differences for a centre pixel in a $3 \times 3$ neighbourhood (shown in Figure~\ref{fig:3x3}) can be written as, $\textbf{I}$, 
\begin{equation}
\label{eqn:centrepixel_eqn}
\textbf{I} = [i_{c}- i_{1},..,i_{c} - i_{8}], 
\end{equation}
where $i_{c}$ is the intensity of centre pixel $c$ and $i_{1,..,8}$ are the intensities of the neighbouring pixels.

Finite directional differences are approximately intensity invariant. 
Normalisation of the directional differences further enhances intensity invariance.
 
\begin{equation}
\textbf{I}_\mathrm{norm}=\frac{\textbf{I}}{\sqrt{\sum_{j=1}^{8} \textbf{I}_{j}^2}},
\label{eqn:norm_chap4}
\end{equation}

The Discrete Wavelet Transform (DWT) of the signal $\textbf{I}_\mathrm{norm}$, in the Haar basis, is calculated by passing it through a series of 
filters~\cite{Mallat} whose coefficients are given in Equation~\ref{eqn:haar}.
 In this work, Haar wavelets are used because they are computationally efficient and have the smallest possible support.
\begin{equation}
\textbf h = \left[\frac{1}{\sqrt{2}},\frac{-1}{\sqrt{2}}\right],\,
\textbf g = \left[\frac{1}{\sqrt{2}},\frac{1}{\sqrt{2}}\right]
\label{eqn:haar}
\end{equation}

The signal is decomposed simultaneously a using high-pass filter, $\textbf h$, and a low-pass filter, $\textbf g$. The outputs of the high-pass filter are known as the detail coefficients and those from the low-pass filter are referred to as the approximation coefficients. The filter outputs are then downsampled by $2$. The $n^{\text{th}}$ component of downsampling a vector $\textbf y$ by $k$ may be written as: 
\begin{equation}
(\textbf y\downarrow k)[n] = \textbf {y}[k n]
\label{eqn:downsample}
\end{equation}
where $\downarrow$ is used to denote the downsampling operator. Noting that the wavelet transform operation corresponds to a convolution followed by downsampling by $2$ allows the 
 filter outputs to be written more concisely as,
\begin{equation}
\textbf{G}=(\textbf{I}_\mathrm{norm}\ast \textbf g) \downarrow 2 ,\ ,
\textbf{H}=(\textbf{I}_\mathrm{norm}\ast \textbf h) \downarrow 2
\label{eqn:filter_out}
\end{equation}
where, $\textbf{G}$ and $\textbf{H}$ are the approximation and detail coefficients respectively.

The texture representation, $Z$, of a $3\times 3$ neighbourhood of pixels is given by Equation~\ref{eqn:tex_represent}.
\begin{equation}
Z = [\textbf{G},E_{\textbf{H}} ],
\label{eqn:tex_represent}
\end{equation}

where

\begin{equation}
E_{\textbf{H}} = \sum_{j=1}^{4}{\textbf{H}_{j}^2}
\label{eqn:H_energy}
\end{equation}
The detail coefficients would consist of high frequency components including associated noise. We represent the detail coefficients, $\textbf H$, by its collective energy, $E_{\textbf H}$, given by Equation~\ref{eqn:H_energy}, in order to retain the influence of  high frequency components whilst minimising the effect of noisy-coefficients, thereby reducing noise.


Illumination will essentially be uniform (except in pathological cases) over a \textit{small} neighbourhood, such as the $3\times3$ neighbourhood shown in Figure~\ref{fig:3x3}.
Distributions collected over a $m\times n$ block of pixels, consisting of numerous $3\times3$ neighbourhoods, will provide texture representations with substantial illumination-invariant property.

Consider a frame $f$, divided into $p \times q$ blocks with each block consisting of $m \times n$ pixels, then the texture representation of $(p,q)^{th}$ block, $\textbf{T}_f(p,q)$, from here on referred to as \textit{block-textures}, is given by,   
\begin{equation}
\label{eqn:pixel_eqn}
\textbf{T}_{f}(p,q)= \frac{1}{mn} \sum_{x=1}^{m} \sum_{y=1}^{n}Z_{x,y}, 
\end{equation}
where $Z$ is  texture representation with $d$ dimensions. 
For a frame with $N$ pixels, texture representation takes $O(8N)=O(N)$ computations. This texture representation effectively captures local variations in pixel intensities and is also inherently illumination-invariant, noise-resilient per block and computationally inexpensive.

\subsection{Modelling and Detection}
 The first frame is always the initial seed for modelling and is assumed to be a \textit{stationary model} ($\textbf{M}$) of block-textures which is updated using the current matrix of block-textures, $\textbf{T}$. The stationary model is estimated on-the-go, thus  there is no explicit training phase involved.
The sum of absolute differences between the stationary model, $\textbf{M}$,  and the current matrix of block-textures, $\textbf{T}$,  provides the \textit{motion-likelihood} ($\Delta$) measure. Comparing the motion-likelihood measure to a derived tolerance value, $\tau$, we generate a motion map, $\mathbf{\phi}$, depicting \textit{moving} blocks. 

Pseudo-code for initialisation, update and detection is provided in Algorithm \ref{alg:update}. The function $Sgn$ is defined as $Sgn(a) = [sgn(a_{1}),..,sgn(a_{d})]$, where, $a$ is a vector with $d$ dimensions and  $sgn(b)= 1$ if $b > 0$, $sgn(b)=-1$ if $b<0$, and $sgn(b)=0$ if $b=0$.

The stationary model is updated at the learning rate ($\pm\alpha$). By using increments $\pm\alpha$ for the learning rate, we avoid using any Gaussian fitting techniques.  This comparison and elementary increment/decrement can be interpreted as the simulation of a digital conversion of a time-varying analog signal using sigma-delta modulation, except that the increments are $\pm\alpha$, with $0 \leq \alpha \leq 1$. Intuitively, using $\alpha = 0$, would mean that \textit{no} information should be learned and using $\alpha=1$, would mean \textit{all} information should be learned. Using a low value for $\alpha$, for example $\alpha < 0.2$, would mean only \textit{some} information should be learned. An interesting property of this technique is that the error is proportional to the variation rate of the input signal corresponding to a motion-likelihood  measure  of the block-textures. We compare this motion-likelihood  measure to a tolerance to classify the block as \textit{stationary} or \textit{moving}.

The limit of tolerance, $\tau$, will eventually determine the number of detections made by the system. The texture measure in each dimension can only vary between $\frac{-1}{\sqrt 2}$ and $\frac{+1}{\sqrt 2}$, thus the motion-likelihood lies in $[0, \frac{5}{\sqrt 2}]$. In order for an user to be able to vary the limit of tolerance, $\tau$, we introduce a tradeoff parameter, $\lambda$, so that the limit of tolerance, $\tau$, will be proportional to the user-defined tradeoff, $\lambda$: 


\begin{equation}
\label{eqn:tolerance}
\tau = (1-\lambda)\frac{5}{\sqrt 2}
\end{equation}   
where, $0 \leq \lambda \leq 1$. $\lambda=0$ would result in the system showing \textit{no} detections and $\lambda=1$ would result in system showing \textit{all} the detections.


\begin{algorithm}
\emph{Initialisation}\;
\SetKwInOut{Input}{input}
\SetKwInOut{Output}{output}
\Input{$\textbf{T}_{0}$: block-textures of first frame ($f = 0$),\\ $p,q$ : block size}
\Output{$\textbf{M}_{0}$: Stationary model at first frame}
\BlankLine
\ForEach{$k \leftarrow 1$ \KwTo $p$}{
\ForEach{$l\leftarrow 1$ \KwTo $q$}{\nllabel{forins}
$\textbf{M}_0(k,l) \gets \textbf{T}_{0}(k,l)$\;
}
}
\hbox{}
\emph{Update and Detect}\;
\SetKwInOut{Input}{input}
\SetKwInOut{Output}{output}
\caption{Initialise, Update and Detect}
\label{alg:update}
\Input{$\textbf{T}_{f}$: block-textures of frame $f$ ($f \neq 0$),\\ $p,q$: block size,\\ $\alpha$ : learning rate,\\ $\tau$ : limit of tolerance}
\Output{$\textbf{M}_{f}$: Stationary model at frame $f$,\\ $\mathbf{\phi}_{f}$ : Motion Map at frame $f$}
\BlankLine
\ForEach{$k \leftarrow 1$ \KwTo $p$}{
\ForEach{$l\leftarrow 1$ \KwTo $q$}{
\emph{Update stationary model}\;
$\textbf{M}_{f}(k,l) \gets \textbf{M}_{f-1}(k,l) + \alpha *Sgn(\textbf{T}_{f}(k,l) - \textbf{M}_{f-1}(k,l))$\;
\emph{Calculate motion-likelihood measure}\;
$\Delta \gets  ||\textbf{M}_{f-1}(k,l) - \textbf{T}_{f}(k,l)||_{1}$\;
\emph{Generate motion-map}\;
\eIf{$\Delta < \tau$} {
$\mathbf{\phi}_{f}(k,l) \gets 0$\; }{
$\mathbf{\phi}_{f}(k,l) \gets 1$\; }
}
}
\end{algorithm}

%
\section{Experimental Setup}
\label{sec:exp_setup}


Six different (indoor and outdoor) video sequences were used for performance evaluation. Table~\ref{tab:videoproperties} shows the properties of all video sequences used in this experimental setup. The \textit{Afternoon} and \textit{Morning} traffic video sequences captured by traffic cameras installed on Sydney roads are a proprietary dataset of Roads and Traffic Authority of New South Wales, Australia. \textit{CAVIAR}, \textit{Multiple\_Flows} and  \textit{WavingTrees}  are publicly available on the Internet(\cite{inria}, \cite{pets} and \cite{wavingtrees} respectively). \textit{Hall\_Monitor} is a well-known sequence used for MPEG4 standards. 

The proposed methodology was compared with three state-of-the-art, widely used methods that are considered as the most computationally efficient techniques (GMM~\cite{gmm,Zoran06}, Sigma-Delta~\cite{sigdel} and LBP-based~\cite{lbp}).

In order to showcase the \textit{non-dependency} or \textit{dependency} of a methodology on fine-tuning of its parameters, we fine-tune the parameters of a methodology using an input  video ``A" and then test the methodology (retaining the previously fine-tuned parameters) for its performance using an input  video ``B". Here, the assumption is that video ``A" is \textit{unrelated} (w.r.t. semantics) to video ``B" in terms of scene-type, etc. If a methodology can attain maximum performance for input video ``B" then it is \textit{independent} of parameter fine-tuning, otherwise it is \textit{dependent} on parameter fine-tuning. We believe that a system that is \textit{independent} of parameter fine-tuning is much more desirable for real world applications compared to a system that is \textit{dependent} on parameter fine-tuning. We fine-tune the parameters for all algorithms using the \textit{Multiple\_Flows} sequences. 

RGB video frames are used in the evaluation of the GMM-based technique~\cite{gmm,Zoran06}, with three Gaussian distributions per pixel per color component. Learning rate and foreground threshold are  set to $0.01$ and $0.25$, respectively. A match is defined as a pixel value being within $2$ standard deviations of the distribution's mean. We have used the GMM implementation provided by Zivkovi et al.~\cite{zoranGMM}.  Grayscale video frames are used in the evaluation of Sigma-Delta based technique~\cite{sigdel}. Temporal pixels are identified as the pixels whose variation rate is $N =4$ times the non-zero differences. Grayscale video frames are also used in evaluation of LBP based technique~\cite{lbp} using three histogram mixtures. The rate of learning for the histogram mixture and weights were both set to $0.005$. The pixel block-size was set to $8\times8$ pixels, with a detection threshold of $0.25$. Heikkil\"{a} and Pietik\"{a}inen~\cite{lbp} use overlapping blocks in order to achieve better shape contour; we set the overlapping region as half the block-size.   

Results for the proposed methodology are reported for the overlapping and non-overlapping block-based techniques. The proposed methodology requires only three parameters to be pre-set: block-size,  learning-rate($\alpha$) and tradeoff($\lambda$). There are no other explicit system parameter that need to be set. Parameter setting for the proposed methodology is presented in Table~\ref{tab:para}.  Though increasing the block-size will reduce the computational load, we have instead used a small block-size in order to achieve good shape-contours for moving objects. Increasing the learning rate will result in the model following the input closely, leading to very few detections.  Decreasing the learning rate significantly results in a high number of false detections. Moving objects are detected by thresholding the motion-likelihood measure by the limit of tolerance, derived using the tradeoff as given by Equation~\ref{eqn:tolerance}. Decreasing the tradeoff reduces the number of correct detections and vice-versa. We have set a low learning rate and tradeoff of $0.05$ and $0.98$, respectively.

We claim here that the proposed technique does not need parameter-tuning for individual videos. Hence, irrespective of the input video, we retain the same parameter settings for all techniques used in our experiments.

\begin{table}
\begin{center}
\caption{Parameters used for conducting experiments using proposed methodology}
\label{tab:para}
\begin{tabular}{|c|c|c|c|}\hline
 & Block size (pixels)  & $\alpha$ & $\lambda$ \\\hline
Overlap & $8\times8$   & $0.05$ & $0.98$ \\\hline
Non-overlap & $12\times 12$   &$0.05$ & $0.98$ \\\hline
\end{tabular} 
\end{center} 
\end{table}

Quantitative and qualitative results are presented for these experiments. For quantitative evaluation, we manually labelled four frames from the \textit{Afternoon}, \textit{Morning}, \textit{Hall\_Monitor} and \textit{CAVIAR} video sequences and four frames from each of the six mini-sequences of \textit{Multiple\_Flows}. Hand-segmented ground truth for \textit{WavingTrees} is available online~\cite{wavingtrees}. We then calculated the average scores for the true positives/negatives and false positives/negatives reported by each methodology.

We have quantified the noise\footnote{Noise-free video for the sequences used are not available hence the noise calculation is an approximate estimation. PSNR is calculated as a reference for supporting the claims made in this paper.} for each video  in the dataset (except \textit{WavingTrees}\footnote{The technique used for PSNR calculation in this work is not valid for videos with dynamic background.}) by calculating the peak signal-to-noise ratio (PSNR) in decibels (dB), so as to gauge the performance of all algorithms for different PSNRs. The average signal-to-noise ratio  for all pixel-locations over the entire collection of background frames was considered as an approximate quantification of noise for the video. Consider a video with $n$ frames consisting of a stationary background with $r \times c$ frame size. We have $p_{i,j}$ as a collection of all pixel intensities from location $(i,j)$  over $n$ frames, where $i \in \{1,..,r\}$  and $j \in \{1,..,c\}$. Thus, the signal-to-noise ratio for pixel 
location $(i,j)$ is given by Equation~\ref{eqn:snr}.

\begin{equation}
\text{SNR}_{i,j} = 20 \times \log_{10}(\frac{\max(p_{i,j})-\min(p_{i,j})}{\sigma_{p_{i,j}}})
\label{eqn:snr}
\end{equation}
where $\sigma_{p_{i,j}}$ is the standard deviation of the collection of pixel intensities, $p_{i,j}$.

The average signal-to-noise ratio over all pixel locations, as shown in Equation~\ref{eqn:psnr_avg}, is considered as the PSNR for the video.

\begin{equation}
\text{PSNR} = \frac{1}{rc}\sum_{i=1}^{r}\sum_{j=1}^{c}\text{SNR}_{i,j}
\label{eqn:psnr_avg}
\end{equation}

\begin{table*}
\begin{center}
\caption{Properties of all video sequences used in this experimental setup}
\label{tab:videoproperties}
\begin{tabular}{|>{\small}c|>{\small}c|>{\small}c|>{\small}c|>{\small}c|>{\small}c|>{\small}c|>{\small}c|>{\small}c|}\hline
\multirow{11}{*}{\rotatebox[]{90}{Properties}}
& \multicolumn{2}{>{\small}c|}{Name} & Afternoon & Morning & CAVIAR & Hall$\_$& Multiple$\_$ & WavingTrees\\
& \multicolumn{2}{>{\small}c|}{}& & & & Monitor& Flows & \\ \cline{2-9}
& \multicolumn{2}{>{\small}c|}{Location} & Outdoor & Outdoor& Indoor& Indoor& Outdoor& Outdoor\\ \cline{2-9}
& \multicolumn{2}{>{\small}c|}{Objects} & Vehicles & Vehicles& People & People & People & Person\\ \cline{2-9}
& \multicolumn{2}{>{\small}c|}{Resolution} & $320\times320$ & $320\times320$ & $384\times288$ & $352\times288$ & $768\times576$ & $160\times120$\\ \cline{2-9}
& \multicolumn{2}{>{\small}c|}{Num. of Frames} & $3000$ & $3000$ & $1050$ & $300$ & $874$ & $287$\\ \cline{2-9}
& \multicolumn{2}{>{\small}c|}{Light Source} & natural & natural & natural + artificial & artificial & natural & natural \\ \cline{2-9}
& \multirow{2}{*}{\small{Illumination Change}} & gradual & $\checkmark$ & $\checkmark$  & $\times$ & $\times$ & $\times$ & $\times$ \\ \cline{3-9}
&  & sudden & $\checkmark$ & $\checkmark$  & $\times$ & $\times$ & $\times$ & $\times$\\ \cline{3-9}
&  & no change & $\times$ & $\times$  & $\checkmark$ & $\checkmark$ & $\checkmark$ & $\checkmark$ \\ \cline{2-9}
& \small{Noise/Artifacts} & aperture & $\checkmark$ & $\checkmark$ & $\times$ & $\times$ & $\times$ & $\times$\\ \cline{3-9}
& & banding & $\checkmark$ & $\checkmark$ &  $\times$ & $\times$ & $\times$ & $\times$\\ \cline{3-9}
& & random/shot & $\checkmark$ & $\checkmark$ &  $\checkmark$ & $\checkmark$ & $\checkmark$ & $\checkmark$\\ \cline{3-9}
& & compression & $\checkmark$ & $\checkmark$ & $\checkmark$ & $\checkmark$ & $\times$ & $\checkmark$ \\ \cline{2-9}
& \multicolumn{2}{>{\small}c|}{Dynamic Background} & $\times$ & $\times$ & $\times$ & $\times$ & $\times$ & $\checkmark$ \\ \cline{2-9}
& \multicolumn{2}{>{\small}c|}{Camera on}  & $\checkmark$ &  $\checkmark$ & $\times$ & $\times$ & $\times$ & $\times$\\
& \multicolumn{2}{>{\small}c|}{Auto-Calibration mode}&&&&&&\\ \hline
\end{tabular}
\end{center}
\end{table*}

%
\section{Results and Analysis}
\label{sec:analy}


\subsection{Qualitative Analysis}

\begin{figure}[H]
\begin{center}
\subfigure[frames($f$): $797$, $832$, $889$]{
\includegraphics[width=20mm,height=20mm]{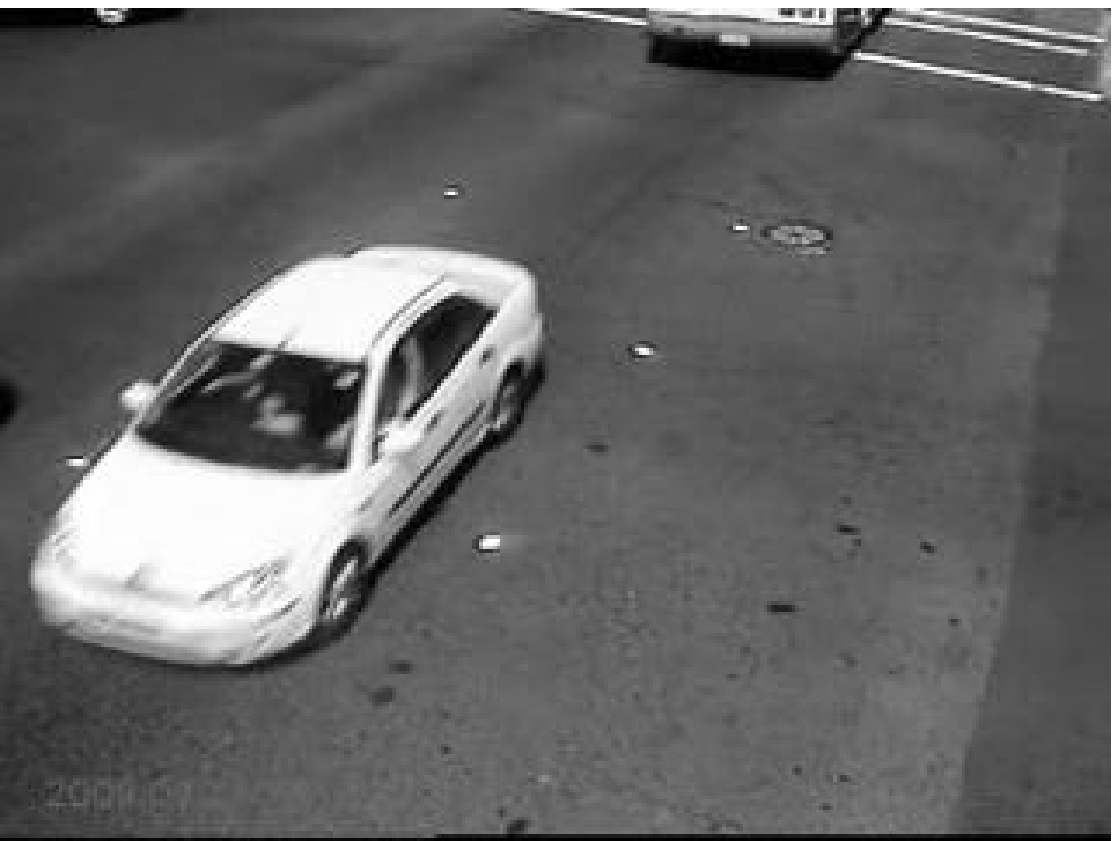}
\includegraphics[width=20mm,height=20mm]{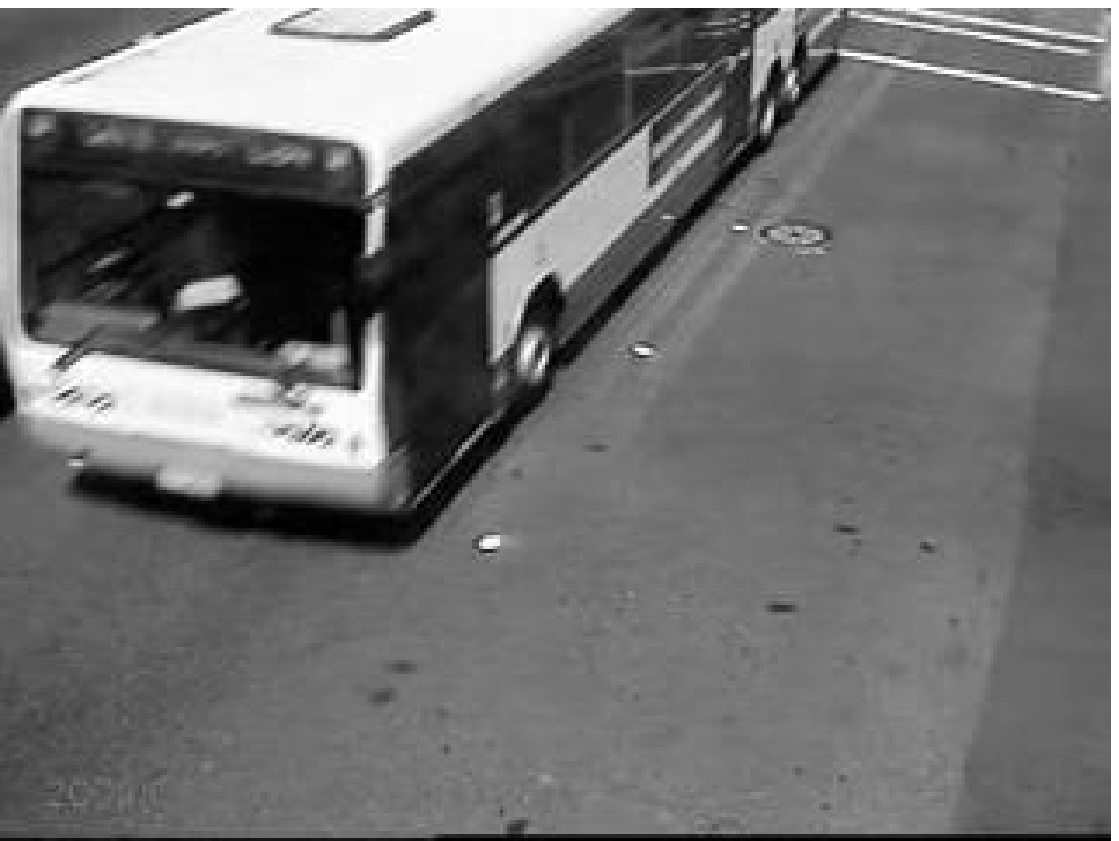}
\includegraphics[width=20mm,height=20mm]{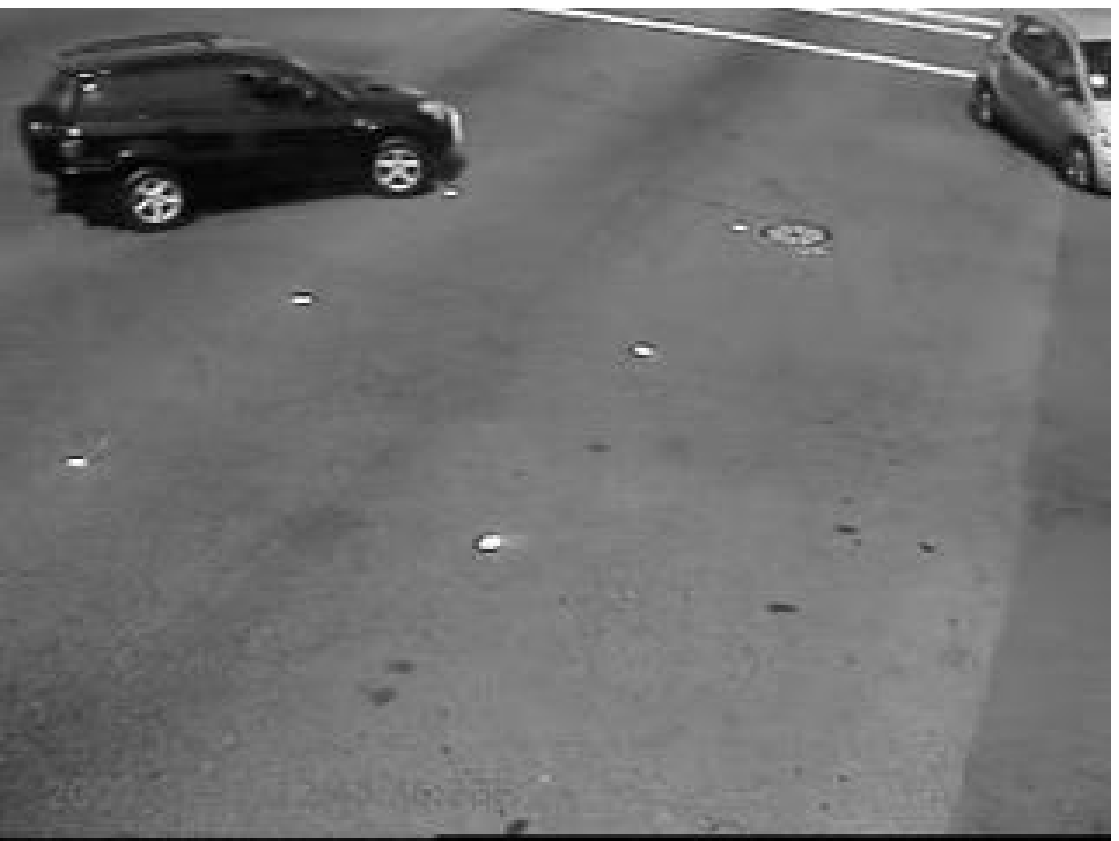}

}

\subfigure[Ground-truth]{
\includegraphics[width=20mm,height=20mm]{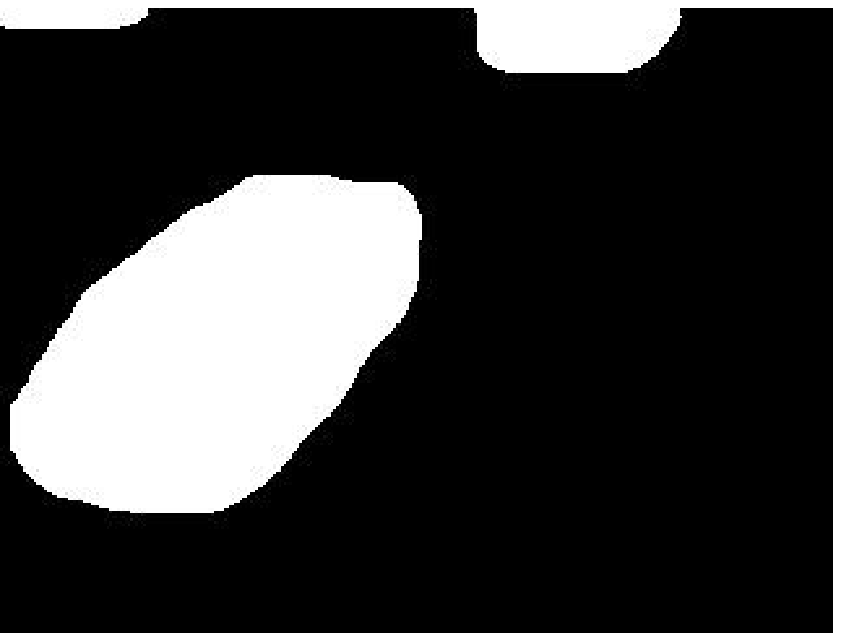}
\includegraphics[width=20mm,height=20mm]{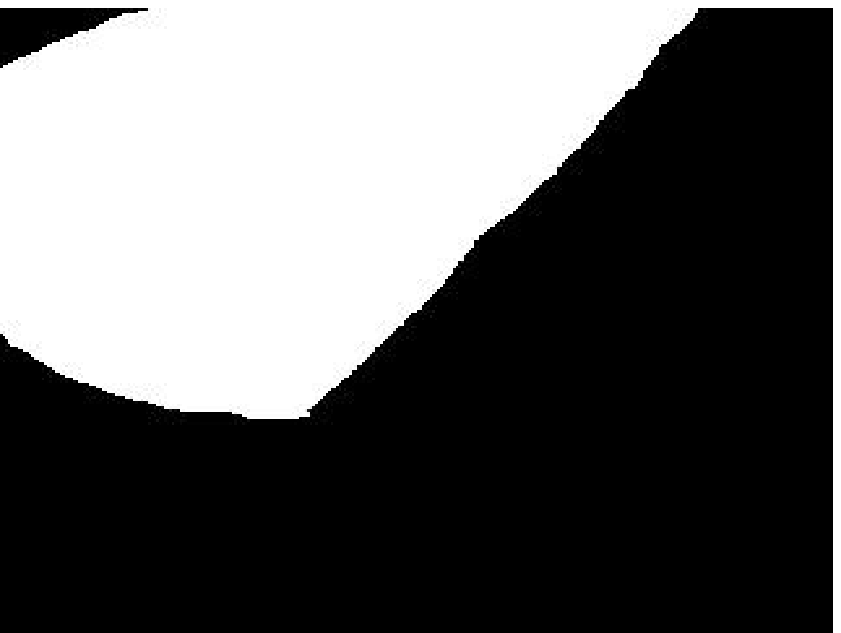}
\includegraphics[width=20mm,height=20mm]{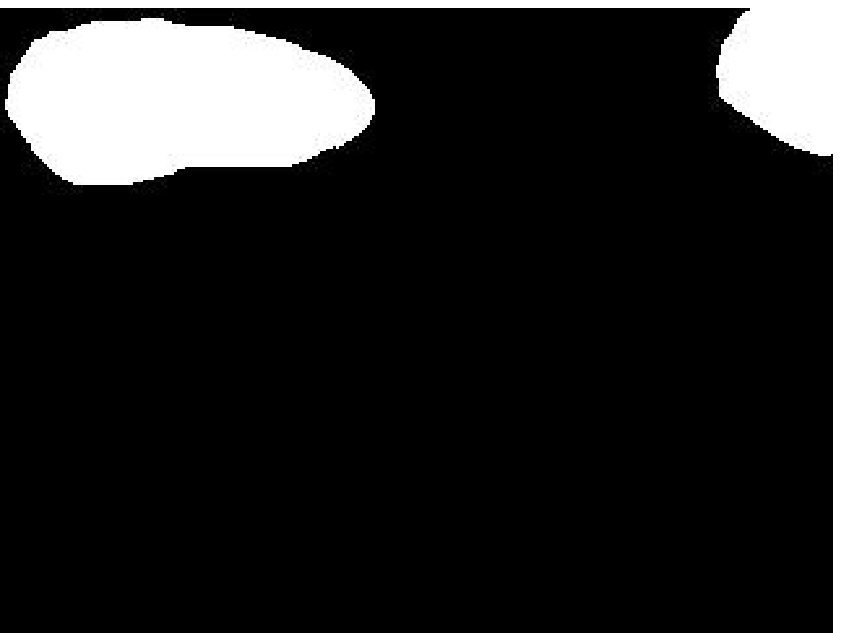}
}

\subfigure[Proposed Methodology]{
\includegraphics[width=20mm,height=20mm]{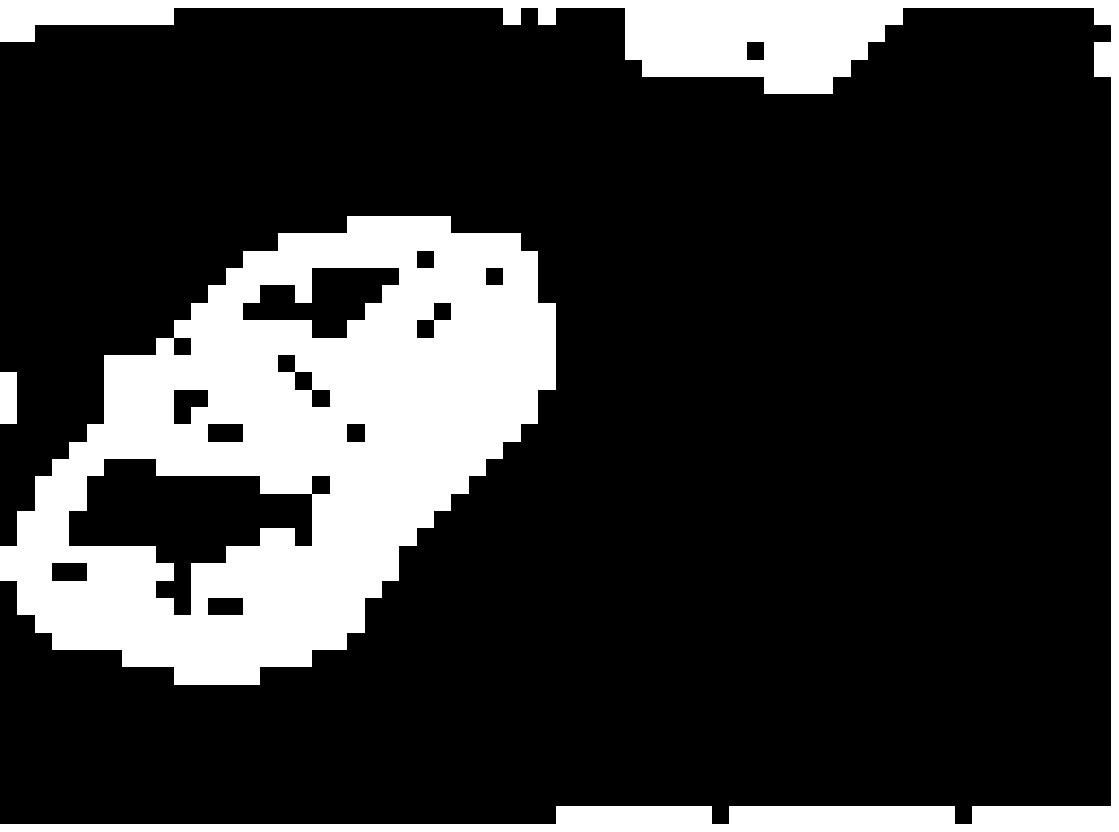}
\includegraphics[width=20mm,height=20mm]{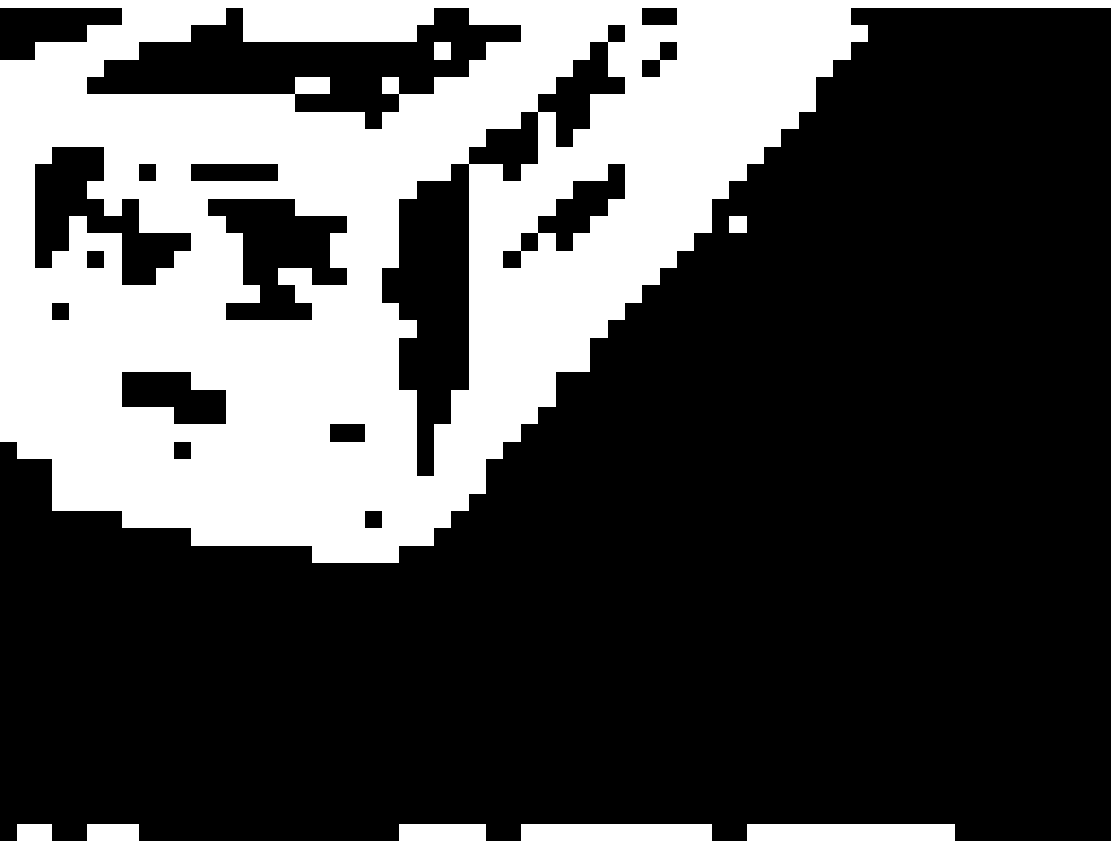}
\includegraphics[width=20mm,height=20mm]{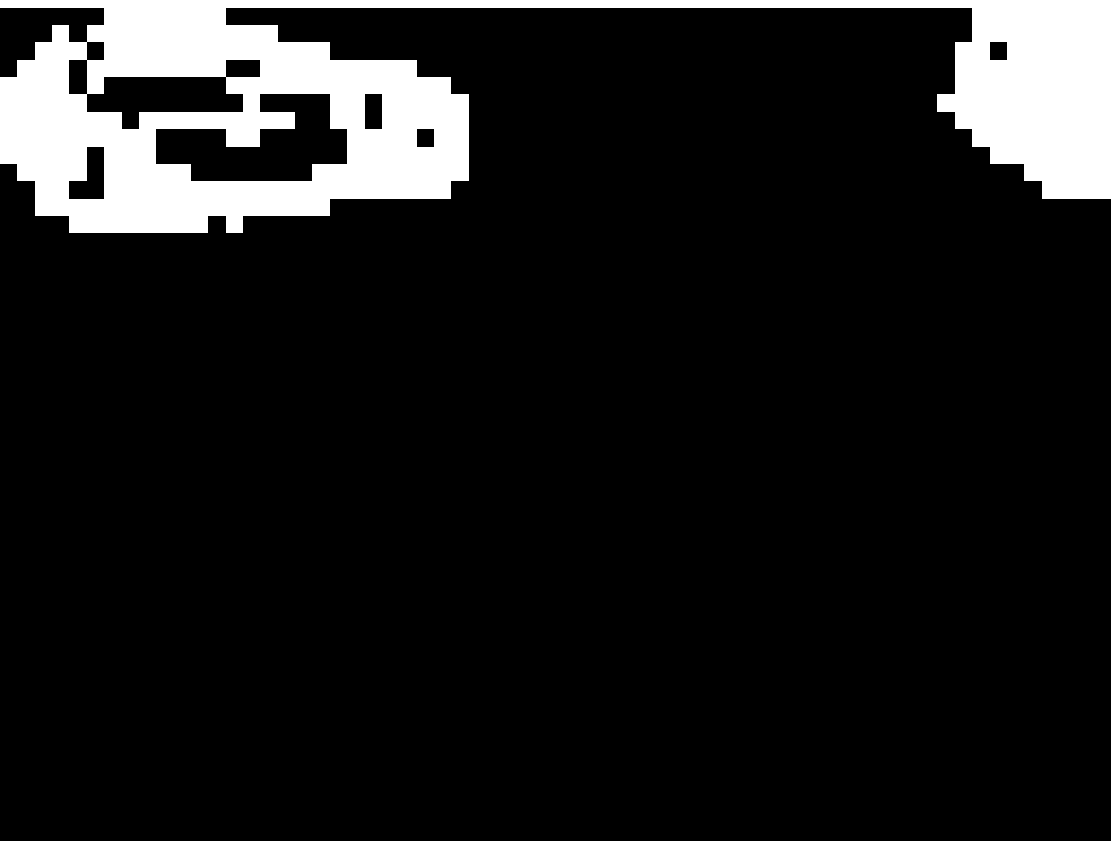}
}

\subfigure[GMM]{
\includegraphics[width=20mm,height=20mm]{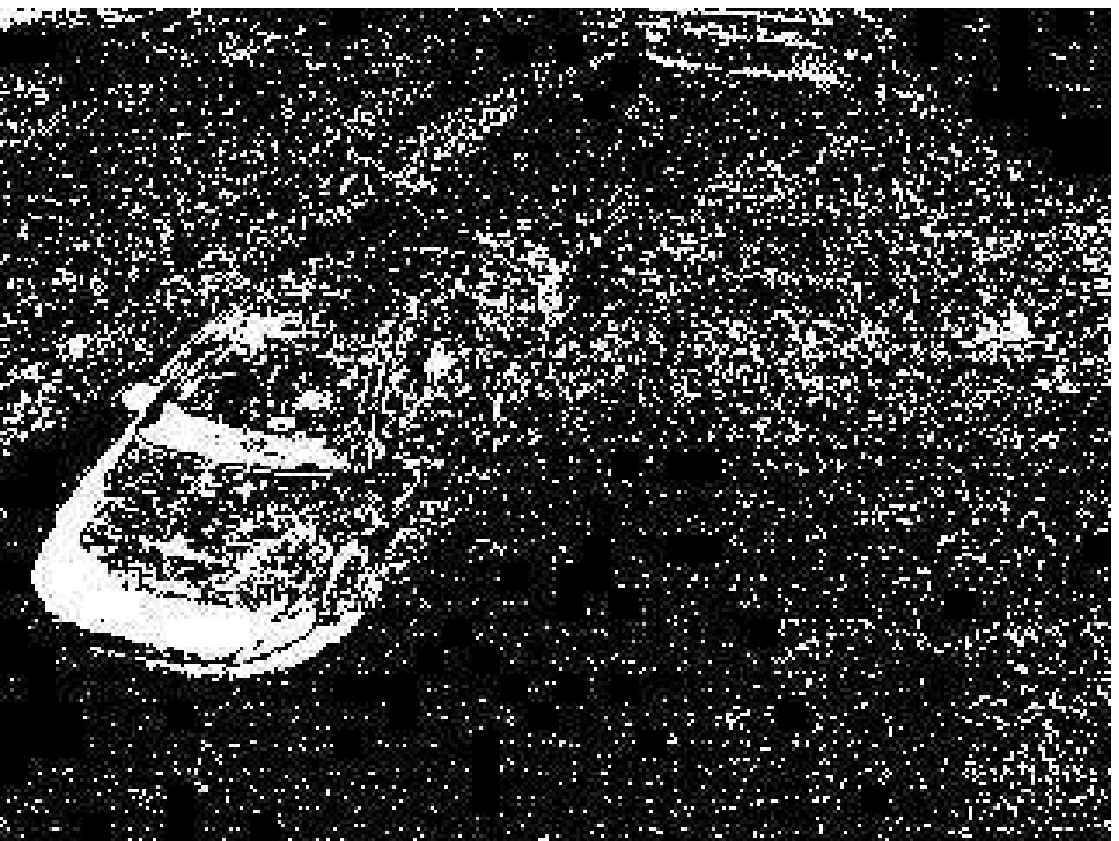}
\includegraphics[width=20mm,height=20mm]{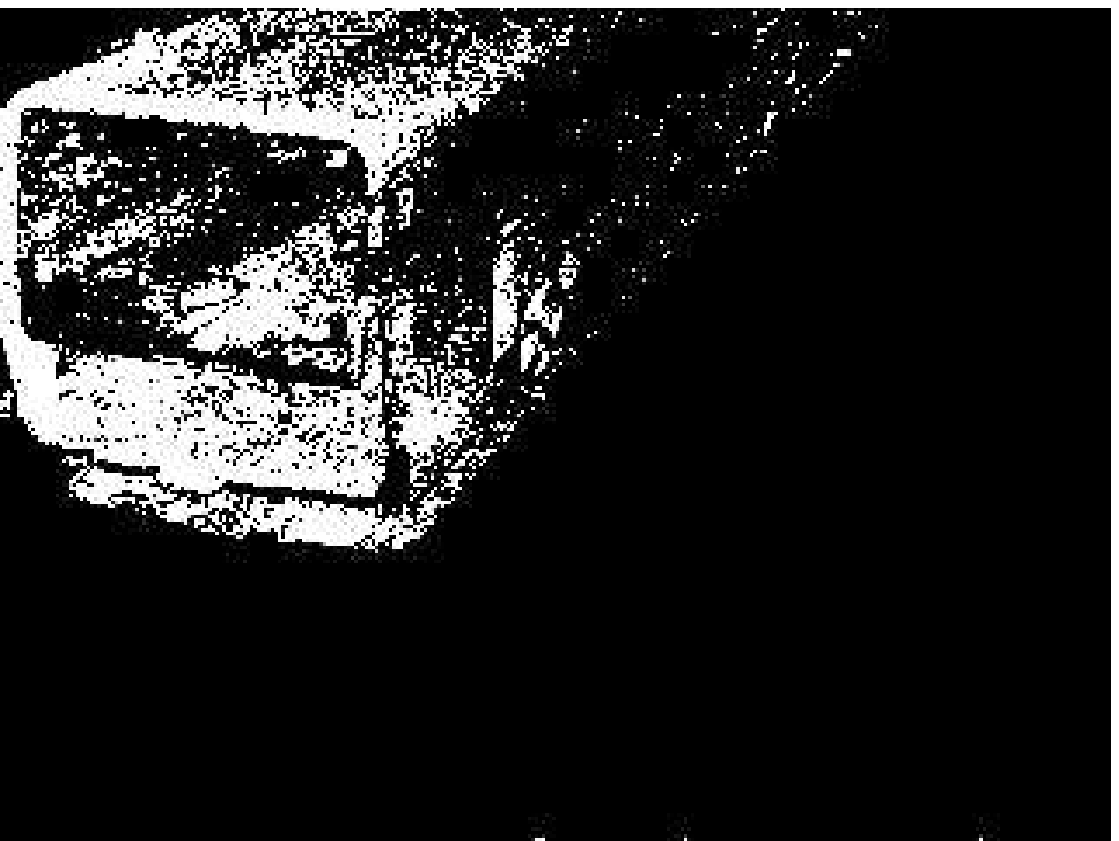}
\includegraphics[width=20mm,height=20mm]{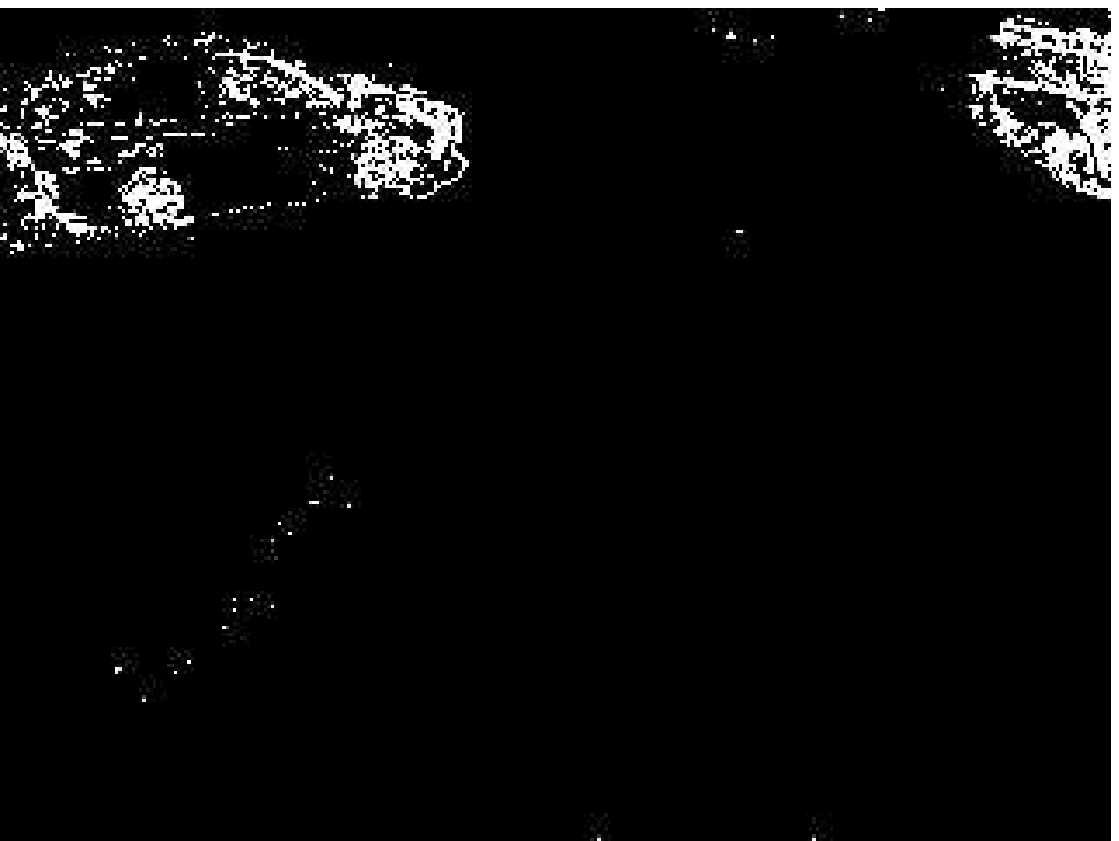}
}

\subfigure[Sigma-Delta]{
\includegraphics[width=20mm,height=20mm]{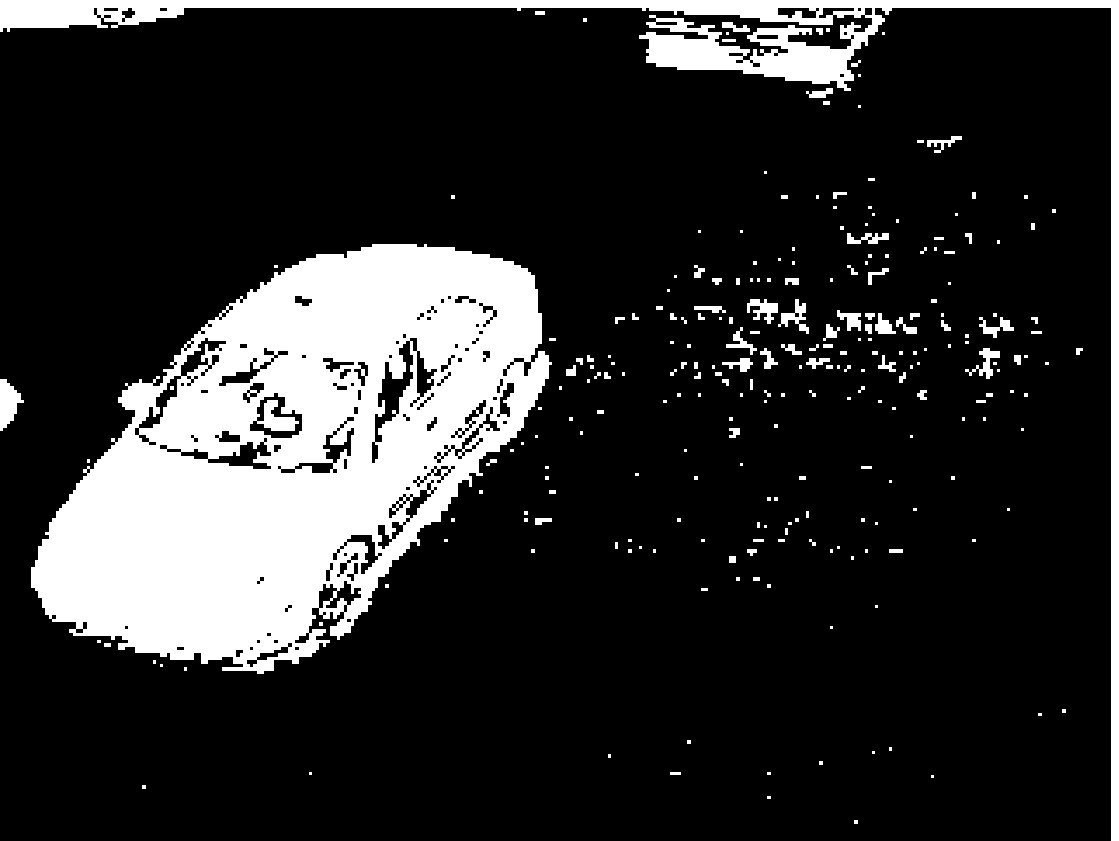}
\includegraphics[width=20mm,height=20mm]{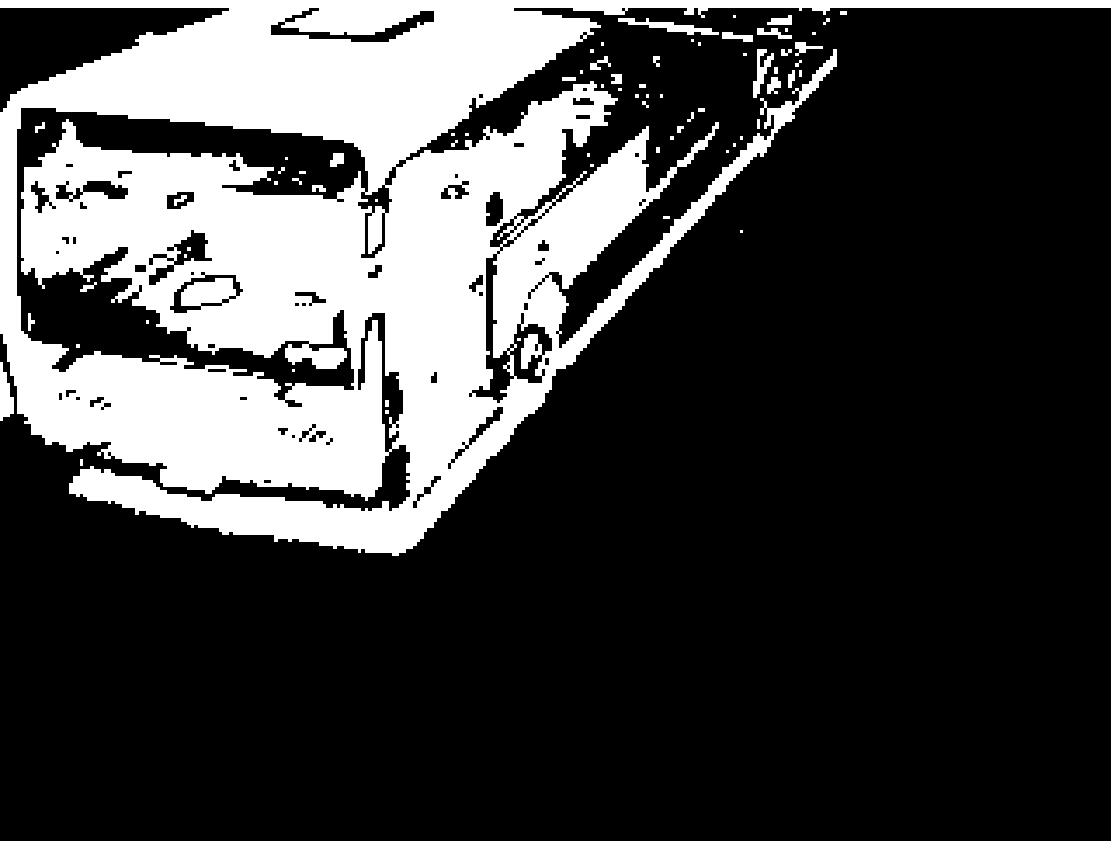}
\includegraphics[width=20mm,height=20mm]{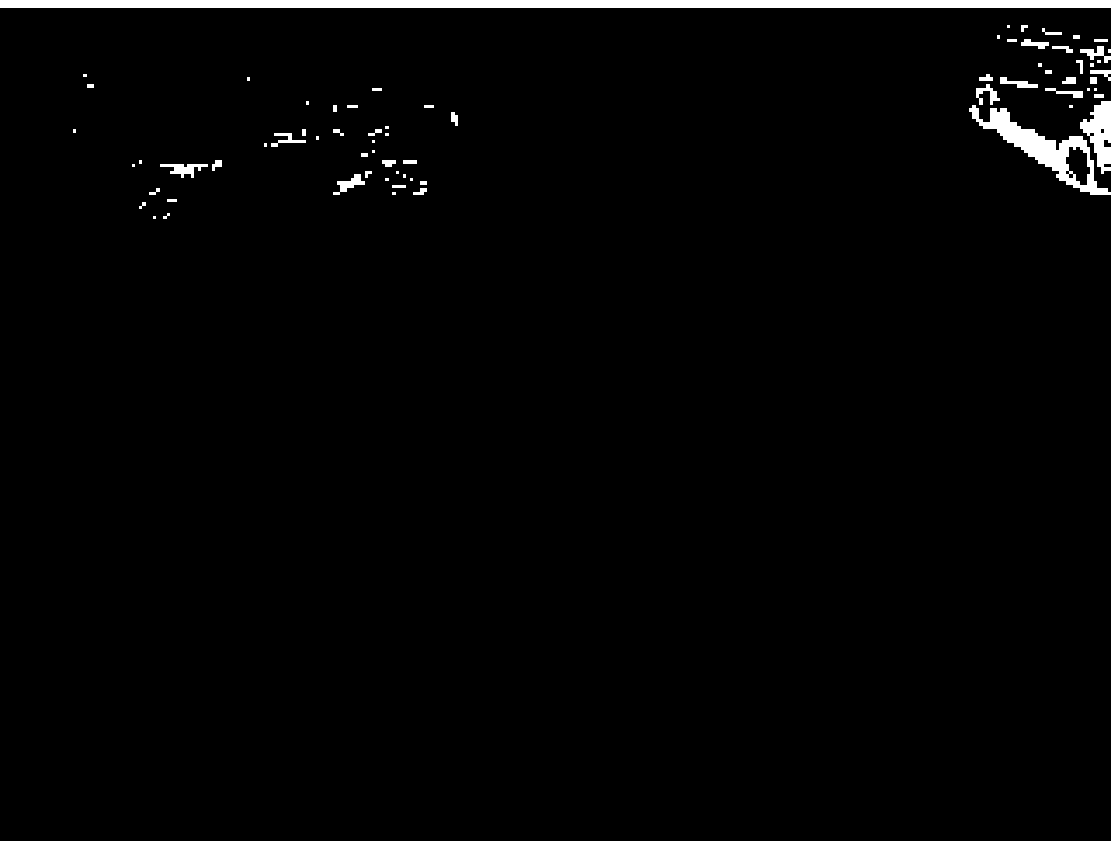}
}

\subfigure[LBP]{
\includegraphics[width=20mm,height=20mm]{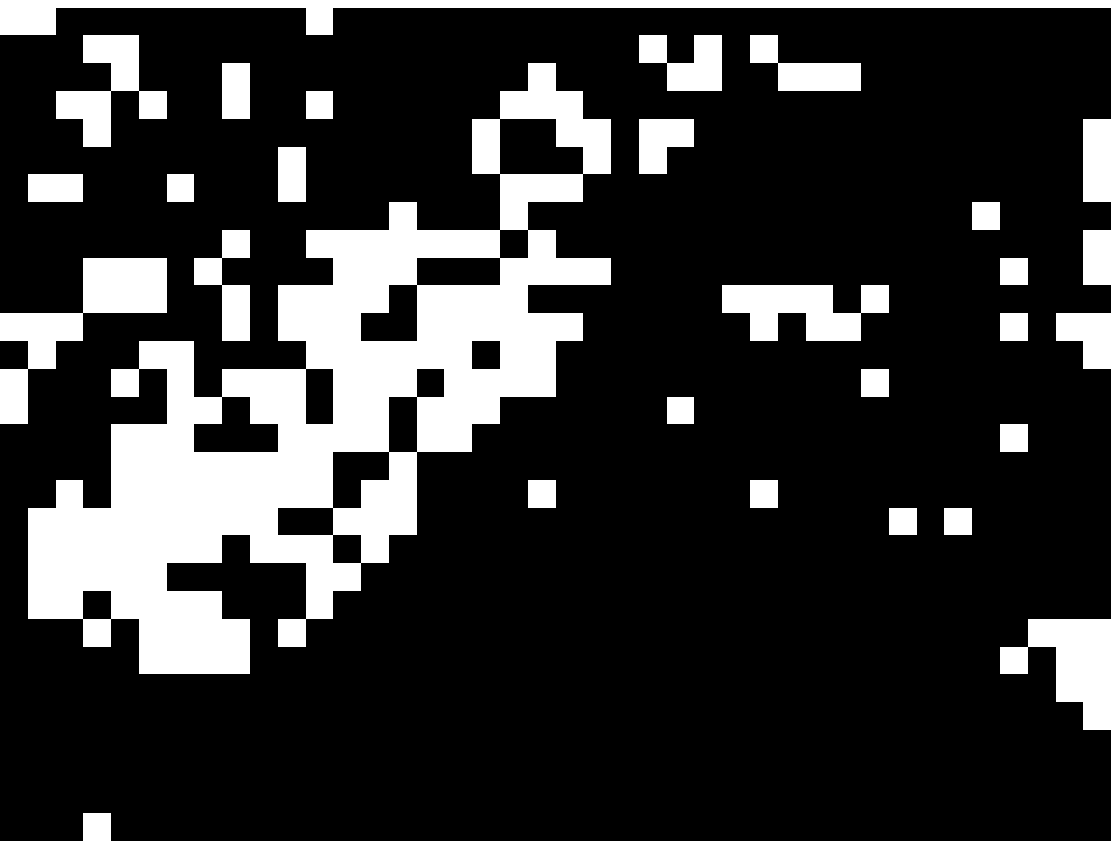}
\includegraphics[width=20mm,height=20mm]{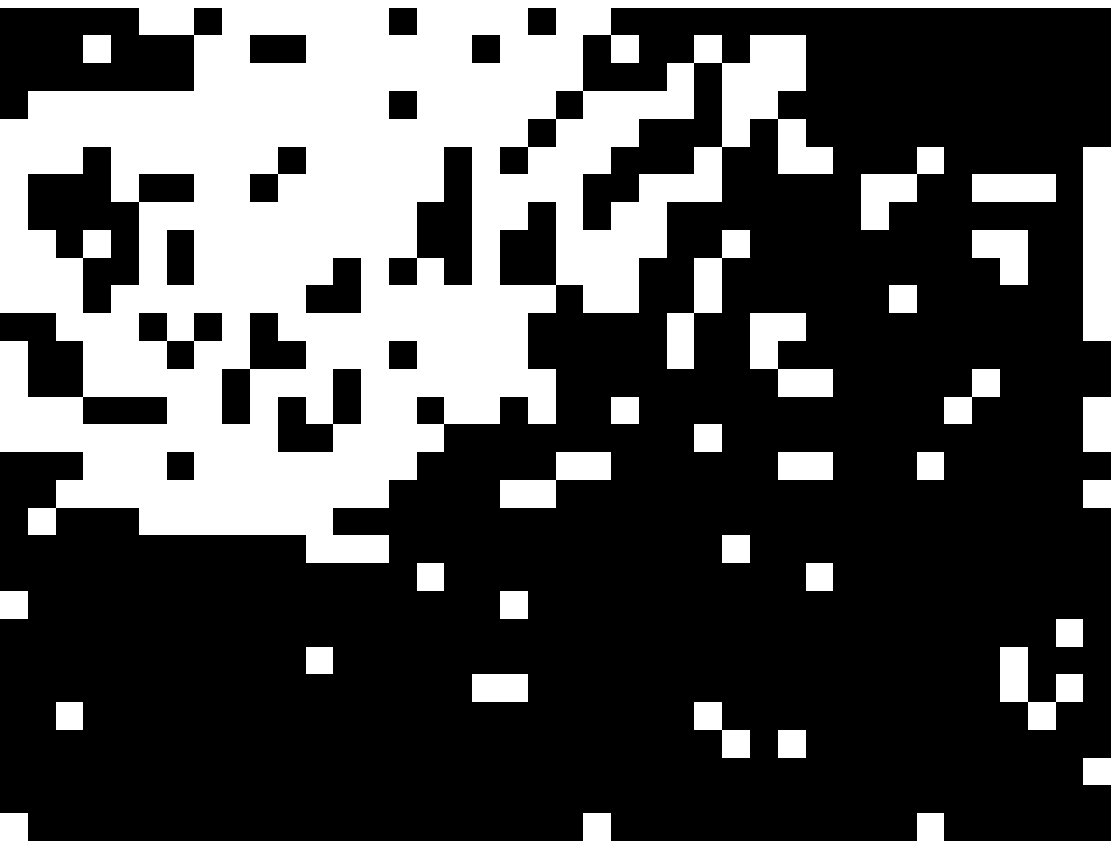}
\includegraphics[width=20mm,height=20mm]{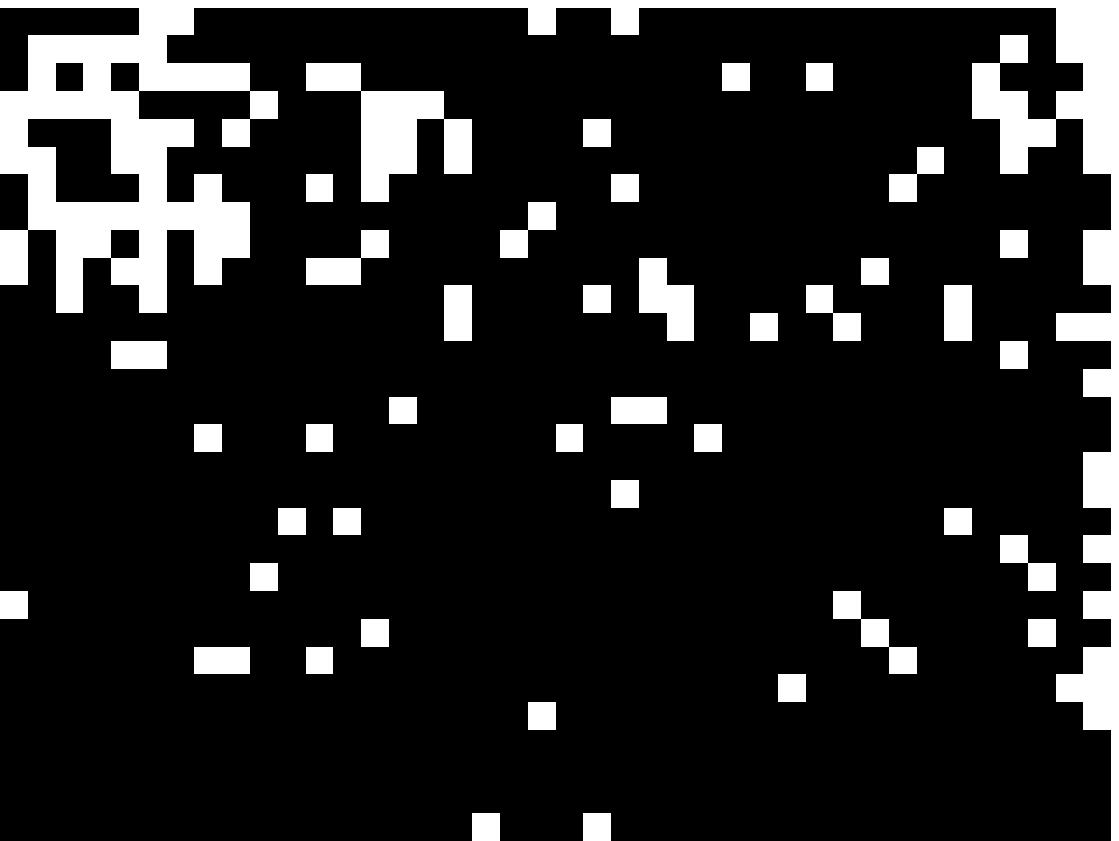}
}

\caption{Ground-truth along with results for all algorithms for a sequence of frames ($f:797,832$ and $889$) from \textit{Afternoon} video where the goal is to detect the moving vehicle that has entered the frame from behind the bus.}
\label{fig:car_miss}
\end{center}
\end{figure}
Figure~\ref{fig:car_miss} shows the results for a sequence of frames from the \textit{Afternoon} video for all algorithms. A car waiting to turn just after the bus has passed has been clearly detected using the proposed methodology but not by other techniques. GMM does have a few true positives, but these true positives are small in number and less dense. The Sigma-Delta approach fails to detect any movement and it is hard to differentiate between true detections and false detections in the case of the LBP-based methodology.  It is quite interesting to note that ``white'' vehicles reflect light onto the camera forcing the camera to adjust its colour settings as they are on auto-calibrate mode. Hence, the GMM-based technique  detects more false-positives even though it uses RGB data.

\begin{figure}[H]
\begin{center}
\subfigure[Aperture effects, Noise, Low-resolution human and Dynamic background]{
\includegraphics[width=20mm,height=20mm]{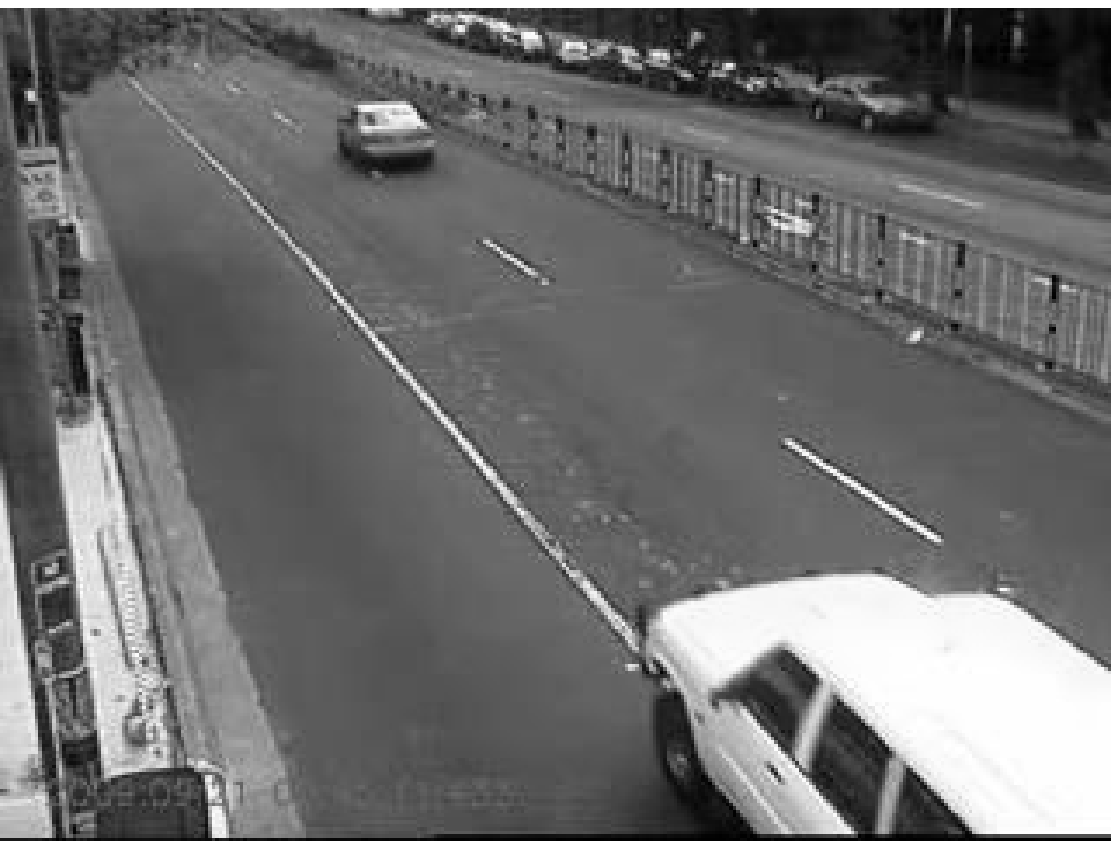}
\includegraphics[width=20mm,height=20mm]{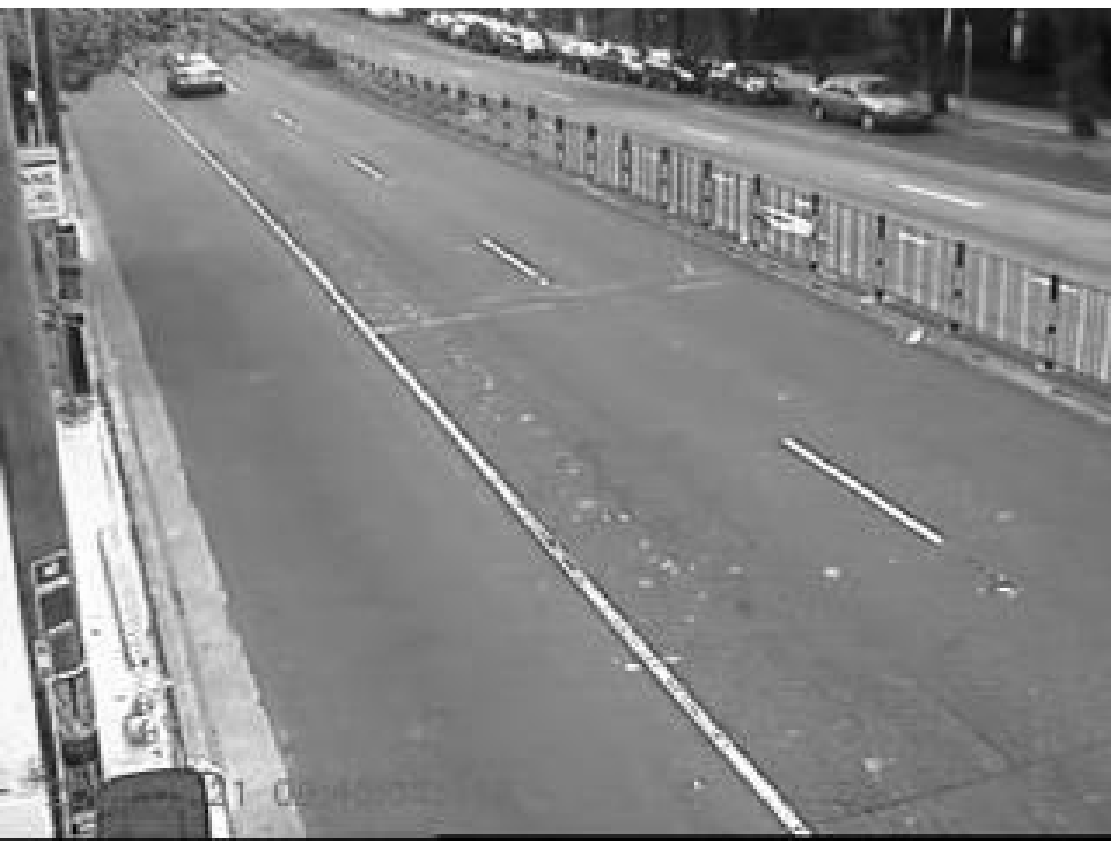}
\includegraphics[width=20mm,height=20mm]{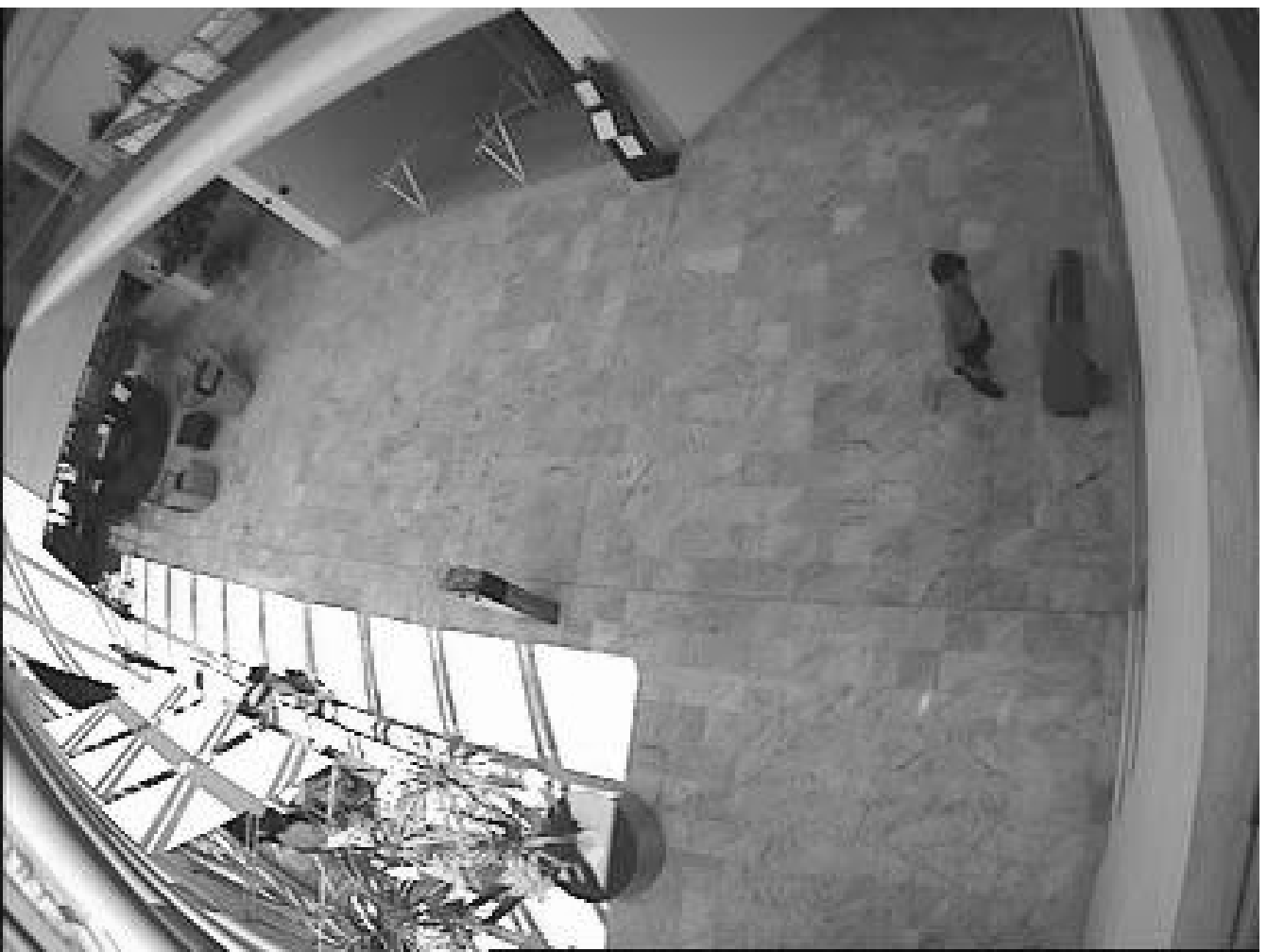}
\includegraphics[width=20mm,height=20mm]{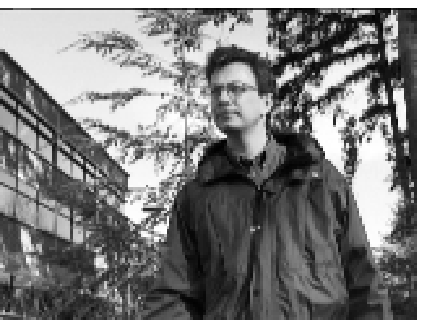}
}

\subfigure[Ground-truth]{
\includegraphics[width=20mm,height=20mm]{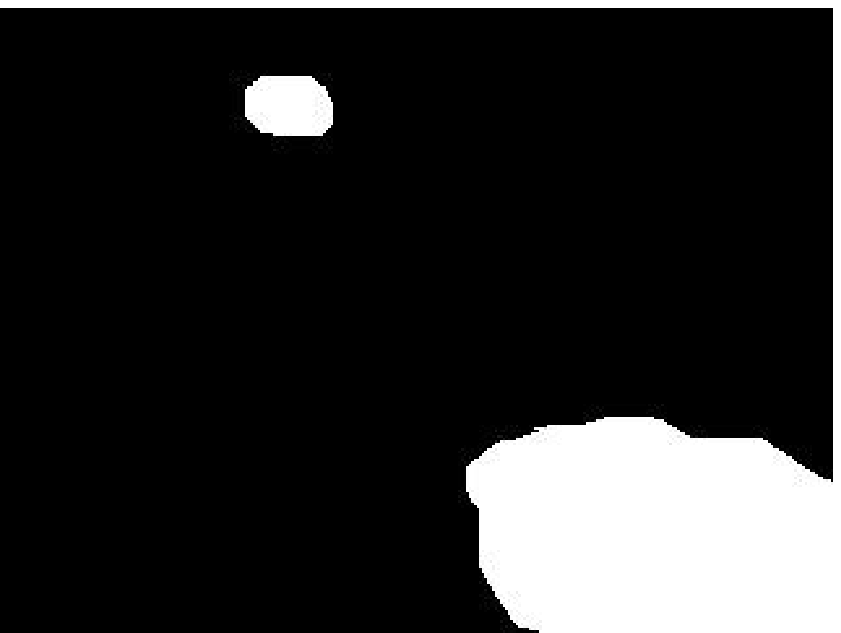}
\includegraphics[width=20mm,height=20mm]{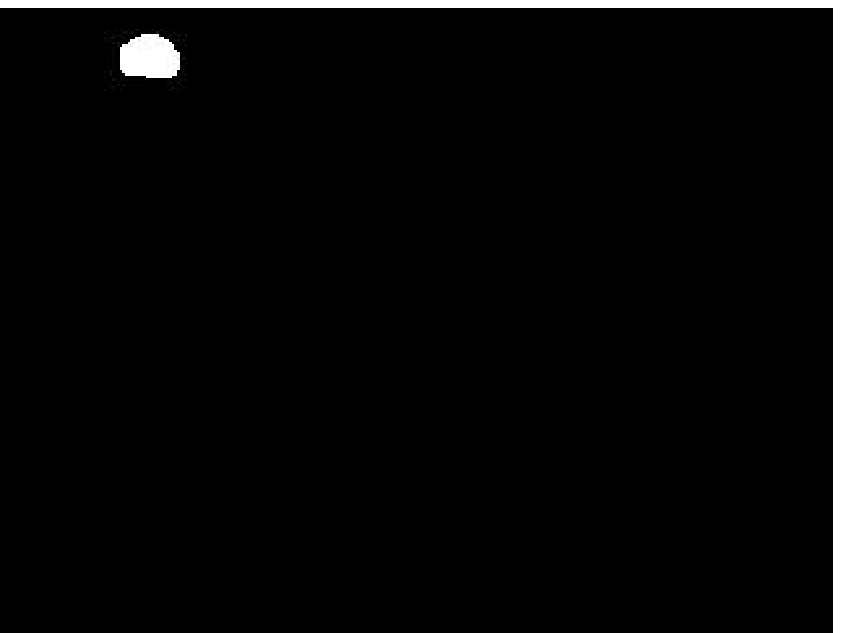}
\includegraphics[width=20mm,height=20mm]{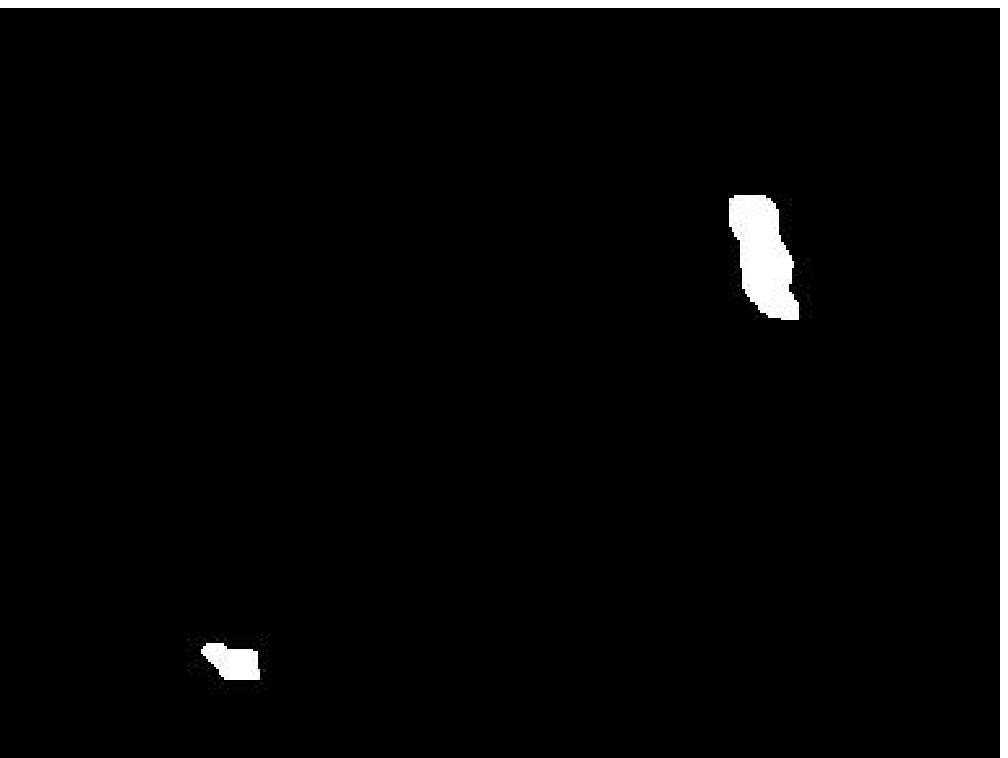}
\includegraphics[width=20mm,height=20mm]{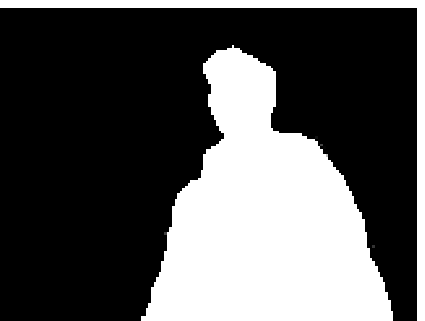}
}

\subfigure[Proposed Methodology]{
\includegraphics[width=20mm,height=20mm]{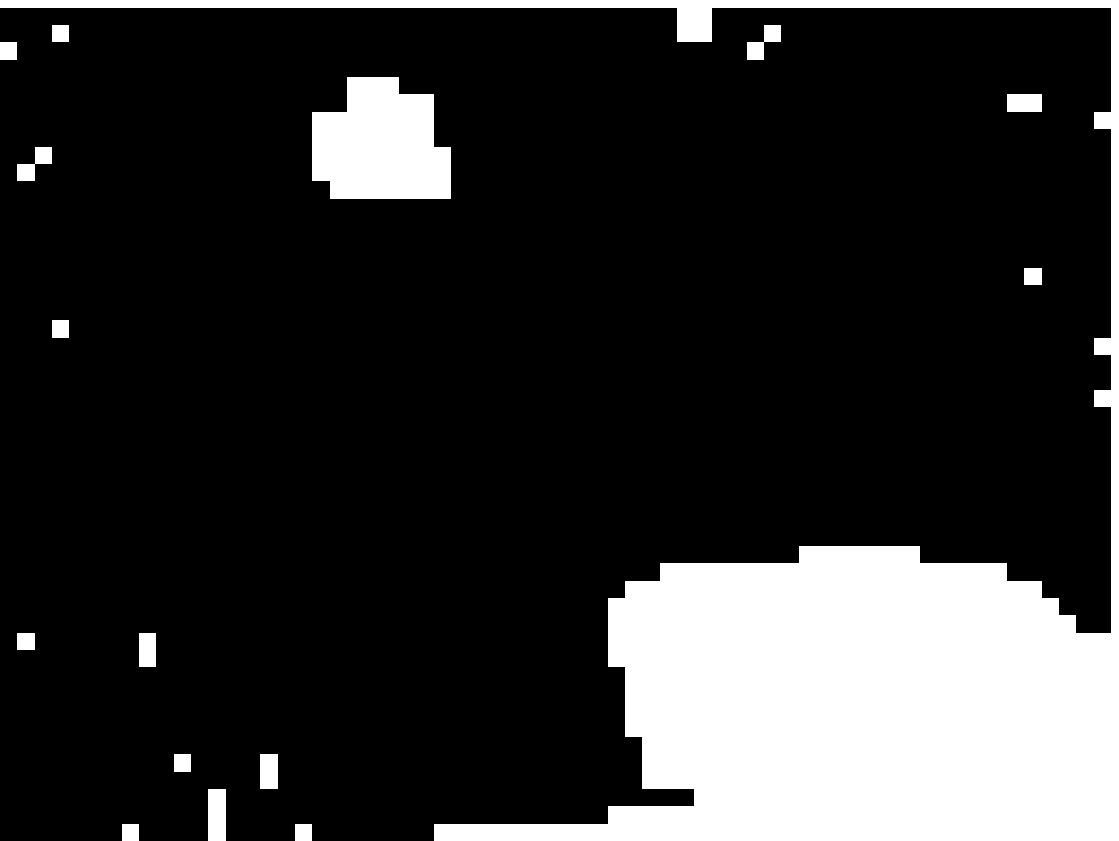}
\includegraphics[width=20mm,height=20mm]{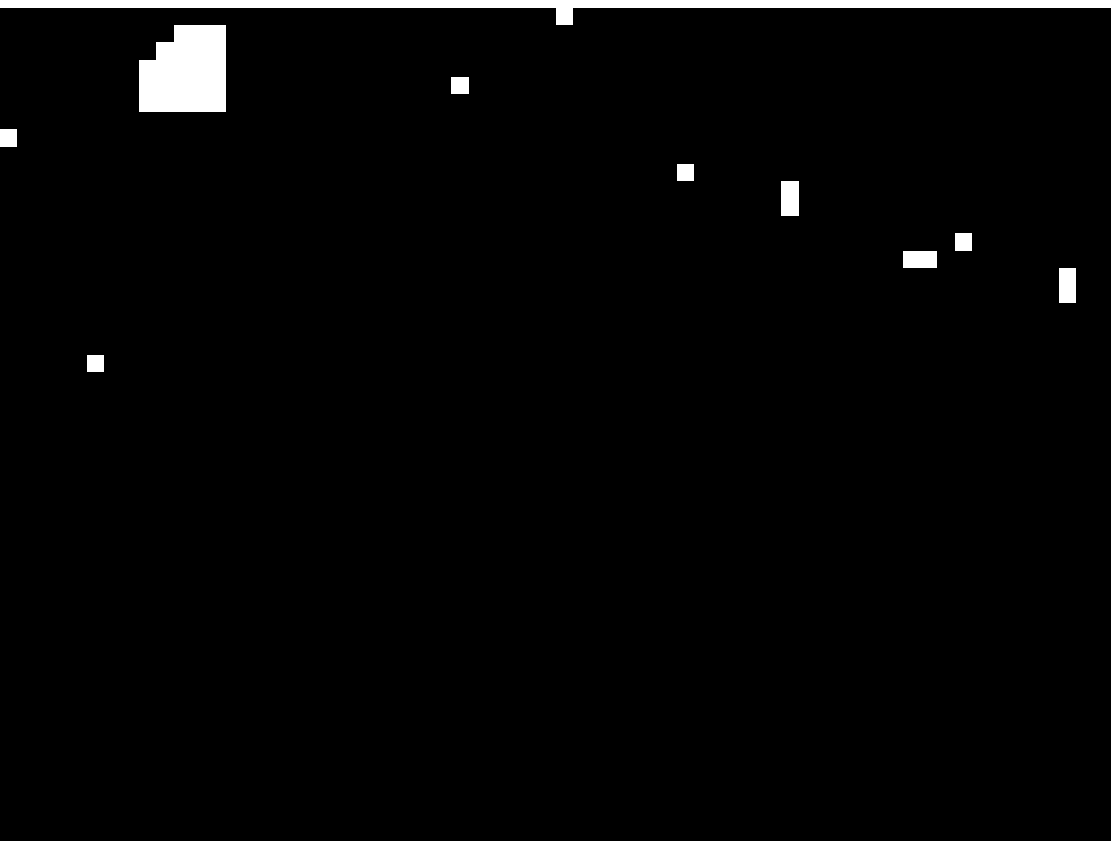}
\includegraphics[width=20mm,height=20mm]{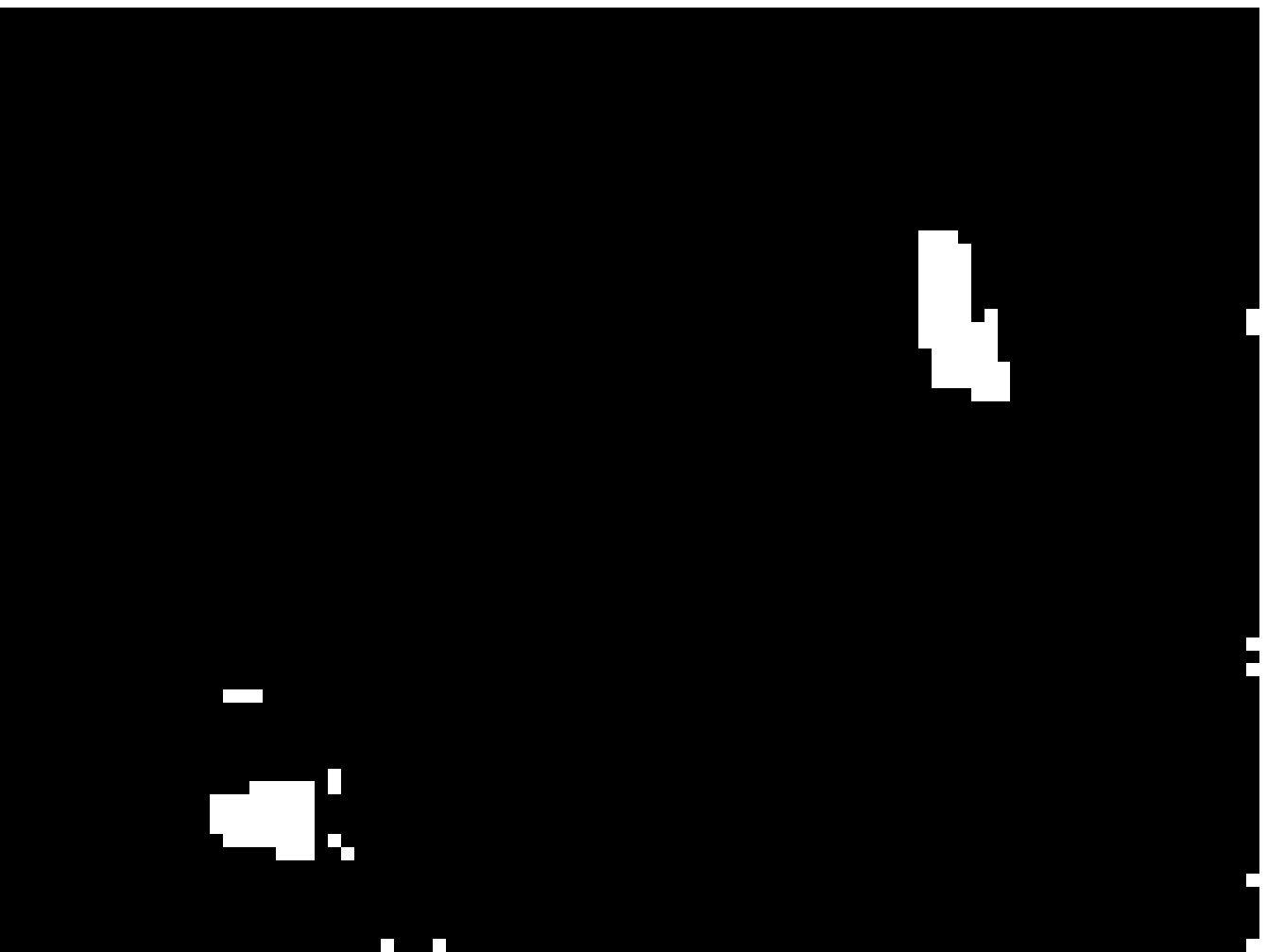}
\includegraphics[width=20mm,height=20mm]{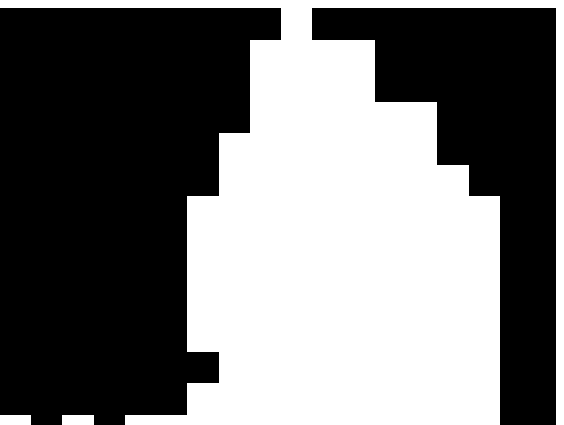}
}

\subfigure[GMM]{
\includegraphics[width=20mm,height=20mm]{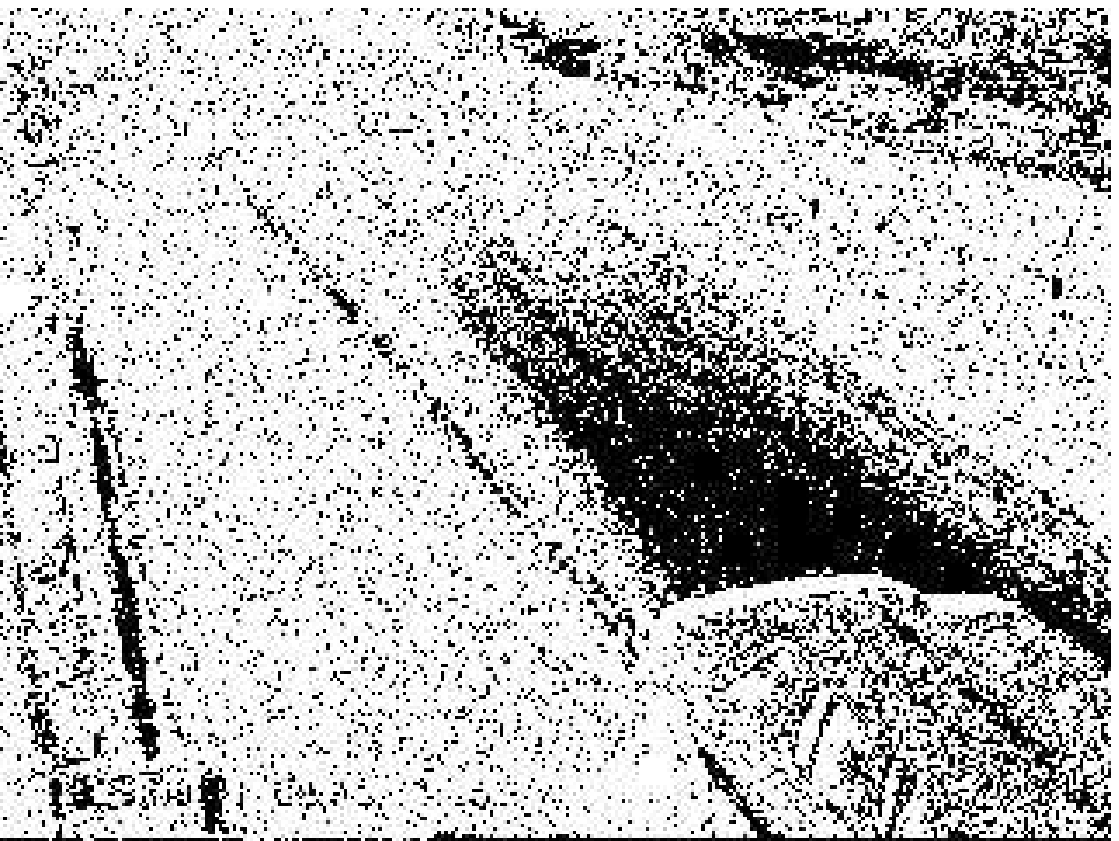}
\includegraphics[width=20mm,height=20mm]{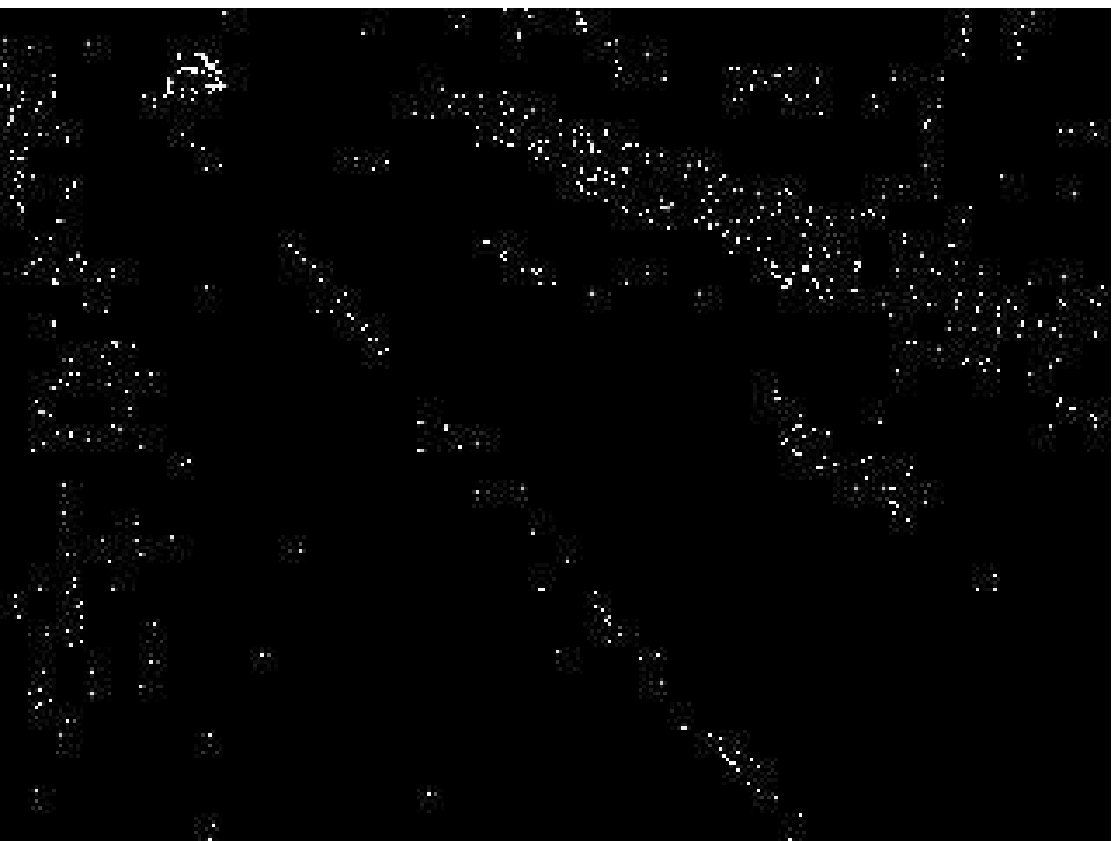}
\includegraphics[width=20mm,height=20mm]{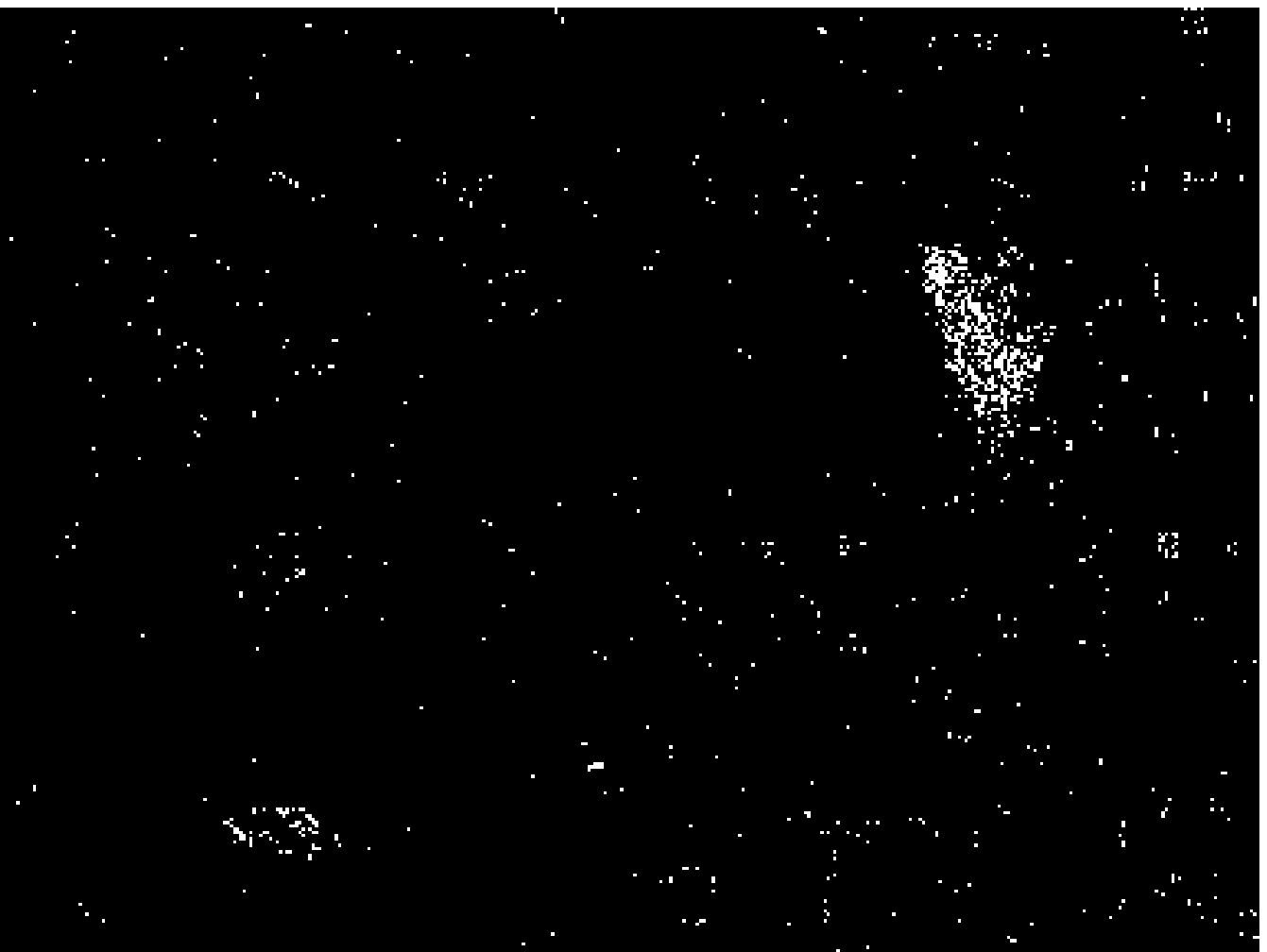}
\includegraphics[width=20mm,height=20mm]{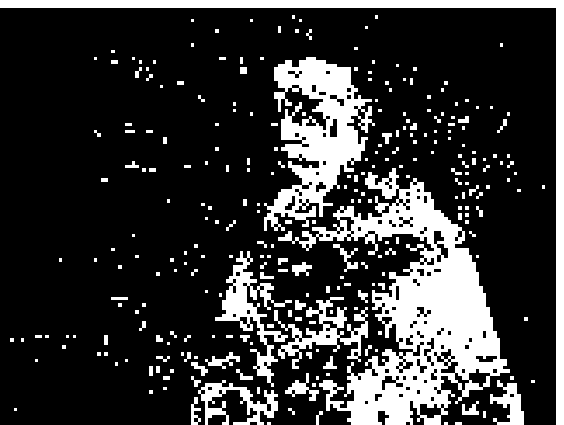}
}

\subfigure[Sigma-Delta]{
\includegraphics[width=20mm,height=20mm]{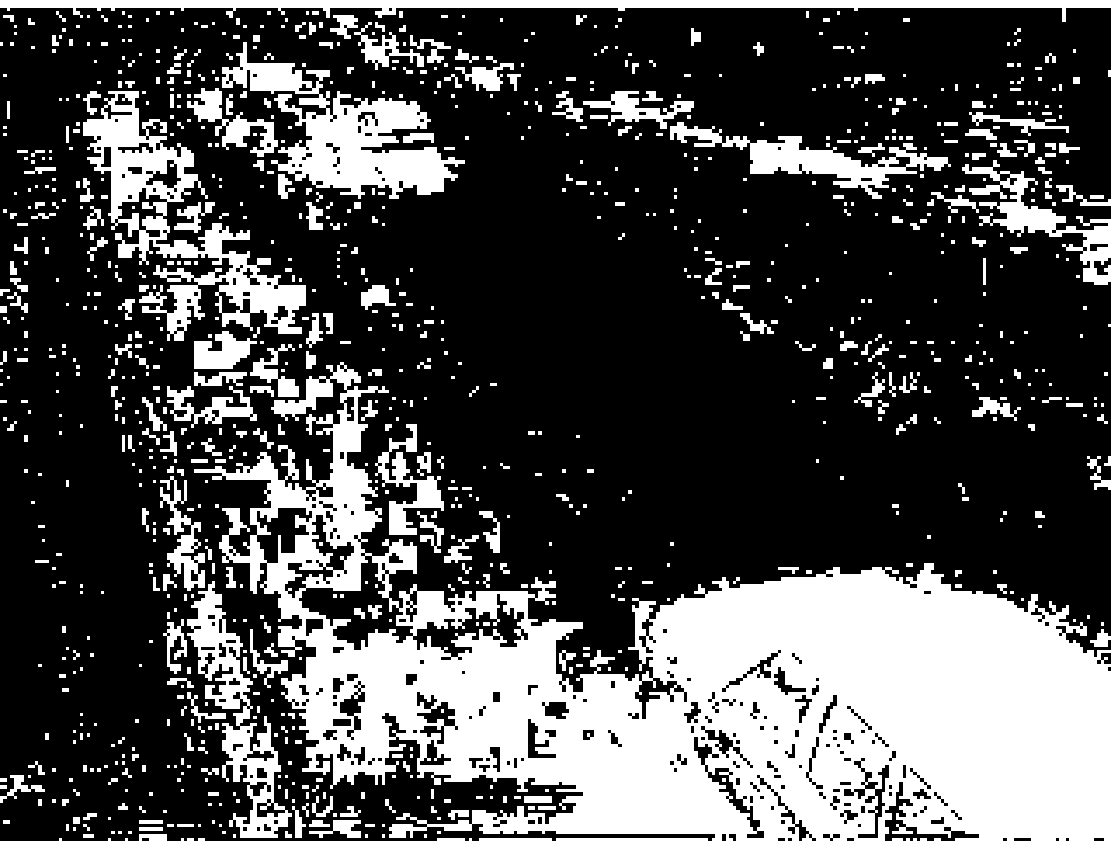}
\includegraphics[width=20mm,height=20mm]{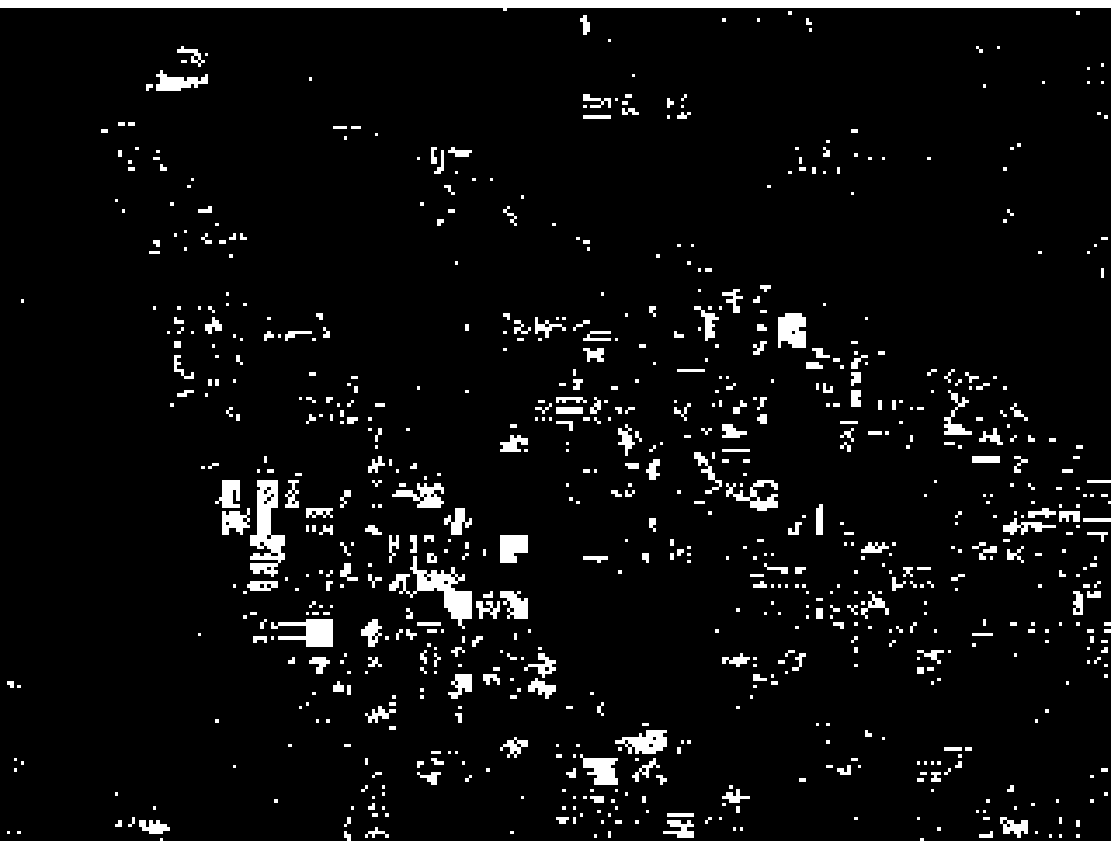}
\includegraphics[width=20mm,height=20mm]{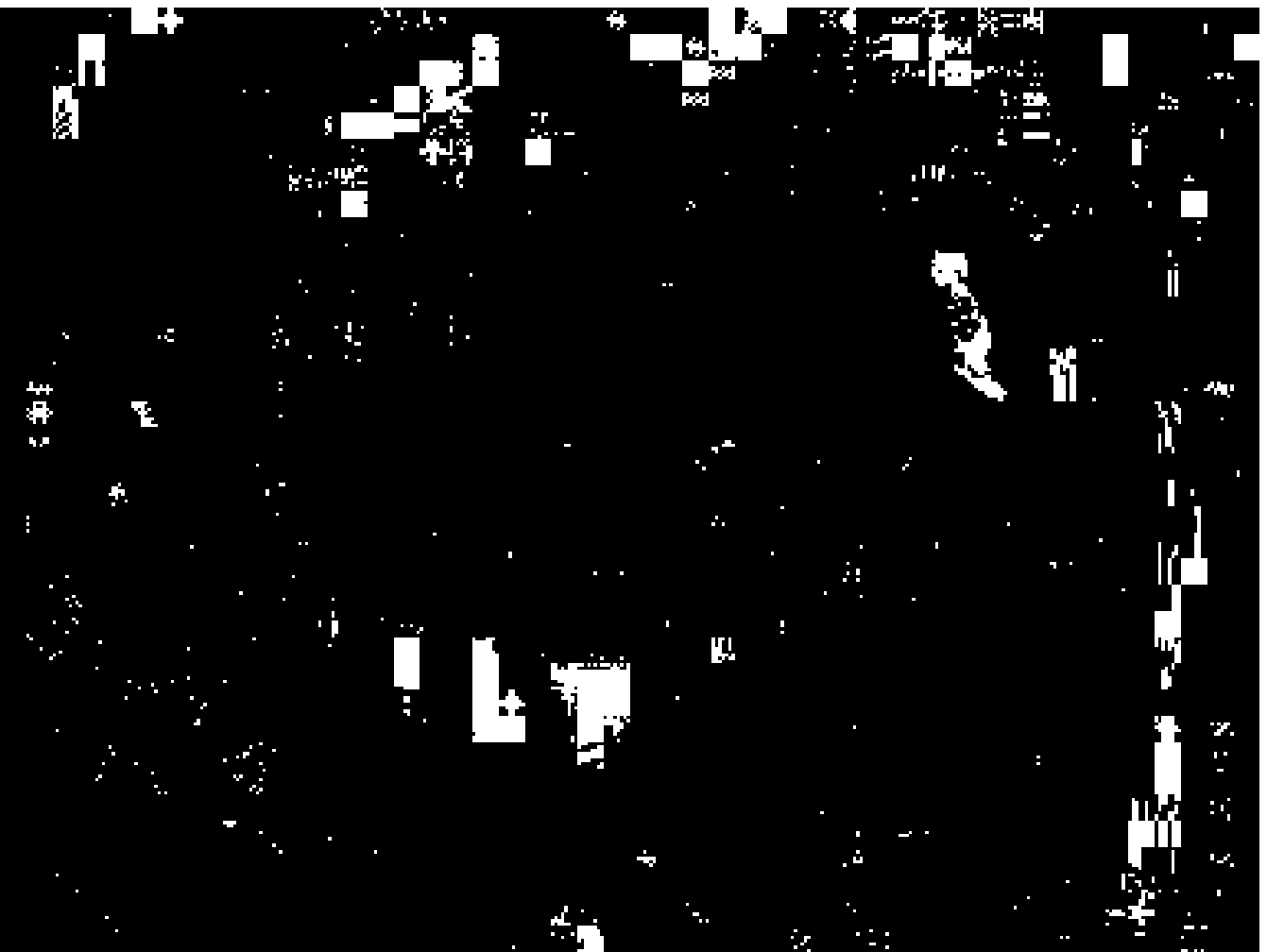}
\includegraphics[width=20mm,height=20mm]{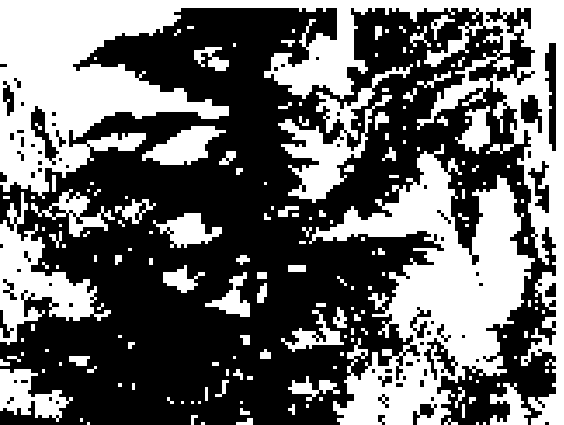}
}

\subfigure[LBP based]{
\includegraphics[width=20mm,height=20mm]{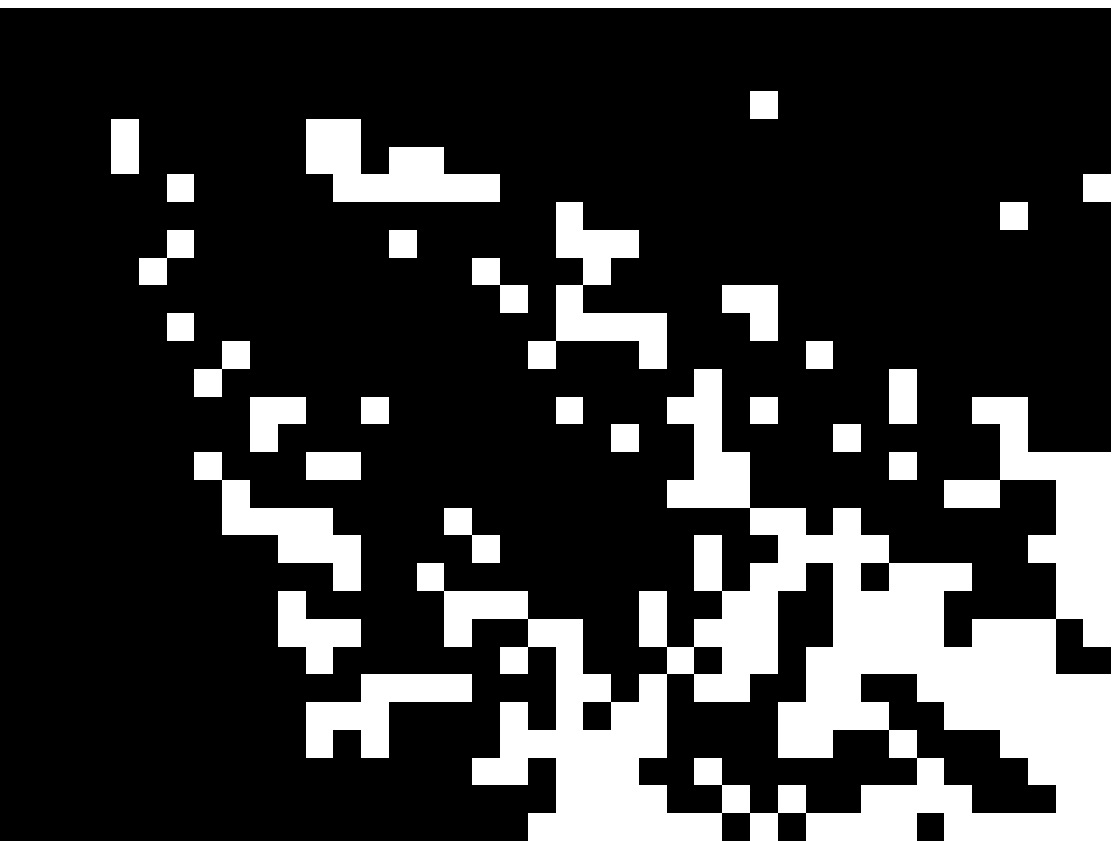}
\includegraphics[width=20mm,height=20mm]{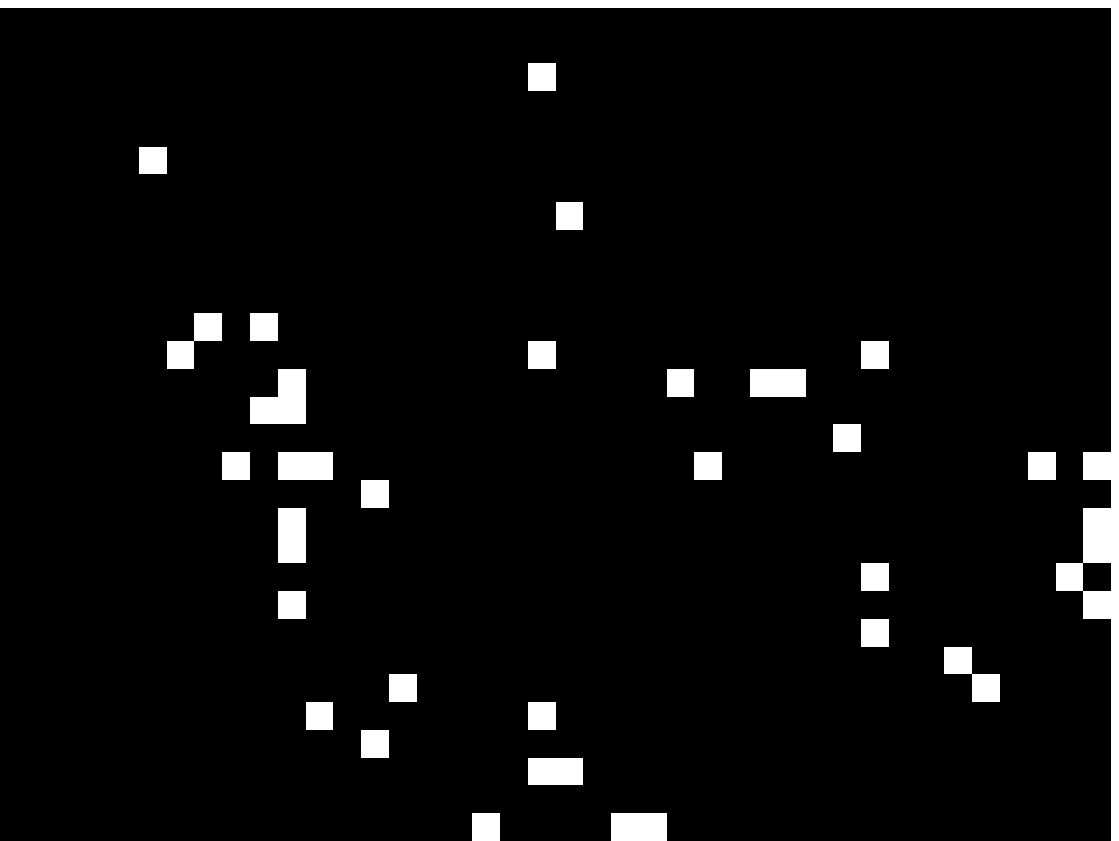}
\includegraphics[width=20mm,height=20mm]{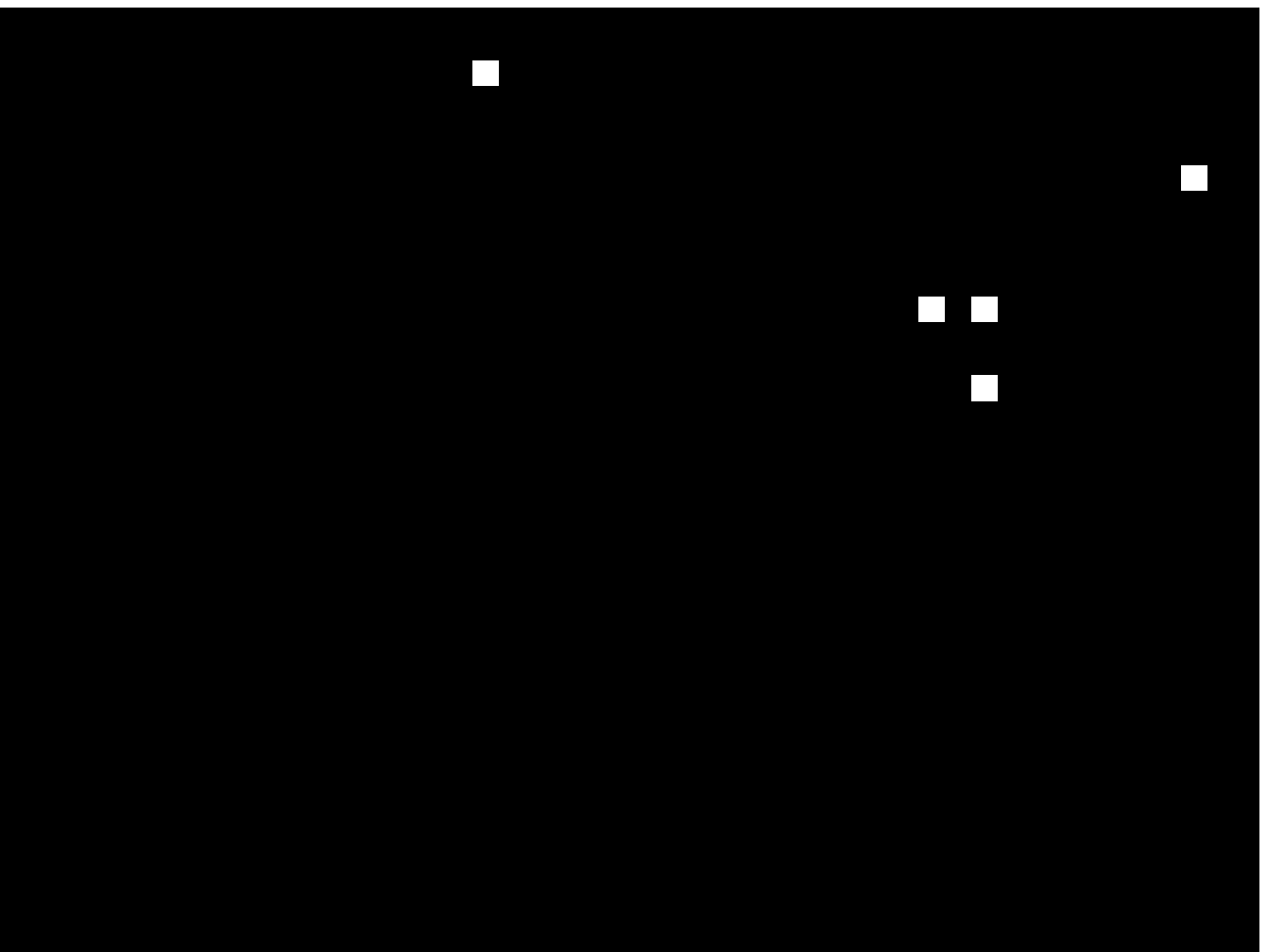}
\includegraphics[width=20mm,height=20mm]{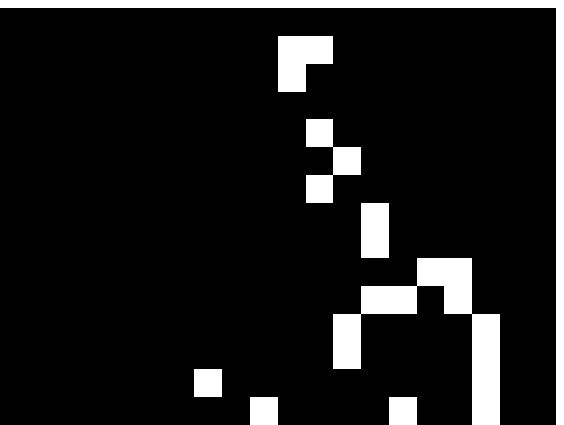}
}

\caption{Ground-truth along with results for all algorithms in the presence of Aperture effects ($f:206$, \textit{Morning}), Noise ($f:572$, \textit{Morning}), Low-resolution human ($f:618$, \textit{CAVIAR}) and Dynamic background ($f:247$, \textit{WavingTrees}).}
\label{fig:frames_effects}
\end{center}
\end{figure}

Figure~\ref{fig:frames_effects} shows the results for frames with different effects. In the presence of aperture effects, the proposed methodology produces a clear outline of the moving object, whereas the GMM and Sigma-Delta approaches produce a high number of false detections in comparison to true detections. The LBP-based technique fails to present any meaningful output. In the presence of noise (banding and random noise) effects, the proposed methodology clearly outperforms the other algorithms. In the presence of low-resolution objects such as a low-resolution human in the \textit{CAVIAR} video, the proposed methodology reported better detection than the other methodologies. The proposed methodology also clearly detects the moving-object (i.e. person) in the \textit{WavingTrees} video in the presence of dynamic background (i.e. waving trees). Detection by state-of-the-art techniques is not as good as the proposed method with the GMM-based technique producing the best performance among the state-of-the-art techniques.

As expected, on fine tuning the parameters for GMM, Sigma-Delta and LBP-based techniques for the \textit{Afternoon} and \textit{Morning} videos individually, there was  improvement in their individual performances, and similarly for the \textit{WavingTrees} video as well. We have not presented those results here.

\subsection{Quantitative Analysis}

An ideal moving object detection algorithm should achieve $100\%$ detection for both \textit{moving} (referred to as positives) and \textit{stationary} (referred to as negatives) pixels.

\begin{figure}[H]
\begin{center}
\includegraphics[width=3.75in,height=2.25in]{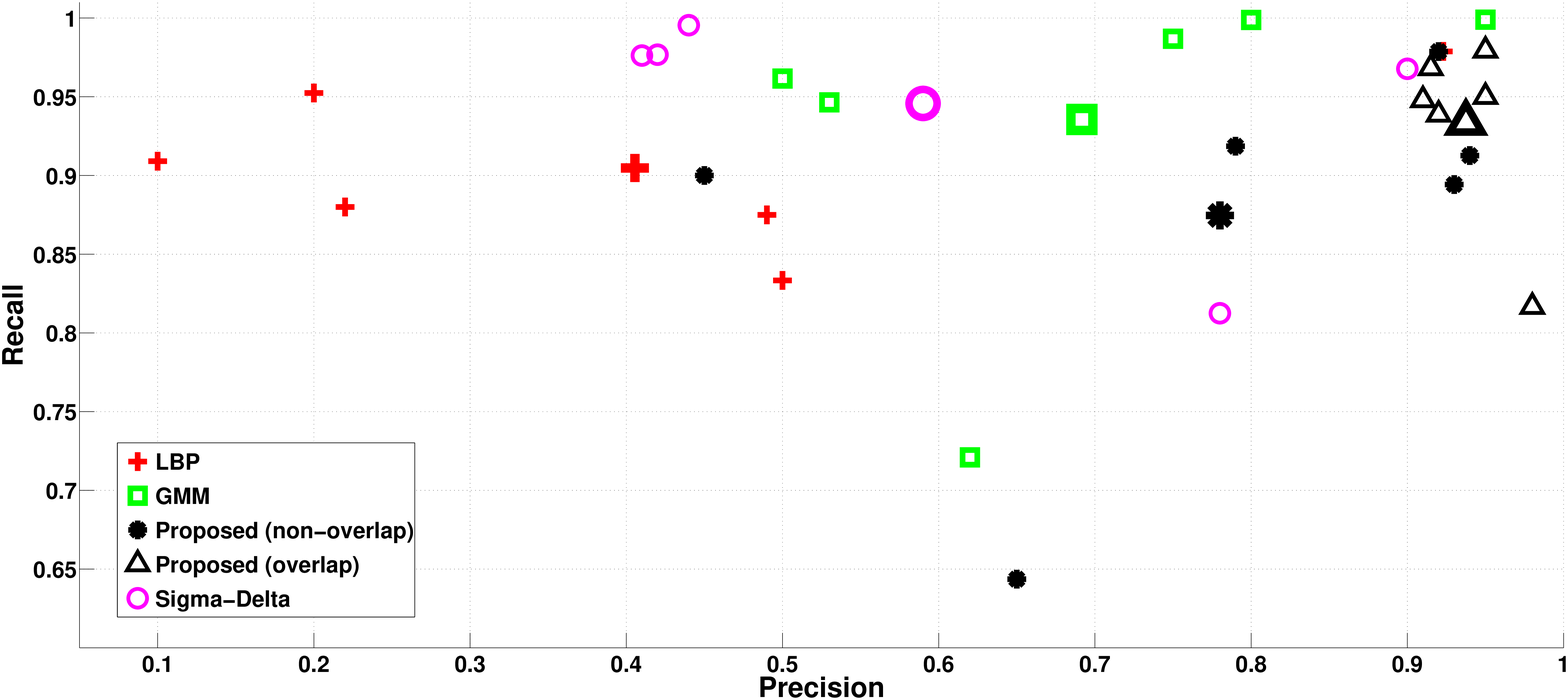}
\caption{Precision vs Recall for all sequences: \textit{Afternoon}, \textit{Morning}, \textit{CAVIAR}, \textit{Hall\_Monitor},\textit{WavingTrees}, and \textit{Mutliple\_Flows}. Average precison-recall for each method is depicted using large fontsize markers.}
\label{fig:tp_vs_tn}
\end{center}
\end{figure}

Figure~\ref{fig:tp_vs_tn} shows the Precision and Recall for all methods.   The proposed methodology outperforms the other algorithms in terms of precison-recall. Between the two, the overlapping method outperforms the non-overlapping method as it can achieve better shape-contour fit to the moving objects. The proposed technique is capable of delivering high performance with generic calibration whereas the performance of existing methods deteriorates significantly.  As expected, the existing methods require parameter fine-tuning for each individual video sequence in order to achieve better performance.

\begin{figure}[H]
\begin{center}
\includegraphics[width=3.75in,height=2.25in]{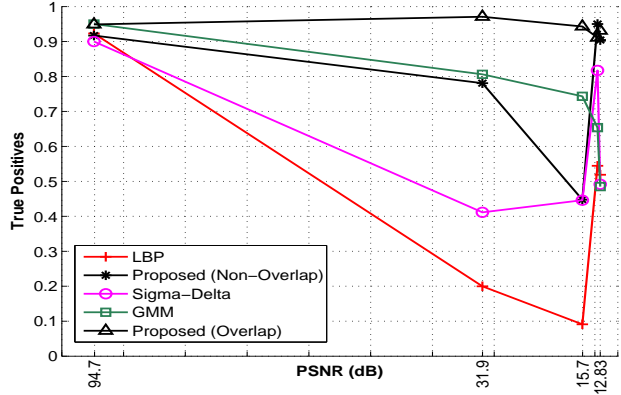}
\caption{Proportions of true positives over the range of peak signal-to-noise ratios (PSNR).}
\label{fig:overall_psnr}
\end{center}
\end{figure}

Figure~\ref{fig:overall_psnr} shows the detection rate for \textit{moving}(positives) pixels with reference to the PSNR of the videos. As the PSNR of video sequences decreases the proposed method clearly outperforms the other systems.  The proposed methodology (overlap) achieves a high detection rate and is also highly consistent for varying PSNR values.
 
\begin{figure}[H]
\begin{center}
\includegraphics[width=3.75in,height=2.75in]{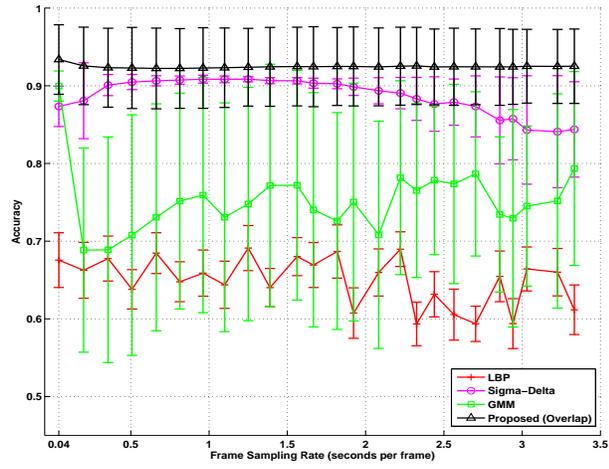}
\caption{Accuracy vs Frame Sampling Rate for all algorithms over all datasets.}
\label{fig:accu_vs_framerate}
\end{center}
\end{figure}

Video cameras are often designed to operate with various frame rates. However, video-based surveillance systems could be designed to further sample the frames in time for processing, depending on the available infrastructure i.e., data transfer capabilities, number of cameras, storage, etc. Figure~\ref{fig:accu_vs_framerate} shows the output accuracy of each algorithm for various frame sampling rates (seconds per frame). The output accuracy is calculated using Equation~\ref{eqn:accu}.

\begin{equation}
\mathrm{accuracy}=\frac{N_{\mathrm{correct}}}{N_{\mathrm{Total}}}, 
\label{eqn:accu}
\end{equation}
where $N_{\mathrm{correct}}$ is the total number of correct detections i.e., true positives $+$ true negatives and $N_{\mathrm{Total}}$ is the total number of pixels in the frame.

The proposed methodology provides high accuracy consistently for various frame sampling rates of the input video. The Sigma-Delta method is the second best with LBP-based methodology achieving the lowest accuracy compared to the other techniques. Error-bar depicts the standard deviation of accuracy over the entire dataset. The GMM-based technique has high accuracy for frame sampling rate at $0.04$ seconds per frame (i.e, $25$ frames per second). However, as the frame sampling rate increases, accuracy decreases and the size of the error-bar increases across the entire dataset. This is a clear indication that at low-frame rates the GMM-based foreground object detector is not reliable. For the Sigma-Delta technique, as the frame sampling rate increases the accuracy increases and the size of the error-bar decreases until the accuracy reaches a peak between $1$ and $1.5$ seconds per frame. Then the accuracy starts to decline and the size of the error-bar starts to increase.  The LBP-based methodology consistently provides low accuracy with low standard deviation. The proposed methodology provides high accuracy irrespective of the frame sampling rate with consistent standard deviation across the entire dataset. 

\begin{figure}[H]
\begin{center}
\includegraphics[width=3.75in,height=2.75in]{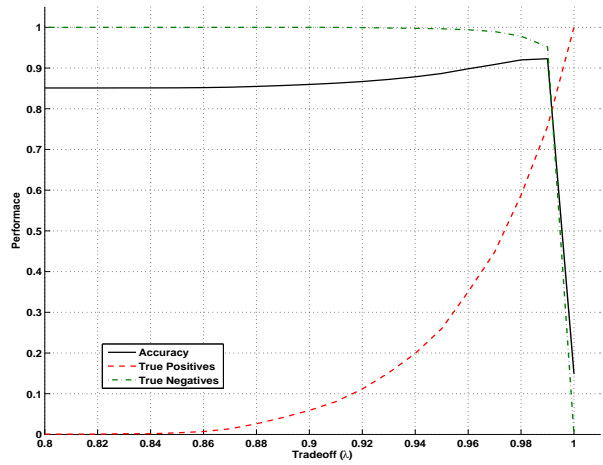}
\caption{Performance of the proposed methodology over the entire dataset w.r.t. accuracy, and proportions of correctly classified pixels (i.e.,true-positives and true-negatives) for varying tradeoff values.}
\label{fig:accu_tp_tn_vs_tradeoff}
\end{center}
\end{figure}

\begin{figure}[H]
\includegraphics[width=\textwidth,height=3in]{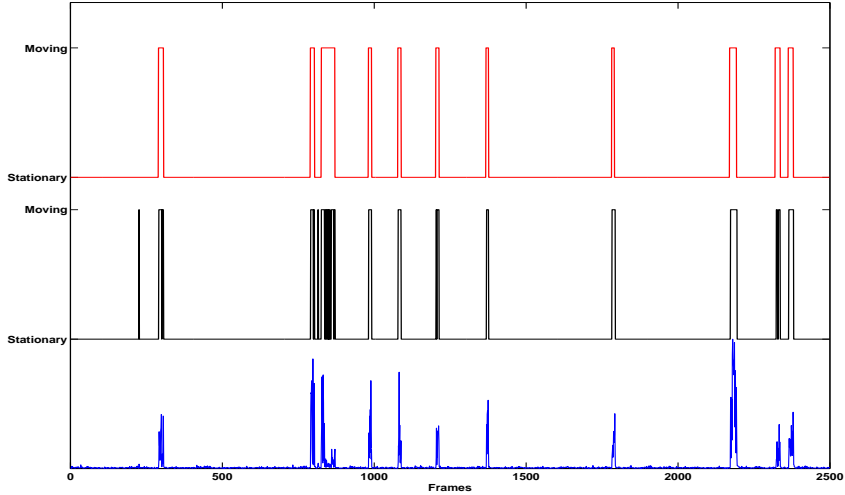}
\caption{From top to bottom: Ground-Truth , System Output and Motion-likelihood ($\Delta$) graphs for $(10,5)^{th}$ texture-block from $2500$ frames of the \textit{Afternoon} sequence using the proposed methodology. Parameters used: Learning rate of $\alpha = 0.05$ and tradeoff of $\lambda = 0.98$. Detection error $\approx 1.96\%$ (incorrect classifications $=49$).}
\label{fig:tex_example}
\end{figure}

\begin{figure}[H]
\includegraphics[width=\textwidth,height=3in]{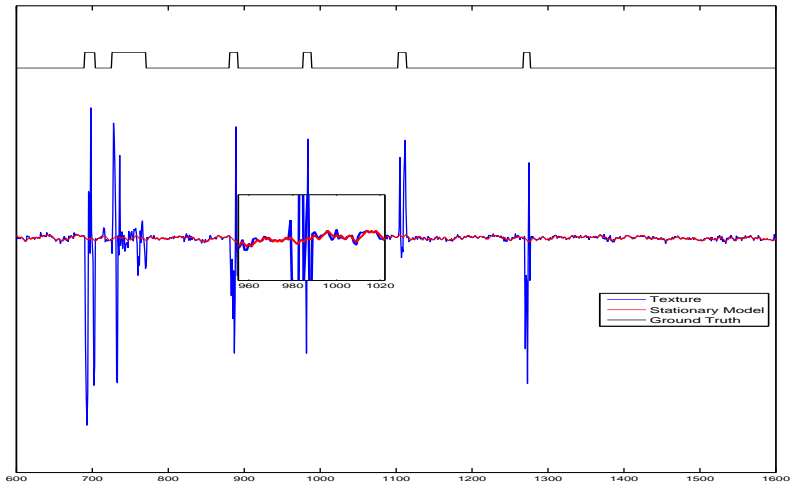}
\caption{Ground Truth, Texture measure ($1^\text{st}$ dimension only, blue) and Stationary Model ($1^\text{st}$ dimension only, red) for $(10,5)^{th}$ texture-block from frames of the \textit{Afternoon} sequence using the proposed methodology. Parameters used: Learning rate of $\alpha = 0.05$ and tradeoff of $\lambda = 0.98$. Inset: Magnified version of a small section.}
\label{fig:1dtex_eg}
\end{figure}

Figure~\ref{fig:accu_tp_tn_vs_tradeoff} shows the performance of the proposed methodology when varying the tradeoff parameter. Highest accuracy is achieved for a tradeoff of $\lambda=0.99$, and appears to remain constant at around $0.85$ for any tradeoff less than $0.86$. This accuracy, however, is not valid as the proportions of true positives is zero.  The range of tradeoff is quite limited, with the possibility of detecting true moving-pixels using a tradeoff between $0.86$ and $0.99$. For a tradeoff of $\lambda < 0.86$, the system considers all pixels as background and for a tradeoff of $\lambda \geq 1$, the system considers all pixels as moving.  

A sample output for a texture-block is shown in Figure~\ref{fig:tex_example}. There were $49$ incorrect classifications (false positives $+$ false negatives) out of a total of $2500$ instances i.e. an error rate $\approx 1.96\%$. Figure~\ref{fig:1dtex_eg} shows the $1^\text{st}$ dimension of the derived texture measure along with the generated stationary model. The stationary model is effective in modelling background texture, including the associated noise.

\section{Application: Vehicular Traffic Density Estimation}
\label{sec:appln}


\label{sec:intro}
Intelligent Transportation Systems (ITS) are used in many cities to provide information about traffic conditions on road networks and  as an aid to streamlining vehicular traffic flow in an effort to reduce  traffic congestion. An ITS typically uses various kinds of sensors,  such as video cameras and inductive loop detectors to measure the significant properties of vehicular traffic flow. Traffic density is a useful property that an ITS can use to  perform higher level functions such as traffic light sequencing.  

Video monitoring systems promise many advantages over the now-dominant inductive loop detectors  which are point detectors that sense the vehicles passing over them~\cite{Beymer97}. Cameras are cost effective and easier to maintain than other road-side mounted sensors. They also offer the potential of providing a much richer data stream than the simple loop.

Most of the state-of-the-art techniques have relied upon modelling all density states using training data and then classifying traffic density state of the testing data~\cite{Faith04,Pork04,Jing07,pranamits09}.There is a real need to develop systems that can analyse surveillance footage especially in the field of vehicular traffic analysis in varying noise and illumination conditions.

Vehicle tracking is another popular technique for traffic density estimation~\cite{Koller94,Tseng02,Kamijo00}. A vehicle detection algorithm provides vehicle regions in the frame which are used by a tracking algorithm to perform vehicle correspondence from one frame to the next and generate tracks.  A count of individually tracked vehicles will provide the percentage of lane coverage.

Vehicle detection approaches can be placed into four different categories, namely (a) Point-detectors such as Difference-of-Gaussian detector~\cite{Lowe04} and Harris detector~\cite{Harris88}; (b) Segmentation-based methods such as Mean-shift~\cite{Coma99} and Graph-cut~\cite{Shi97}; (c) Background modelling-based methods such as Mixture of Gaussians~\cite{gmm} and  (d) Supervised learning-based methods such as Adaptive boosting~\cite{Viola03} and Support Vector Machines~\cite{Papa98}.  Tracking algorithms  have been classified by Yilmaz et al.~\cite{Yilm06} into three different categories: (a) Point-based tracking (MGE tracker~\cite{Sala90}, Kalman filters~\cite{Broi86}, etc), (b) Kernel-based tracking (Mean-shift~\cite{Coma03}, KLT~\cite{Shi94}, SVM tracker~\cite{Shai04}, etc) and (c) Silhouette-based tracking (Hough-transform~\cite{Sato04}, Variational methods~\cite{Bert00}, etc). 

Point-based tracking is effective for small objects, because small objects can be represented using a single point. Large objects have multiple points, so all points of an object need to be automatically clustered so as to register the points as belonging to that object. Automatic clustering is another difficult problem considering that a frame could consist of multiple foreground objects and a background. Kernel-based tracking is used for estimating the object motion.  An object motion model can be used to estimate the object region and object orientation in the next frame, thereby tracking the object through frames. This approach is reliant on the geometric shapes of object representation and is computationally very expensive. Silhouette-based methods provide accurate shape descriptions of objects and then trackers are used to find the region in each frame using the object model generated. Vehicle tracking algorithms are generally flexible enough to determine almost any type of traffic information. Selecting 
the best object detector and object tracker for a particular application is dependent on many parameters such as type of object, scenario, type of background, etc. In terms of just determining the traffic density, the performance of vehicle tracking algorithms tends to degrade in heavy traffic situations due to occlusion, clutter and false background estimation. However, individual object tracking in the presence of occlusion, as seen in the dataset, is considered in this paper. 
 
 An alternate approach for traffic density estimation is to detect objects inside a traffic lane and then deduce the traffic density. In this section, we explore the possibility of using a foreground object detector for detecting vehicles and estimating vehicular traffic density.   


We present the test dataset and the framework based on the foreground object detector including the experimental evaluation in sections~\ref{sec:dataset} and ~\ref{sec:MOD} .

%
\subsection{Dataset}
\label{sec:dataset}

\begin{figure}
\begin{center}
\includegraphics[width=1.\linewidth,scale = 0.5,height=3.5cm]{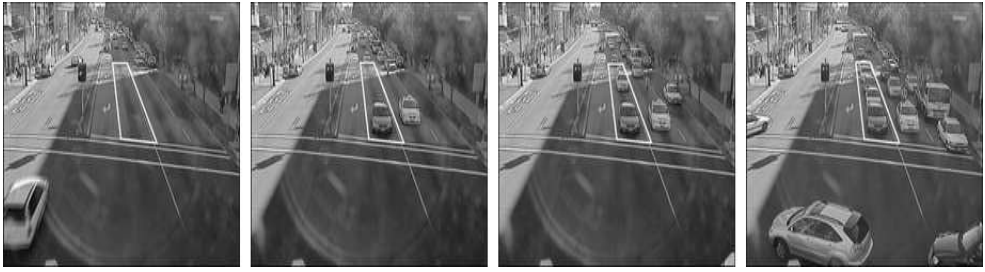}
\caption{\textit{Empty}, \textit{Low}, \textit{High} and \textit{Full} states of a lane of traffic at the Anzac Parade/Barker Street junction (the circular reflection is that of the camera lens onto the window of the camera housing).}
\label{fig:fourstates}
\end{center}
\end{figure}

All our experiments use video sequences captured from a busy road junction located just outside the Sydney CBD (Anzac Parade/ Barker Street Junction) and are proprietary dataset of the Roads and Transport Authority, New South Wales, Australia.  All video sequences  were captured during daytime, exhibiting the expected  range of natural lighting variation; the cameras were on auto-calibration mode. Each video sequence is of 6 minutes---9000 frames---duration. The resolution of each frame is $320 \times 240$. In our experiments we have used only one lane as the ROI. Figure~\ref{fig:fourstates}  shows four sample frames  (with ROI outlined in white) that are representative of the \textit{Empty}, \textit{Low}, \textit{High} and \textit{Full} traffic density states used in this paper. Table~\ref{tab:videoinfo}  details the properties of each of the video sequences.

%
%
\begin{table}
\caption{Properties for each of the video sequences used for testing. All these video sequences were captured by traffic cameras installed at Anzac Parade/Barker Street Junction, Sydney, Australia. 
 }
\label{tab:videoinfo}
\begin{tabular}{|c|c|c|c|c|c|}
\hline
\multirow{5}{*}{\rotatebox[]{90}{Properties}}
& \multirow{2}{*}{Name} &\multicolumn{4}{|c|}{Anzac Parade / Barker Street Jtn}\\ \cline{3-6}
& & Jtn.-$1$ & Jtn.-$2$ & Jtn.-$3$ & Jtn.-$4$ \\ \cline{2-6}
& Illumination & Bright/ & & & \\
&  level & Sunny & Raining & Cloudy & Sun-Glare\\ \cline{2-6}
& Noise/Artifacts: &  & & &\\
& aperture & $\checkmark$ & $\checkmark$ & $\checkmark$ &  $\checkmark$ \\
& banding & $\checkmark$ & $\checkmark$ & $\checkmark$ & $\checkmark$ \\
& random/shot & $\checkmark$ & $\checkmark$ & $\checkmark$ & $\checkmark$ \\
& compression & $\checkmark$ & $\checkmark$ & $\checkmark$ & $\checkmark$ \\
\hline
%
\end{tabular}
\end{table}
\textit{Traffic density} is defined  as the percentage of the ROI (usually a lane segment) occupied by vehicles. Traffic density of the particular traffic lane is classified into one of four states: 

\begin{enumerate}
\item \textit{Empty:}   less than 5\% of the lane is occupied by vehicles.
\item \textit{Low:} 5--30\% of the lane is occupied by vehicles.
\item \textit{High:}  30--90\% of the lane is occupied by vehicles.
\item \textit{Full:} more than 90\% of the lane is occupied by vehicles.
\end{enumerate}

Human annotated ground-truth for all the frames of all the videos was made available to us. 
%
\subsection{Framework}
\label{sec:MOD}
%

\begin{figure*}
\begin{center}
\includegraphics[width=\textwidth,height=2.25in]{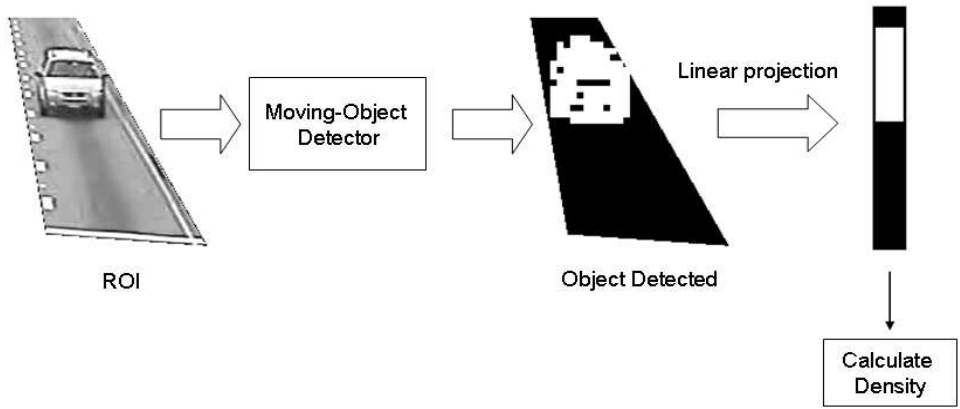}
\caption{Object detection-based framework for vehicular traffic density estimation}
\label{fig:MOD_framework}
\end{center}
\end{figure*}

We consider an ROI within which we need to estimate the traffic density. For every input frame, pixels belonging to the foreground objects are identified using the foreground object detectors and are segmented into regions by a two-pass, connected components algorithm~\cite{HornRobot}. The generated mask  is projected onto a linear scale. The percentage of non-zero elements versus the zero-elements in the linear scale would provide the vehicular traffic density for that particular traffic lane. 
There is neither a explicit training phase nor density-state learning phase. The performance of this framework is entirely dependent upon the performance of the foreground object detector. Figure~\ref{fig:MOD_framework} illustrates the foreground object detection-based approach for vehicular traffic density estimation.

An alternate method of estimating traffic density would be to use the foreground object detector to detect foreground objects and use a tracking algorithm to track individual objects. A count of individual objects will provide the percentage of lane coverage. However, individual object tracking in the presence of occlusion, as seen in the dataset, is beyond the scope of this paper.

In section~\ref{sec:exper_tde}, we present the experimental evaluation consisting of the performance results of the foreground object detection-based framework on the evaluation dataset. 


\subsection{Experiments}
\label{sec:exper_tde}
Experimental results for the foreground object detection-based framework is based on each sequence, and there is no explicit training or testing phase. Parameters for the Proposed, GMM, Sigma-Delta, LBP-based approaches were fine-tuned using the  \textit{Multiple$\_$Flows} sequence as presented in Section~\ref{sec:exp_setup}.

RGB video frames were used in evaluating GMM-based technique, with three Gaussian distributions per pixel per color component. The learning rate and foreground threshold were set to $0.01$ and $0.25$, respectively. A match was defined as a pixel value within $2$ standard deviations of the distribution's mean. Grayscale video frames were used in evaluating of Sigma-Delta based approach. Temporal pixels were identified as the pixels whose variation rate was $N =4$ times the non-zero differences. Grayscale video frames were used in evaluating the LBP-based approach, with three histogram mixtures. The rate of learning for the histogram mixture and weights were both set to $0.005$. The pixel block-size was set to $8\times8$ pixels, with a detection threshold of $0.25$. We set the overlapping region size as half the block-size. 


\begin{figure}
\begin{center}
\subfigure[Traffic densities:~\textit{Empty},~\textit{Low},~\textit{High},~\textit{Full}]{
\includegraphics[width=20mm,height=30mm]{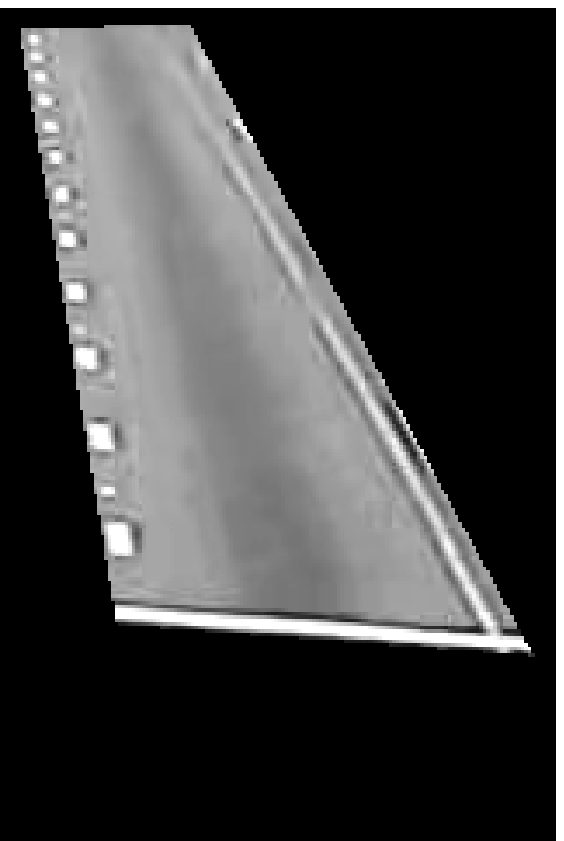}
\includegraphics[width=20mm,height=30mm]{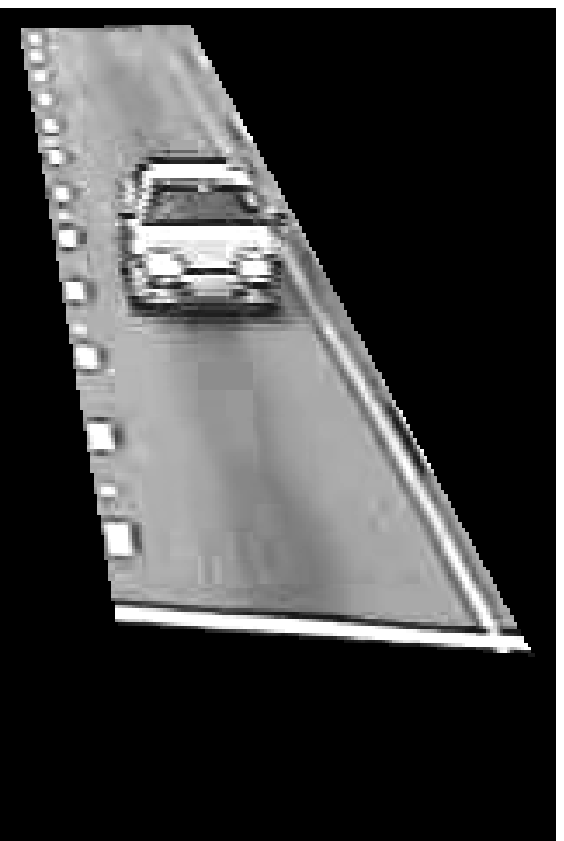}
\includegraphics[width=20mm,height=30mm]{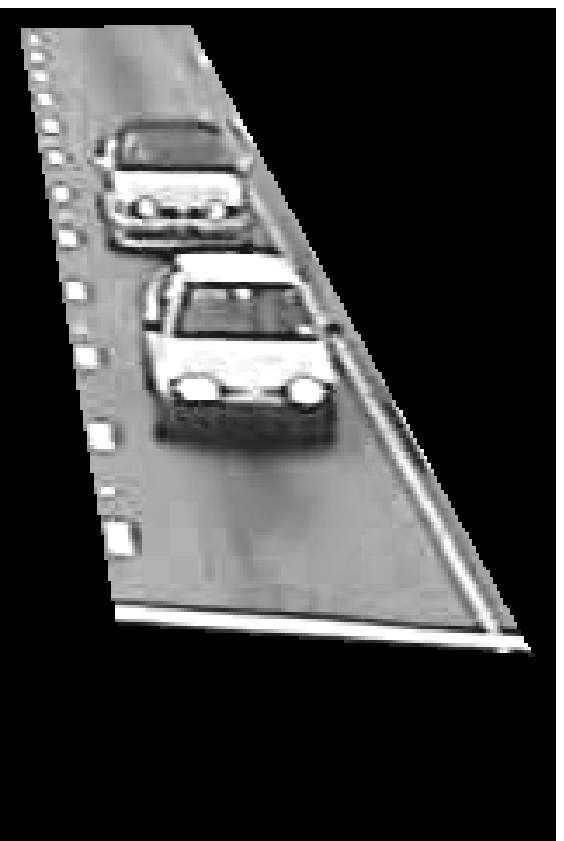}
\includegraphics[width=20mm,height=30mm]{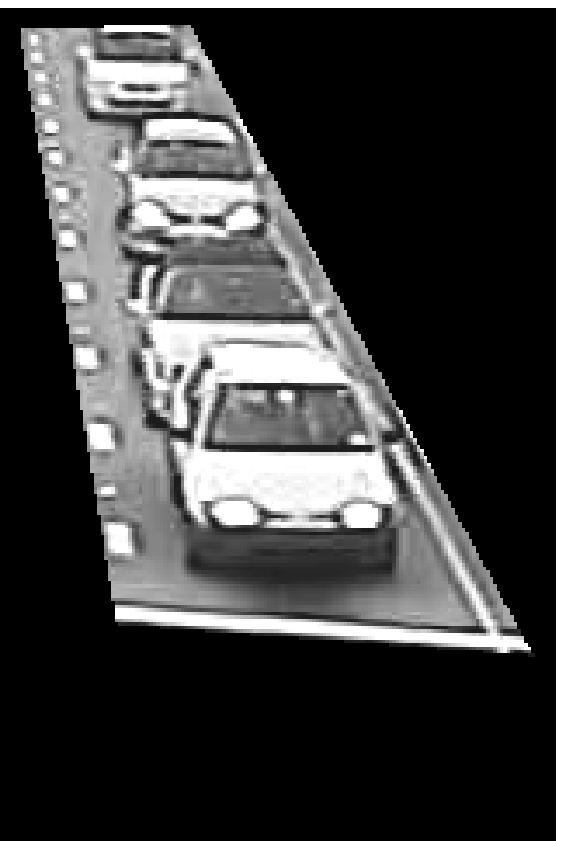}
}

\subfigure[Proposed Methodology]{
\includegraphics[width=20mm,height=30mm]{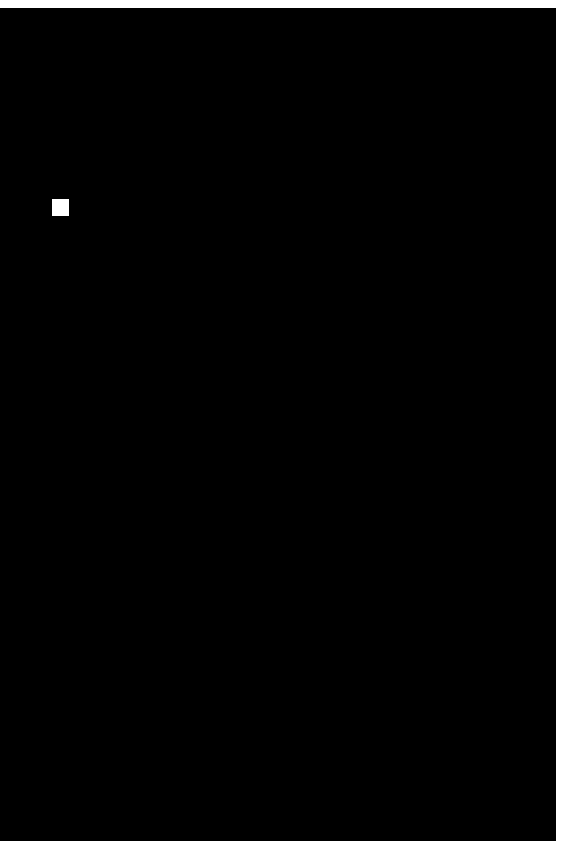}
\includegraphics[width=20mm,height=30mm]{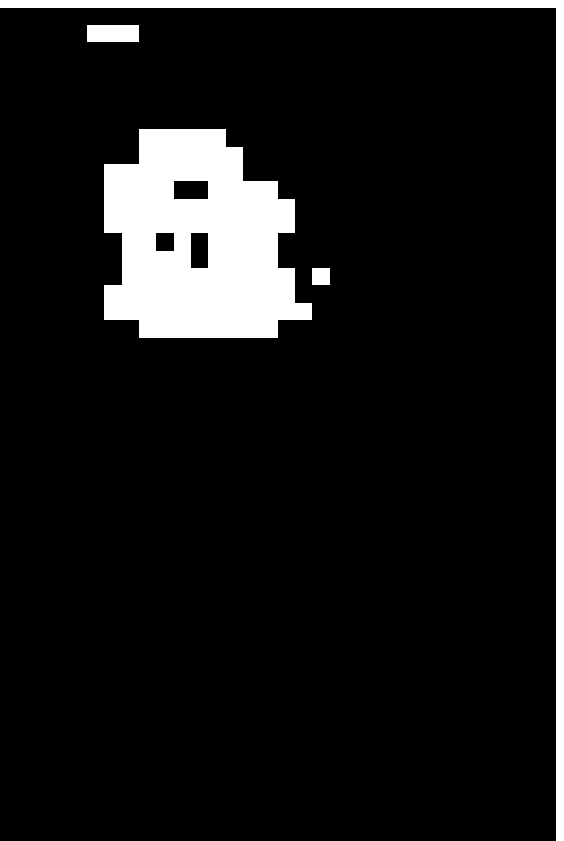}
\includegraphics[width=20mm,height=30mm]{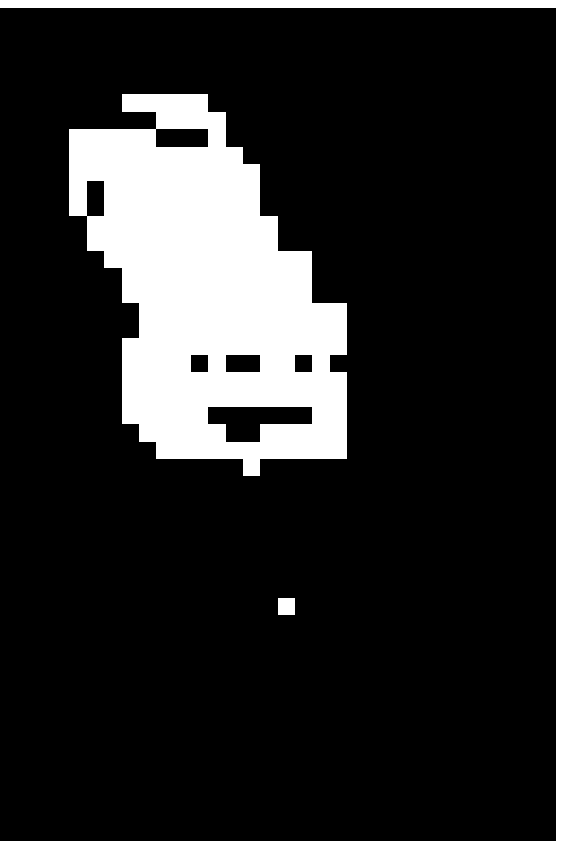}
\includegraphics[width=20mm,height=30mm]{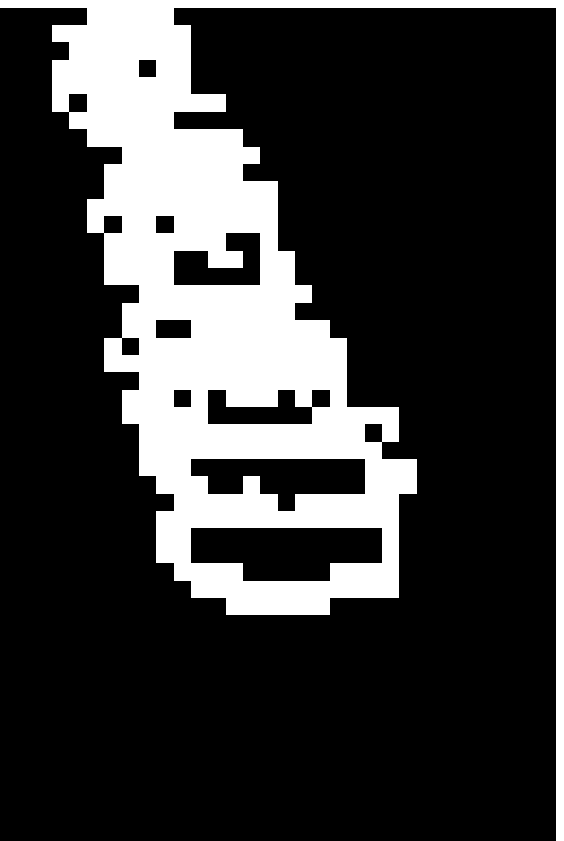}
}

\subfigure[GMM]{
\includegraphics[width=20mm,height=30mm]{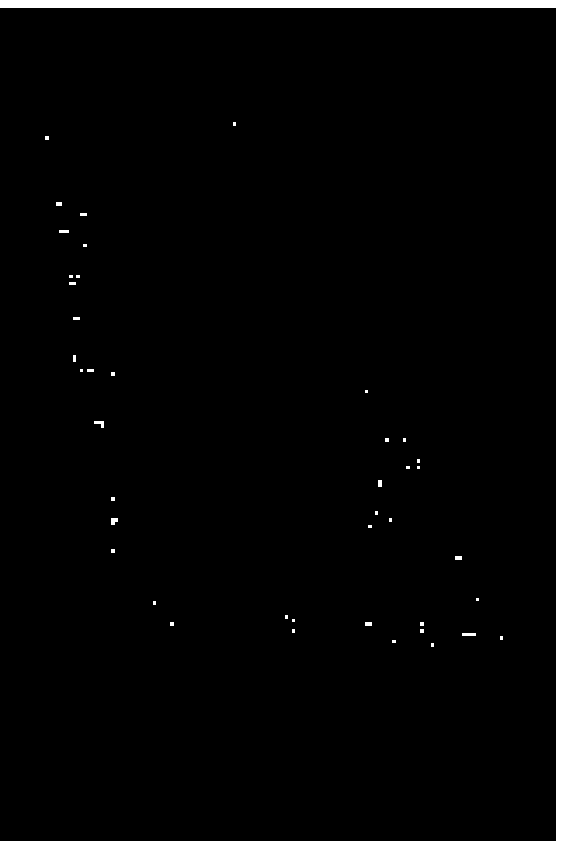}
\includegraphics[width=20mm,height=30mm]{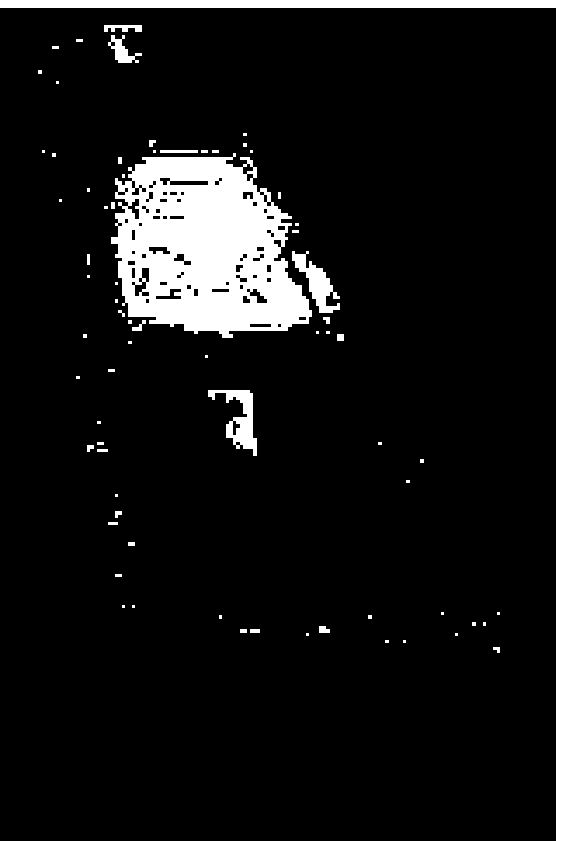}
\includegraphics[width=20mm,height=30mm]{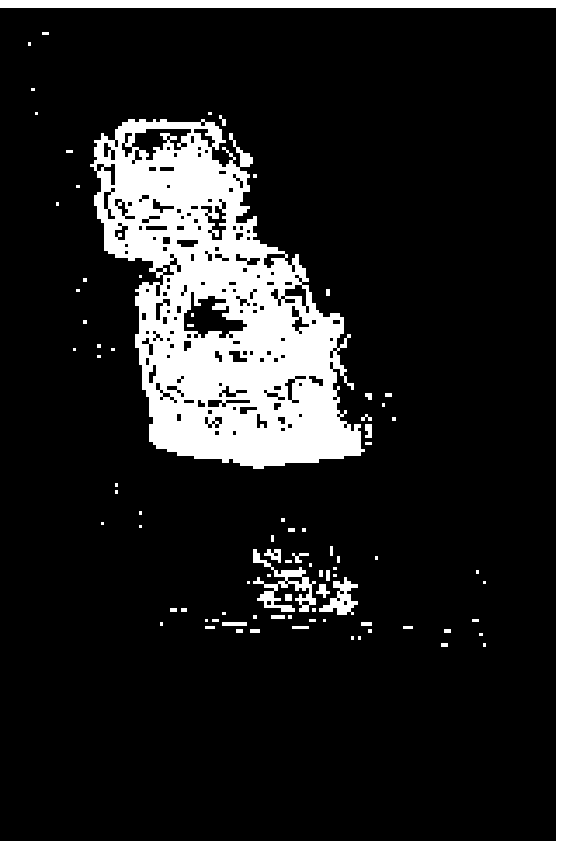}
\includegraphics[width=20mm,height=30mm]{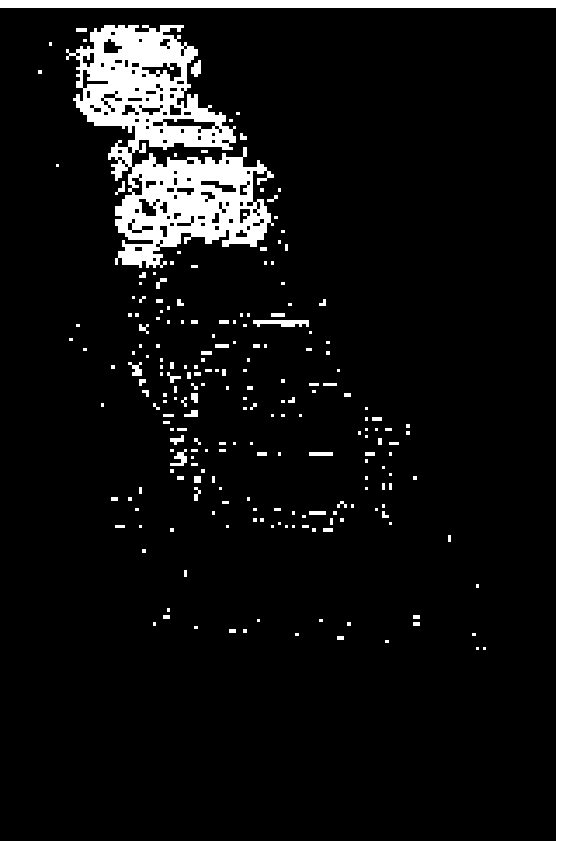}
}

\subfigure[Sigma-Delta]{
\includegraphics[width=20mm,height=30mm]{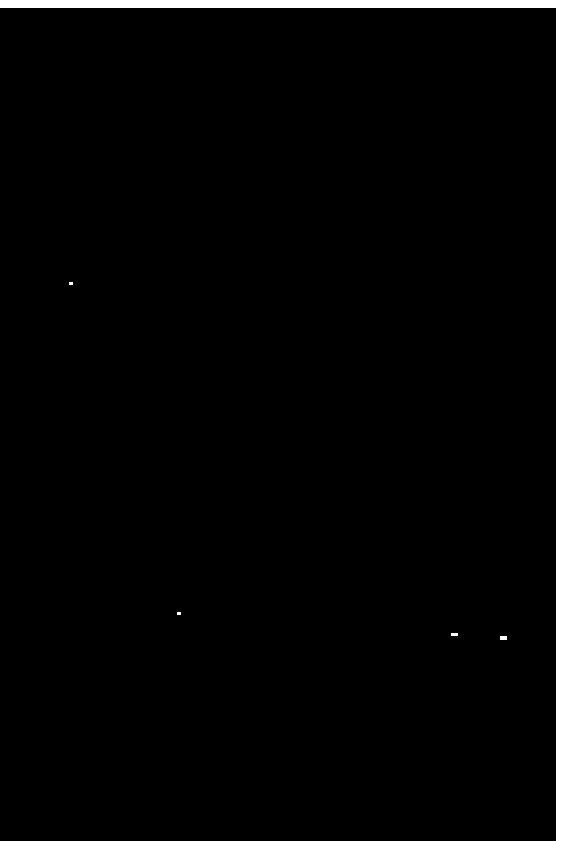}
\includegraphics[width=20mm,height=30mm]{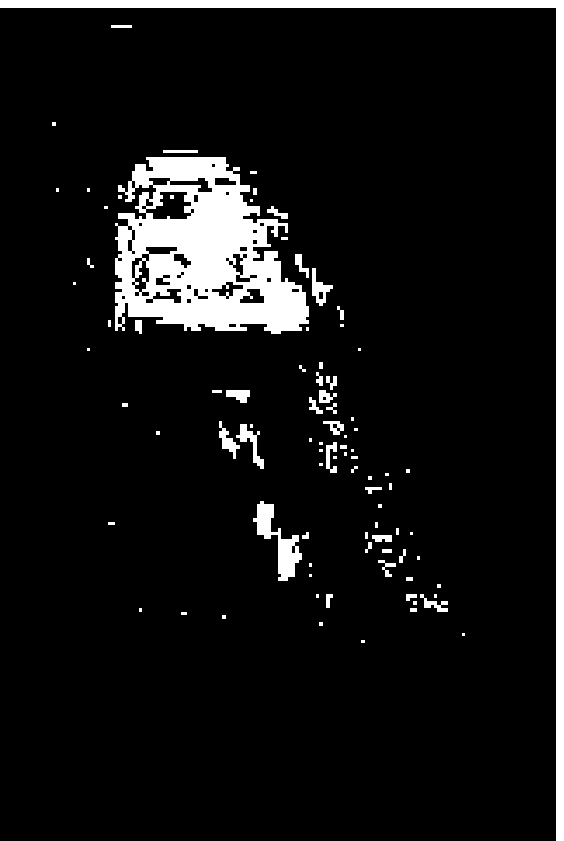}
\includegraphics[width=20mm,height=30mm]{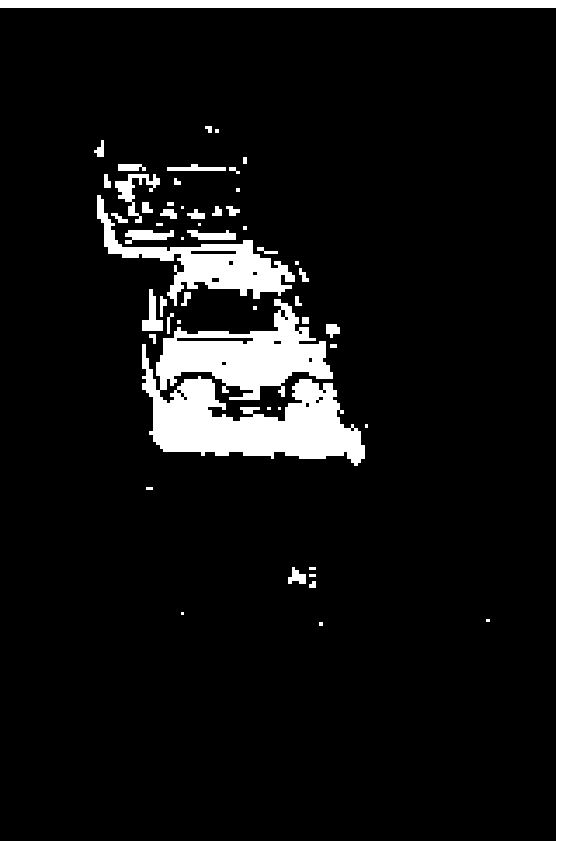}
\includegraphics[width=20mm,height=30mm]{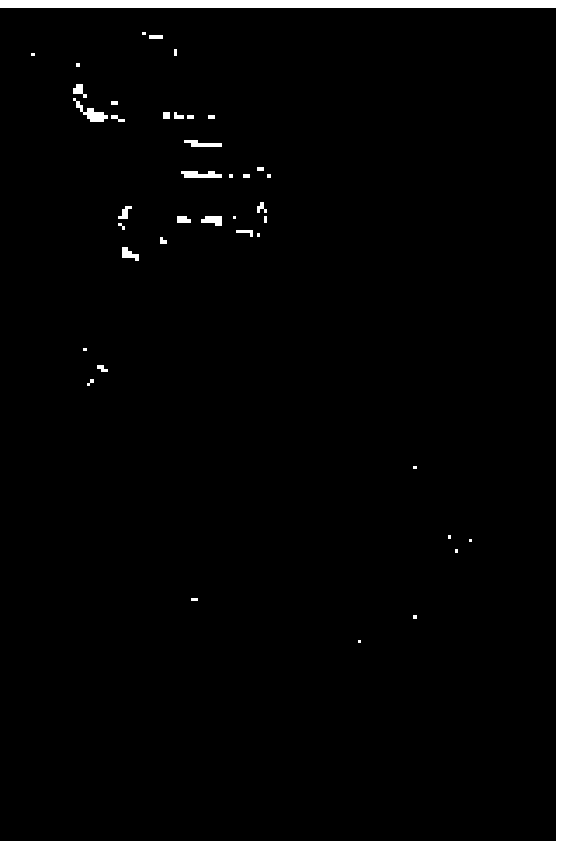}
}

\subfigure[LBP based]{
\includegraphics[width=20mm,height=30mm]{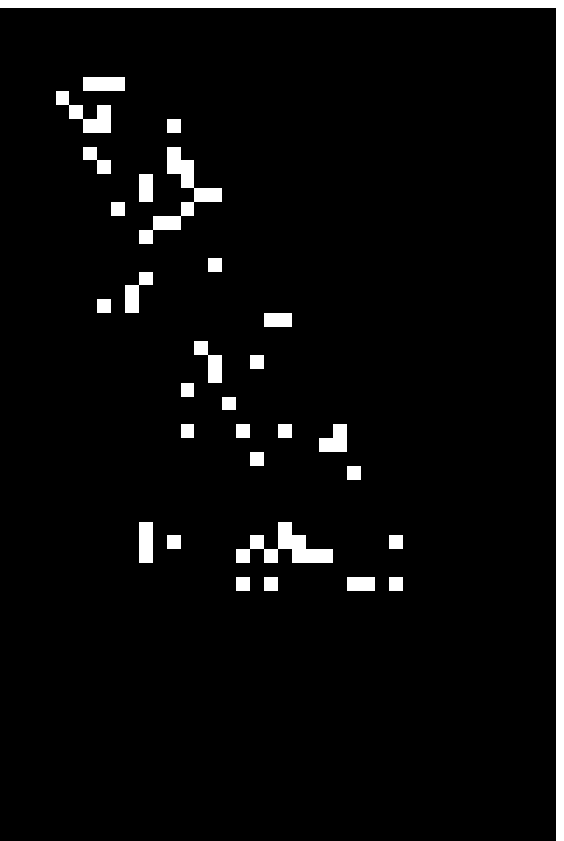}
\includegraphics[width=20mm,height=30mm]{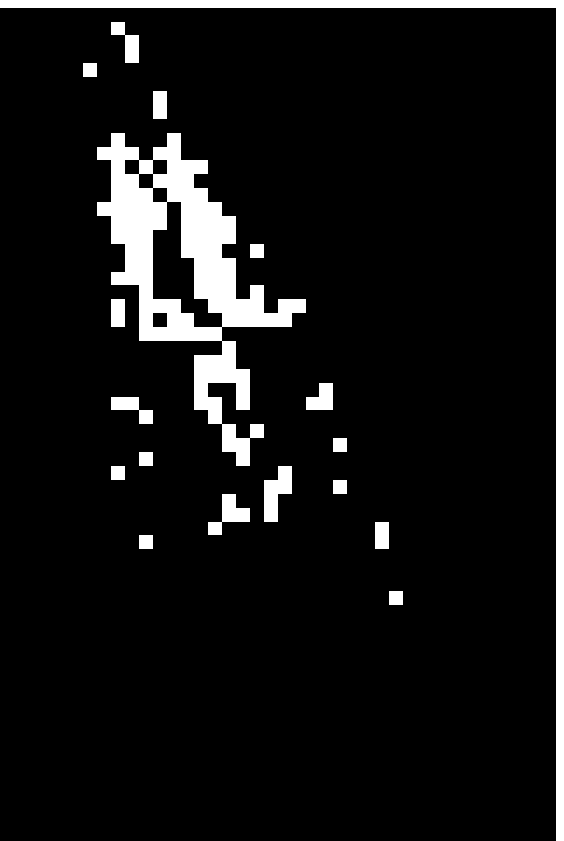}
\includegraphics[width=20mm,height=30mm]{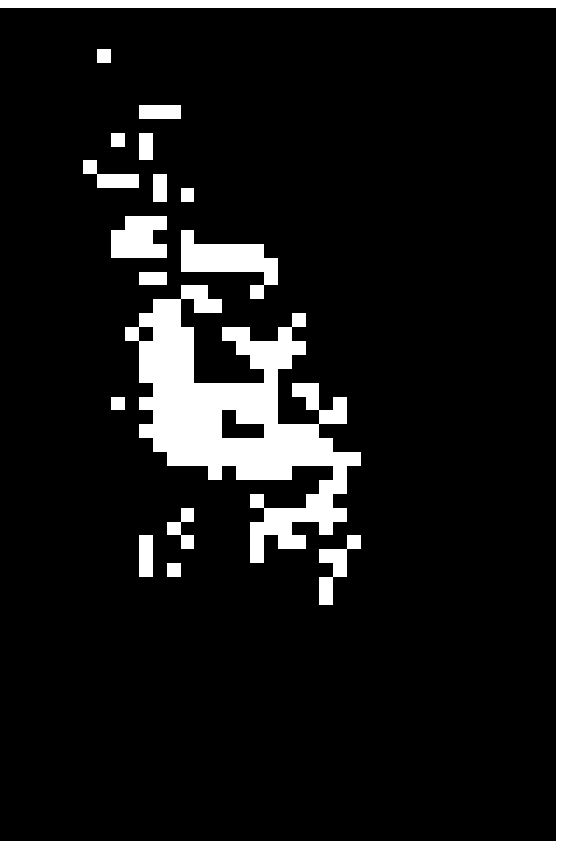}
\includegraphics[width=20mm,height=30mm]{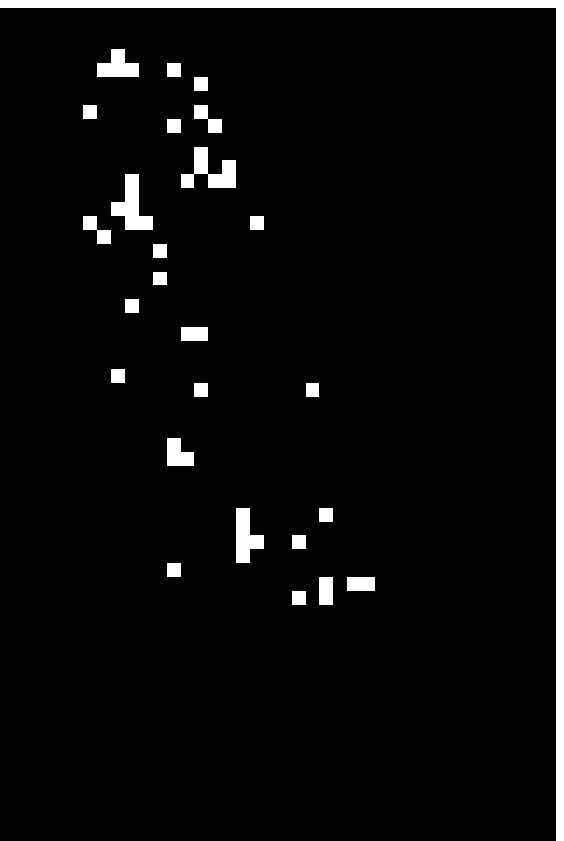}
}

\caption{Sample \textit{Empty}, \textit{Low}, \textit{High} and \textit{Full}  video frames from  \textit{Jtn.-1} along with corresponding foreground detection outputs for all frameworks.}
\label{fig:TD_Jtn1_outputs}
\end{center}
\end{figure}

Figure~\ref{fig:TD_Jtn1_outputs} shows foreground detection results for all frameworks on \textit{Jtn.-1} video sequence for frames with \textit{Empty}, \textit{Low}, \textit{High} and \textit{Full} traffic density states. The detection results for the \textit{Full} frame using GMM, Sigma-Delta and LBP  approaches deteriorate when vehicles are stationary for a span of over $50$ frames. One has to bear in mind that the parameters for each of the methodologies were fine-tuned using the \textit{Multiple$\_$Flows} video for optimal performance and those parameters were \textit{retained} for all experiments with all other input videos. Thus, each method was not fine-tuned or tailored to achieve optimal performance for each input video individually. However, we do believe that by fine-tuning the learning-rate for individual videos, one could achieve optimal detection of stationary vehicles  using the GMM, Sigma-Delta and LBP approaches. There is a trade-off between detecting noisy pixels as foreground pixels and detecting stationary foreground vehicles while fine tuning the learning parameter in the GMM and Sigma-Delta based frameworks. 

\begin{figure*}
\begin{center}
\subfigure[\textit{Empty}]{
\includegraphics[width=1.5in,height=2in]{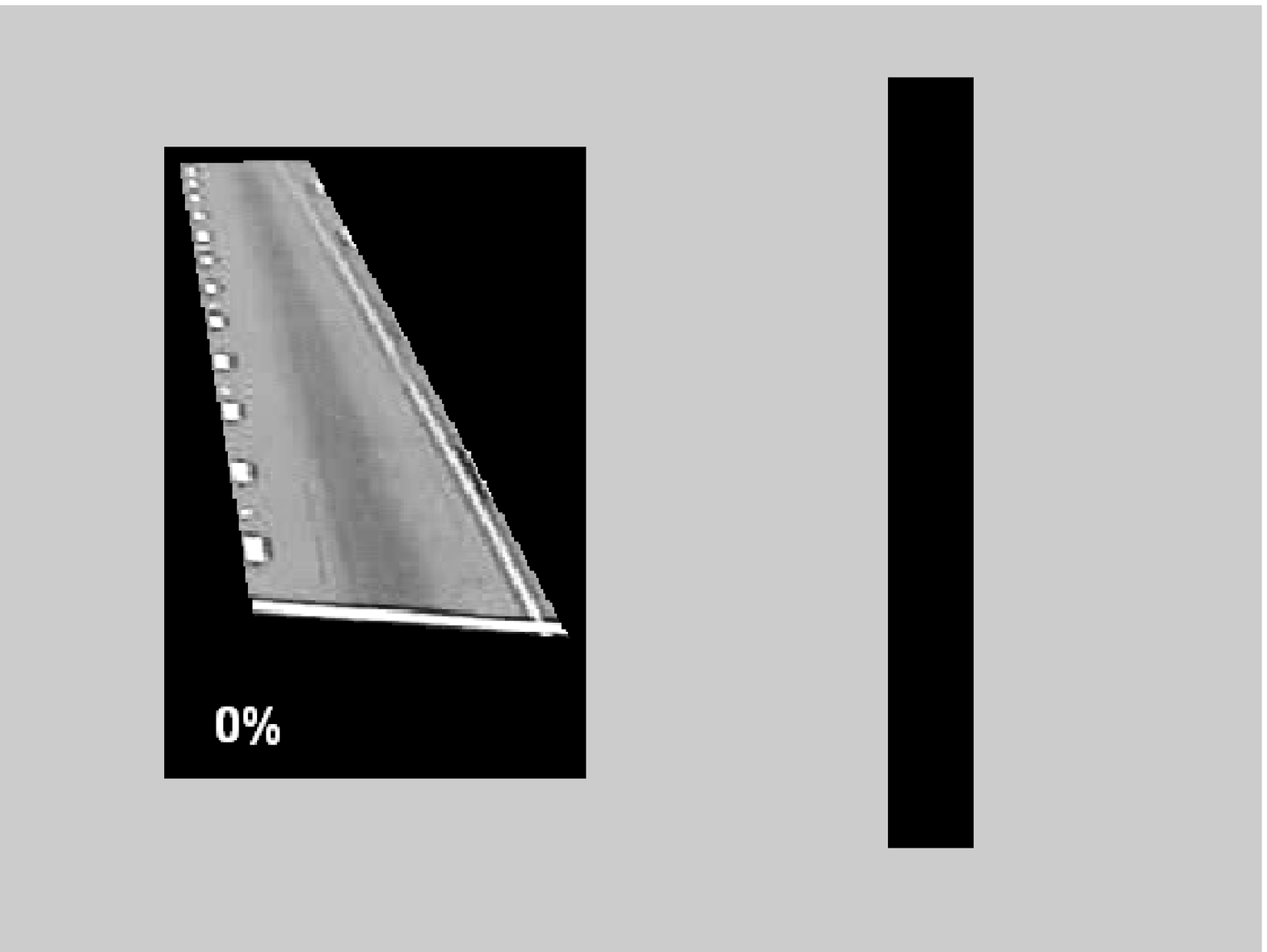}
}
\subfigure[\textit{Low}]{
\includegraphics[width=1.5in,height=2in]{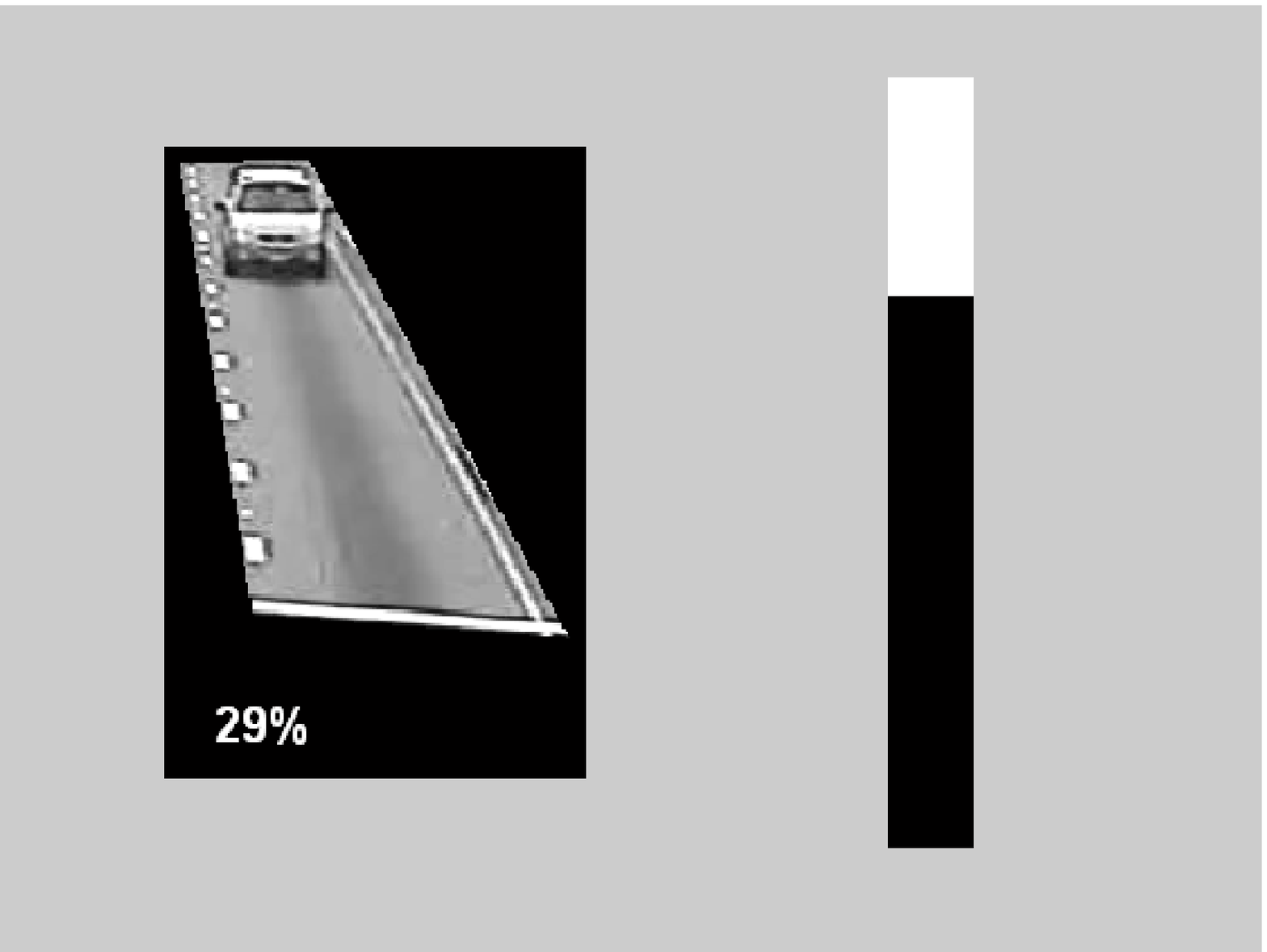}
}
\subfigure[\textit{High}]{
\includegraphics[width=1.5in,height=2in]{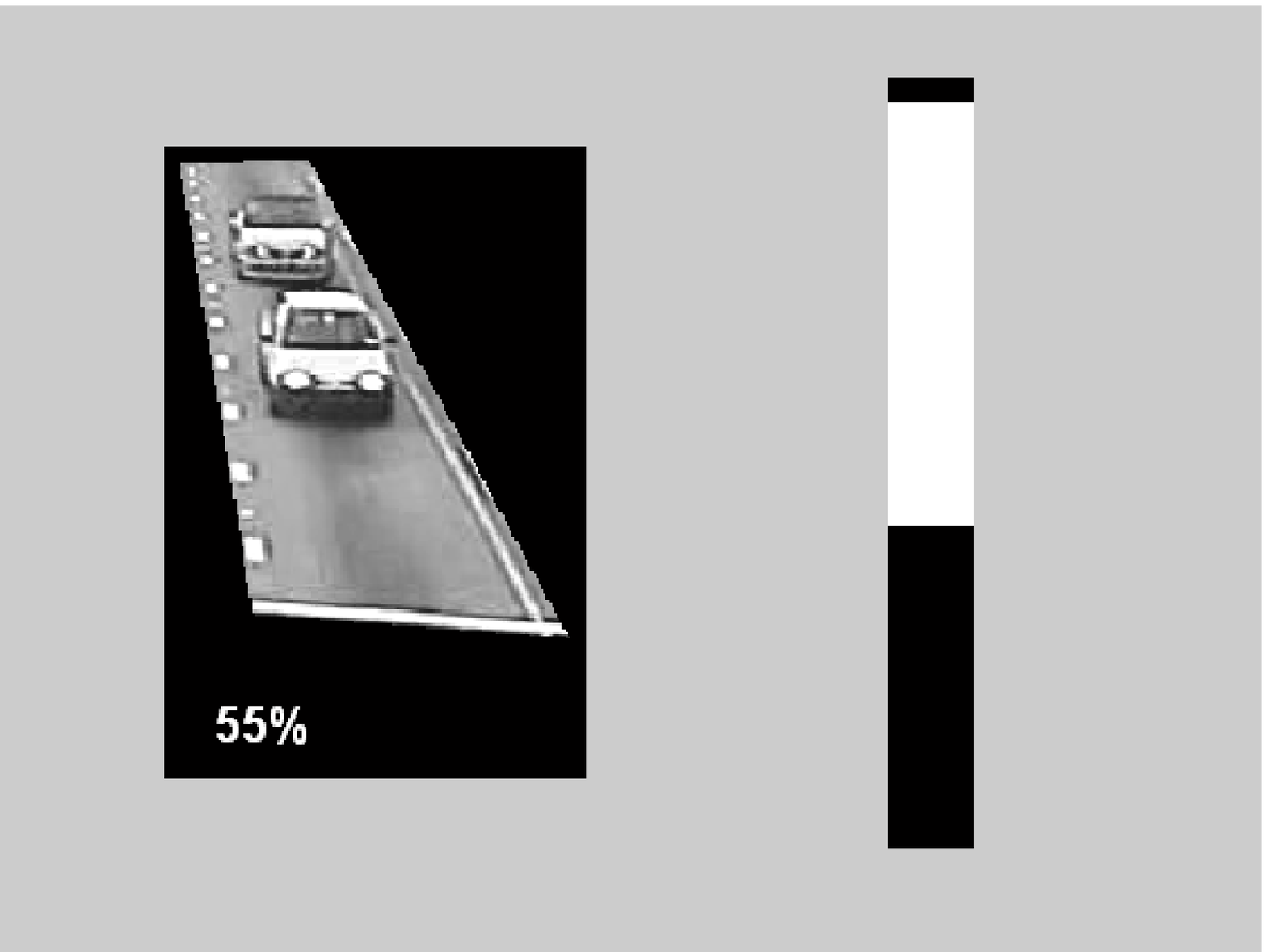}
}
\subfigure[\textit{Full}]{
\includegraphics[width=1.5in,height=2in]{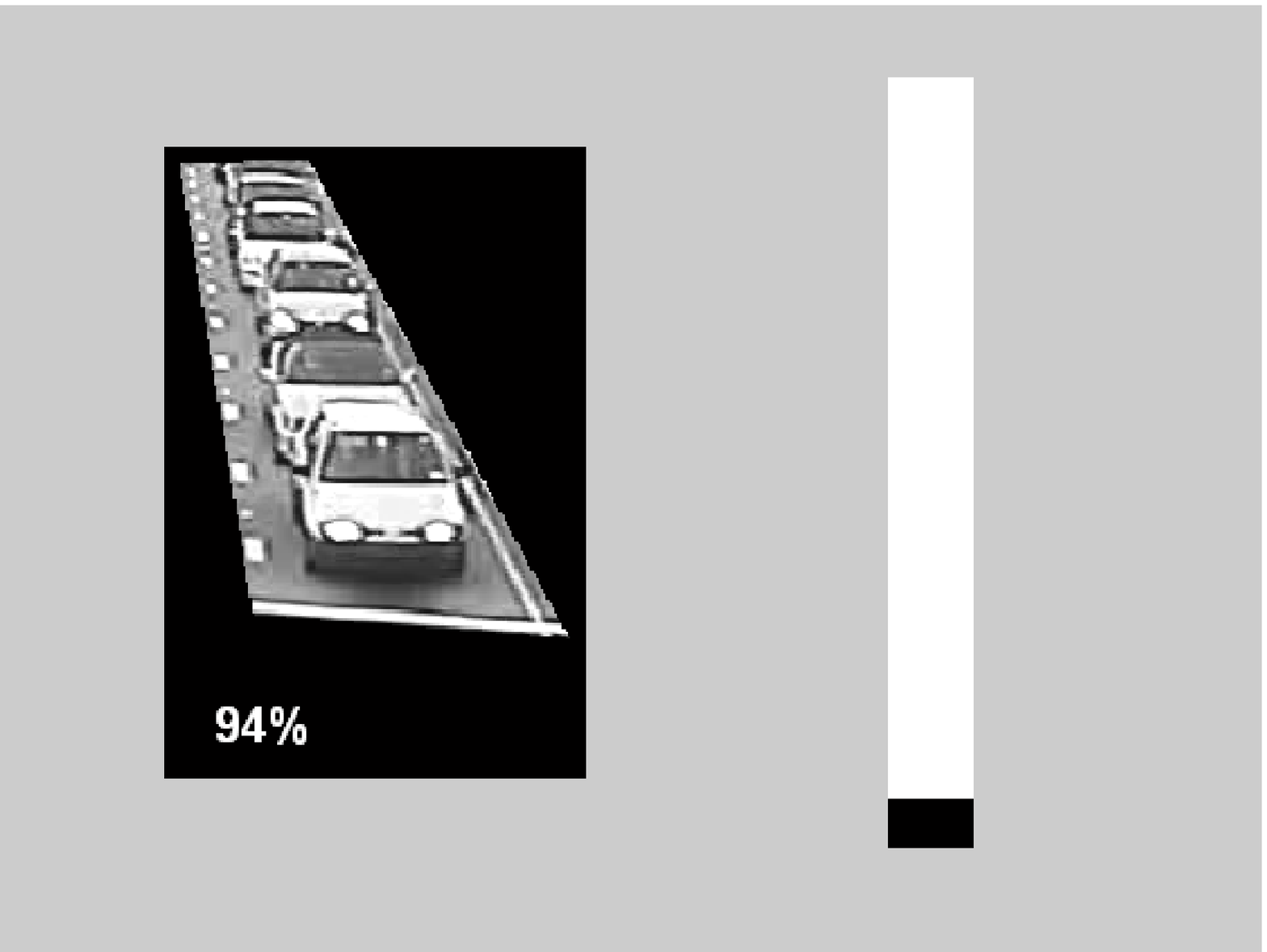}
}
\caption{Sample output for traffic density estimation using proposed moving-object detection approach. The vertical bar is an indicator for lane coverage.}
\label{fig:MOD-sample}
\end{center}
\end{figure*}

Table~\ref{tab:MOD-results} presents the performance results for each sequence tested independently without any training phase. The density state classifications produced by the framework for all frames on a given video were compared to human-annotated ground truth in order to calculate the percentage accuracy of the system.  On average, the proposed moving object detection-based  approach achieves an accuracy of around $86\%$ compared to $79\%$, $74\%$ and $75\%$ for the GMM, Sigma-Delta and LBP-based approaches, respectively. The GMM-based framework does achieve comparable results to the proposed moving object detection-based approach for \textit{Jtn.-1}. However, in the presence of varying illumination conditions such as rain, sun-glare, the performance of the GMM-based framework deteriorates. For all approaches we were able to process upwards of $15$ frames per second for an non-optimized MATLAB$^\mathrm{\textregistered}$ implementation running on a PC with a $2.53$ GHz processor.  Sample outputs of the foreground object detection-based vehicular traffic estimation for all traffic density states is shown in Figure~\ref{fig:MOD-sample}.

\begin{table}
\caption{Performance results for traffic density estimation using different foreground-object detection methodologies.}
\label{tab:MOD-results}
\begin{tabular}{|l|c|c|c|c|}
\hline
Sequence &  GMM       & Sigma-Delta    & LBP & Proposed  \\
         & ~\cite{gmm}& ~\cite{sigdel} & ~\cite{lbp} & \\
\hline
Jtn$.-1$ &  $85.6\%$ & $78.2\%$ & $79\%$  & $86.3\%$\\
Jtn$.-2$ &  $74.1\%$ & $72.3\%$ & $68.2\%$& $85.9\%$\\
Jtn$.-3$ &  $82.1\%$ & $78.3\%$ & $78.9\%$& $86.4\%$\\
Jtn$.-4$ &  $76.3\%$ & $70.7\%$ & $76.2\%$& $87.1\%$\\
\hline
\end{tabular}
\end{table}

Table~\ref{tab:diff_TLD_MOD} highlights the differences between the classical density state modelling-based framework and the proposed foreground object detection-based framework for vehicular traffic density estimation. There is a trade-off between the number of frames that can be processed, accuracy and the computational power required when choosing between either of the two frameworks. However, in our opinion, a foreground object detection-based framework would be the better suited for real-world applications due to the absence of supervised learning modules when compared to a density state modelling-based framework for traffic density estimation.

\begin{table}
\begin{center}
\caption{Differences between texture-model and foreground-object detection based frameworks for an un-optimised MATLAB$^\mathrm{\textregistered}$ implementation on a PC with a $2.53$ GHz processor with $320 \times 240$ frame resolution  }
\label{tab:diff_TLD_MOD}
\begin{tabular}{|l|c|c|}
\hline
  \multirow{2}{*}{Type} &  Density state modelling-based                     & Foreground object \\
       & ~\cite{Faith04,Pork04,Jing07,pranamits09}& detection-based\\
 &  & \\
\hline
 Frames/sec &   2-3 & 15\\
\hline
 Density estimation & \multirow{2}{*}{$92\%$} & \multirow{2}{*}{$86\%$} \\
  accuracy          &  &\\
 \hline
 Learning  & Multi-class SVM, & Mixture Models,\\
 algorithm & $k$-Means,       & simple background-\\
           & HMM-Autoclass    &   filter\\
 \hline
 Supervised &  \multirow{2}{*}{Required} & \multirow{2}{*}{Not-required}\\
   learning &  & \\
 \hline
 Complexity & Complicated & Simple \\
 \hline
 Processor & Needs powerful & Can work on \\
           & processor      & $1$ GHz chipsets \\
\hline
\end{tabular}
\end{center}
\end{table}
%


\section{Conclusion}
\label{sec:final_conclusion}

A methodology to detect moving objects in video using a texture representation  has been presented in this paper. Experimental results have shown that the methodology is resilient to noise, illumination changes, low frame rate and other camera-associated noise. The proposed methodology is capable of working in real-time (i.e., $25$ fps) and does not require parameter fine-tuning for individual videos. The performance of the proposed approach exceeds the performance of other algorithms that are widely used. Traffic videos captured during night time or with long-shadows are still a challenge which is yet to be addressed using the proposed methodology. We have also presented a framework for vehicular traffic density estimation using a foreground object detection-based approach. The foreground object detection-based framework for vehicular traffic density estimation is able to process frames in real-time and comes with an added advantage of not requiring supervised learning.


We have been successful in developing a fundamental process, such as moving-object detection, and showcasing its usage for video analysis applications such as vehicular traffic density estimation. We would like to take this opportunity to point out  the need for  an annotated traffic video database which could be used as ground truth  for research in this area. Such databases are widely available in other image processing specialities (such as face recognition) but are sadly lacking here, making performance comparison difficult. In due course we hope to make our data publicly available. 

%



\section*{Acknowledgment}
NICTA is funded by the Australian Federal Government as
represented by the Department of Broadband,
Communications and the Digital Economy, the NSW
Department of State and Regional Development, the ACT
Government and the Australian Research Council through
the ICT Centre of Excellence Program.



%


\bibliographystyle{siam}



\bibliography{Thesisreferences}

\end{document}